\documentclass[sigconf]{acmart}

\setcopyright{none}                               
\renewcommand\footnotetextcopyrightpermission[1]{}
\usepackage{tikz}
\usepackage{multirow}
\usepackage{xcolor}
\usepackage{colortbl}
\usepackage{graphicx}
\usepackage{subcaption}
\usepackage{tabularx}
\usepackage{array}
\usepackage{wrapfig}
\usepackage{verbatim}
\usepackage{booktabs}
\usepackage{makecell}
\usepackage{adjustbox}
\usepackage{fontawesome5}
\usepackage{pifont}
\usepackage{caption}
\usepackage{tcolorbox}

\tcbuselibrary{skins}
\usetikzlibrary{arrows.meta}

\newcommand{\yes}{$\checkmark$}
\newcommand{\no}{$-\!-$}
\newcommand{\cmark}{\ding{51}}
\newcommand{\xmark}{\ding{55}}

\definecolor{cok}{HTML}{A3C4A8}
\definecolor{cerr}{HTML}{D4848A}
\definecolor{correct}{RGB}{180,220,180}
\definecolor{wrong}{RGB}{240,180,180}
\definecolor{noans}{RGB}{220,220,220}
\definecolor{headerblue}{RGB}{220,235,255}
\definecolor{nullgray}{RGB}{235,235,235}

\definecolor{stepcolor}{RGB}{135,206,250}
\definecolor{imgAcolor}{RGB}{255,187,120}
\definecolor{imgBcolor}{RGB}{152,223,138}
\definecolor{stepbg}{RGB}{240,248,255}
\definecolor{stepframe}{RGB}{100,149,237}
\definecolor{prismbg}{RGB}{245,240,255}
\definecolor{prismframe}{RGB}{150,120,200}

\newlength{\tW}   
\newlength{\tH}   
\newlength{\tlW}  

\AtBeginDocument{%
  }

\acmConference[MM '26]{The 34th ACM International Conference on Multimedia}{November 10--14, 2026}{Rio de Janeiro, Brazil}
\acmYear{2026}
\copyrightyear{2026}
\acmDOI{10.1145/3767308.3832558}
\acmISBN{978-1-4503-XXXX-X/26/11}

\renewcommand{\thetable}{S\arabic{table}}
\renewcommand{\thefigure}{S\arabic{figure}}
\renewcommand{\thesection}{\Alph{section}}

\renewcommand{\arraystretch}{1.15}
\begin{document}

\title{Order Matters: LVLMs as Judges for \\Temporal Reasoning in Image Sequences}

\author{Martina Ianaro}
\affiliation{%
  \institution{University of Bologna}
  \city{Bologna}
  \country{Italy}
}
\email{martina.ianaro3@unibo.it}

\author{Guilherme Fernandes}
\affiliation{%
  \institution{NOVA School of Science and Technology}
  \institution{NOVA Laboratory for Computer Science and Informatics}
  \city{Lisbon}
  \country{Portugal}
}
\email{guidcf28@gmail.com}

\author{Maurizio Gabbrielli}
\affiliation{%
  \institution{University of Bologna}
  \city{Bologna}
  \country{Italy}
}
\email{maurizio.gabbrielli@unibo.it}

\author{João Magalhães}
\affiliation{%
  \institution{NOVA School of Science and Technology}
  \institution{NOVA Laboratory for Computer Science and Informatics}
  \city{Lisbon}
  \country{Portugal}
}
\email{jmag@fct.unl.pt}

\renewcommand{\shortauthors}{Ianaro et al.}

\begin{abstract}
As generative multimedia evolves from static image synthesis to complex, interleaved visual narratives, a foundational bottleneck has emerged: the \textit{judgment crisis}. While human perception naturally synthesizes the temporal and logical flow of a story, automated evaluation systems remain largely ``blind'' to sequential continuity, often failing to distinguish between a coherent narrative and a semantically shuffled or contradictory sequence. This work identifies a critical structural gap in current multimodal evaluation paradigms, arguing that the reliance on Large Vision-Language Models (LVLMs) as judges is fundamentally limited by architectural biases. Our analysis reveals a profound performance dichotomy: while models may appear competent in isolated pointwise scoring, they suffer a catastrophic collapse when required to perform pairwise discrimination of temporal order. We demonstrate that this is not merely a data-scarcity issue but a structural one. Through a series of diagnostic probes, we uncover systematic positional asymmetries, specifically \textit{primacy} and \textit{recency} effects, where a model's judgment of a story is significantly influenced by the placement of a frame, often more than by its semantic consistency. These biases, potentially rooted in causal masking and rotary embeddings, suggest that current transformer-based judges are inherently ill-equipped for long-form visual reasoning. By exposing these ``blind spots,'' we challenge the multimedia community to move beyond snapshot-centric metrics and instead pioneer \textit{Temporally-Aware Evaluation} paradigms that treat visual sequences as unified logical structures rather than unordered collections of frames.
\end{abstract}





\maketitle


\section{Introduction}
\label{sec:intro}
The landscape of generative multimedia is undergoing a radical transition, moving from the synthesis of isolated, static images to the creation of complex, interleaved visual narratives \cite{huang2024surveyevaluationmultimodallarge, chen2024mllm}. Whether in automated storyboarding, procedural video generation, or multi-turn visual dialogues, the value of these systems is no longer defined by pixel-perfect fidelity, but by their ability to maintain temporal and logical continuity across a sequence \cite{Huang_2024_CVPR, Chen_2025_CVPR}. However, as generative capabilities accelerate, our ability to evaluate the narrative logic of these systems has remained stagnant. We argue that the multimedia community is currently facing a foundational \textit{judgment crisis}.

\begin{figure}
  \centering
  \includegraphics[width=\linewidth]{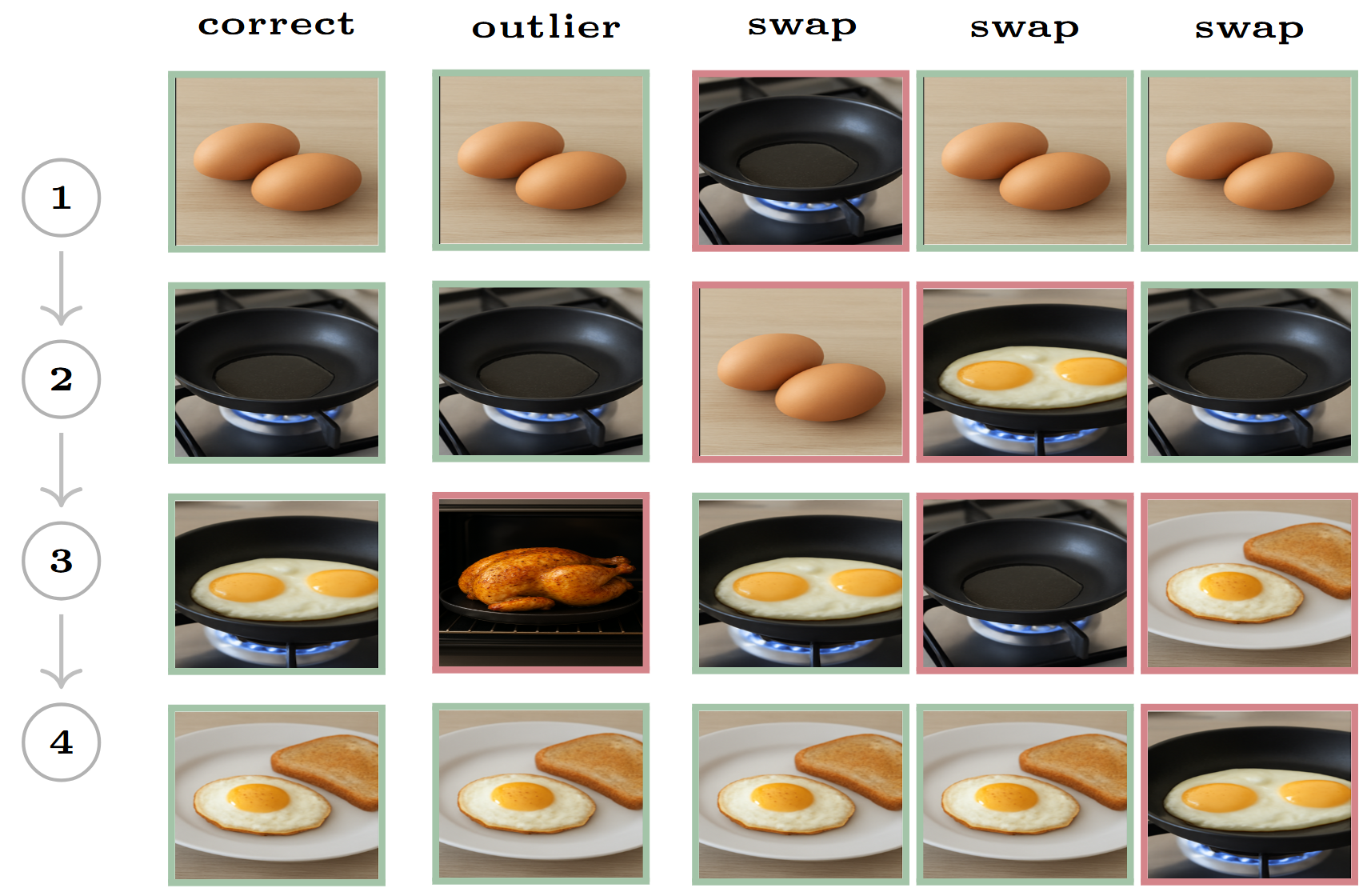} 
  \Description{A grid of cooking photographs. Each column shows a four-step
  fried-egg recipe read top to bottom, with arrows indicating the intended
  order. The leftmost column is the correct reference sequence. The remaining
  columns are perturbed variants in which one frame has been replaced by an
  out-of-context image, or two steps have been exchanged; the altered frames
  are outlined in red.}
\caption{\textbf{Probing temporal and semantic reasoning.} Each column shows the
same four-step procedure (frying an egg) read top to bottom, with perturbed
frames outlined in red. Left to right: the \textsc{correct} reference; a
\textsc{semantic outlier}, where one frame comes from an unrelated recipe; and
three \textsc{temporal swaps} exchanging two steps at different positions.
Because every variant contains individually plausible images, order-blind
metrics score them alike, isolating whether an LVLM judge tracks procedural
logic or relies on positional heuristics.}
  \label{fig:teaser}
\end{figure}

\paragraph{The ``Bag of Frames'' Fallacy}
Currently, the evaluation of multi-image sequences remains tethered to ``pointwise'' metrics, such as FID or CLIPScore, which assess images in total isolation \cite{huang2024surveyevaluationmultimodallarge}. These paradigms are effectively ``blind'' to order; they cannot distinguish a logically sound visual progression from a semantically identical set of frames that have been randomly shuffled. This phenomenon is illustrated in Figure~\ref{fig:teaser}.
As a result, the community has increasingly turned toward LVLMs as ``off-the-shelf'' judges, assuming that their success in single-image reasoning will naturally translate to sequential logic \cite{chen2024mllm, xiong2025llavacriticlearningevaluatemultimodal}.

\paragraph{A Brave New Idea: Chronological Intelligence}
This requires us to measure what we term ``chronological intelligence'' as the ability of a system to understand the causal and sequential ``why'' behind a visual transition.
Our work uncovers a startling performance dichotomy: even state-of-the-art LVLMs, when acting as judges, exhibit a profound ``reasoning chasm.'' While these models assign plausible scores to individual sequences in isolation, their ability to discriminate between ordered and shuffled sequences collapses when required to perform pairwise discrimination. We demonstrate that this failure is rooted in structural blindness: positional asymmetries, specifically \textit{primacy} and \textit{recency} effects,
significantly influence a model's judgment, often more than the actual logical flow of the images \cite{wu2025emergencepositionbiastransformers, liu2023lostmiddlelanguagemodels}.

\paragraph{Vision and Contributions}
This work serves as a critical probe to expose these structural blind spots and redirect the focus of the multimedia community from simple dataset expansion to the development of evaluation paradigms that treat time as a first-class logical citizen. To catalyze this transition, our work provides three key contributions:
\begin{itemize}
    \item \textbf{Analysis of the Reasoning Chasm:} We uncover the performance dichotomy between pointwise and pairwise judging, demonstrating how current paradigms fail to capture the architectural and cognitive constraints of sequence reasoning.
    \item \textbf{Identification of Structural Biases:} We reveal that systematic positional asymmetries persist even after fine-tuning, suggesting that temporal reordering remains a fundamental challenge for current transformer-based LVLM architectures.
    \item \textbf{A Visual-Order Reasoning Tasks:} We release the first dataset to assess LVLM-judges accuracy in sequential multi-image reasoning tasks. It measures LVLM judges abilities to recognize unordered visual sequences, outliers in sequences, semantically poorly-aligned elements of sequences.
\end{itemize}

Our findings expose structural blind spots, laying the groundwork for chronologically-aware generative AI evaluation.

\section{Related Work}
\label{sec:related_work}

\paragraph{\textbf{From Static Recognition to Narrative Reasoning.}}
Traditional evaluation relies on reference-based, pointwise metrics such as FID 
and CLIPScore to assess visual fidelity \cite{huang2024surveyevaluationmultimodallarge}. 
While effective for static quality, these metrics are fundamentally ``order-blind,'' 
treating a visual sequence as a randomized ``bag of frames.'' This limitation has 
sparked a move toward \textit{Temporally-Aware Evaluation} of ``chronological 
intelligence''.
Recent benchmarks probe this frontier, including Panda-70M for large-scale 
captioning \cite{Chen_2024_CVPR}, Assembly101 for procedural activities 
\cite{sener2022assembly101largescalemultiviewvideo}, and cross-task instructional 
substrates \cite{zhukov2019crosstaskweaklysupervisedlearning}. Despite strides via 
contextual calibration \cite{kang2025calibclip} or brain-inspired world models 
\cite{gao2025building}, most frameworks treat temporal order as an implicit byproduct 
rather than a core constraint \cite{Huang_2024_CVPR}. Parallel synthesis methods, 
including retrieval-augmented image generation~\cite{koh2023generatingimagesmultimodallanguage}, 
free-lunch consistent generation~\cite{liu2025onepromptonestoryfreelunchconsistenttexttoimage}, 
illustrated instructions~\cite{menon2024generatingillustratedinstructions}, and 
long-range self-consistent diffusion~\cite{zhou2024storydiffusionconsistentselfattentionlongrange}, 
raise the visual bar for sequential narratives, yet remain agnostic to logical ordering. 
Interleaved datasets such as CoMM~\cite{Chen_2025_CVPR} and procedural corpora like 
COIN~\cite{tang2019coinlargescaledatasetcomprehensive} provide empirical substrate for 
studying temporal structure, while unified consistency frameworks~\cite{11215765} and 
temporal visual semantics approaches~\cite{11399654} show the community's push beyond 
snapshot fidelity. Our work builds on these foundations, requiring non-trivial logical 
operations across controlled perturbations.
There are some to text-to-video evaluation 
suites such as T2V-CompBench~\cite{sun2025t2v} and temporal 
reasoning probes including STEP~\cite{qiu2025step}, Deep Temporal 
Reasoning~\cite{loginova2025deep}, VideoQA-TA~\cite{wu2025videoqa}, 
and ReasVQA~\cite{liang2025reasvqa}, all of which target 
video-native models rather than the interleaved multi-scene regime 
addressed in this paper. Spatial reasoning surveys~%
\cite{zheng2025multimodalspatialreasoninglarge} and 
efficiency-oriented transformer architectures such as 
ASTER~\cite{liu2025aster} further contextualise the broader 
landscape in which our work is positioned.

\paragraph{\textbf{The LVLM-as-a-Judge: A Reasoning Chasm.}}
To overcome static metrics, the community has increasingly adopted LVLMs as automated zero-shot judges \cite{lee2024prometheusvisionvisionlanguagemodeljudge, li2025judgesjudgellmjuryondemand, li2023generativejudgeevaluatingalignment}. However, this paradigm is plagued by a profound ``reasoning chasm.'' While models like GPT-4V \cite{OpenAI2023GPTV} or Gemini \cite{comanici2025gemini25} excel at static frame description \cite{xiong2025llavacriticlearningevaluatemultimodal}, assessing interleaved logic is hindered by failures in aligning multi-step operations with visual grounding \cite{qi2025vcr, fang2024mmbench}.
Progress in video understanding (e.g., TempCompass \cite{liu2024tempcompass}, MMWorld \cite{he2024mmworld}) and reward-oriented extensions like CAREVL \cite{dai2025captionsrewardscarevlleveraging} or InternLM-XComposer2.5 \cite{zang2025internlmxcomposer25rewardsimpleeffectivemultimodal} have narrowed the gap, yet they primarily operate in single-image or clip-level regimes. Recent audits demonstrate that even frontier oracles exhibit systematic calibration failures \cite{gordon2025unblockingfinegrainedevaluationdetailed} and struggle with distributional alignment \cite{chen2025beyond, LI2026104270}. We argue this is a structural crisis of the ``LVLM-as-a-Judge'' paradigm \cite{huang-etal-2025-empirical}, where visual quality is consistently prioritized over temporal consistency \cite{Qi_2025_CVPR}. Broader surveys on aligning multimodal LLMs~\cite{yu2025aligning} and fine-grained visual benchmarking~\cite{yu2025benchmarkinglargevisionlanguagemodels} confirm that judges systematically underweight temporal coherence in favour of perceptual quality. This is further evidenced by studies on image quality criteria~\cite{parthasarathy2025makesgoodgeneratedimage} and fairness audits revealing multi-dimensional biases in evaluation pipelines~\cite{adewumi2024fairnessbiasmultimodalai}. Text-to-visual evaluators~\cite{lin2024evaluating}, unified multimodal CoT reward models~\cite{wang2025unifiedmultimodalchainofthoughtreward}, and pairwise correlation modelling~\cite{jiang2025bridgingmodelingcorrelationspairwise} further illustrate the breadth of the unresolved evaluation landscape.

\paragraph{\textbf{Architectural and Cognitive Constraints}}
These failures are not merely data-driven but rooted in architectural biases. 
Causal masking and Rotary Position Embeddings  (RoPE) \cite{su2023roformerenhancedtransformerrotary} 
often manifest as serial position effects, primacy and recency biases, that lead 
models to neglect a sequence's logical core \cite{wu2025emergencepositionbiastransformers, 
liu2023lostmiddlelanguagemodels}. This phenomenon is linked to ``attention sinks'' 
\cite{gu2024attention}, identified even in diffusion-based architectures 
\cite{rulli2025attention, lu2025artifactsattentionsinksstructured}. Crucially, 
instruction tuning and RLHF \emph{amplify} rather than suppress these 
tendencies~\cite{itzhak-etal-2024-instructed, tjuatja-etal-2024-llms}, making 
fine-tuned judges more susceptible to positional shortcuts than base counterparts. 
These failures also mirror findings in cognitive psychology regarding human memory 
and causal judgment~\cite{rottman2025learning, sinclair2024first}. LVLMs exhibit 
marked sensitivity to ordering \cite{pezeshkpour-hruschka-2024-large}, where reward 
over-optimization can exacerbate primacy biases \cite{zhang2024confronting}. Via diagnostic probes and anchoring techniques like SinkTrack 
\cite{liu2026sinktrack}, we determine whether automated judges evaluate narrative 
logic or fall victim to these constraints \cite{shi-etal-2025-judging, 
drissi2024lesssimulationbasedapproachdynamic}. Serial position effects are 
characterised systematically in LLMs~\cite{guo2025serial}, MCQ sensitivity to option 
ordering further exposes the fragility of transformer-based 
discrimination~\cite{zheng2024largelanguagemodelsrobust}, and bias 
quantification~\cite{ye2024justiceprejudicequantifyingbiases} with mechanistic 
debiasing~\cite{wang2025eliminatingpositionbiaslanguage} represent the state of the 
art in mitigating, yet not resolving, these architectural constraints.

\section{Theoretical Background}
\label{sec:theory}


\noindent We refer to the failure mode described in Section~\ref{sec:intro} as the
\textit{judgment crisis}: evaluation tools that cannot distinguish a logical
progression from a semantically identical permutation~\cite{huang2024surveyevaluationmultimodallarge, chen2024mllm}.

\vspace{2mm}
\noindent\textbf{Evaluation Paradigms.}
We formalize sequential coherence evaluation under two complementary frameworks.

In \emph{pointwise scoring}, a judge $\mathcal{J}$ maps a sequence $s_i$ to an 
independent scalar rating $\hat{y}_i = \mathcal{J}(s_i;\pi) \in \{1,\dots,5\}$. 
Alignment with human references is quantified via Pearson correlation ($r$) and error 
residuals (MAE, RMSE)~\cite{zheng2023judgingllmasajudgemtbenchchatbot, chen2024mllm}. 
This paradigm is inherently susceptible to \textbf{distributional bias}, where a 
systematic skew in the judge's score range, such as the \textit{binary collapse} 
observed in generative judges, can artificially inflate or deflate correlations 
independently of true discriminative accuracy ~\cite{chen2025beyond, LI2026104270, adewumi2024fairnessbiasmultimodalai}. 
Our findings confirm this: models assign 
dangerously similar high scores to both coherent and shuffled sequences, hallucinating consistency from local semantic cues alone.

In \emph{pairwise comparison}, the judge returns a ternary preference label 
$d = \mathcal{J}(s_a,s_b;\pi) \in \{\texttt{A}, \texttt{B}, \texttt{UNK}\}$, grounded 
in the Bradley--Terry model~\cite{sun2025rethinkingbradleyterrymodelspreferencebased}:
\begin{equation}
    \Pr[s_a \succ s_b] = \frac{e^{r(s_a)}}{e^{r(s_a)} + e^{r(s_b)}}
    \label{eq:bt}
\end{equation}
where $r$ denotes a latent reward function. While more robust to distribution shifts, 
pairwise protocols introduce \textbf{positional bias}. A judge is defined as 
\textit{position-invariant} if and only if:
\begin{equation}
    \Pr[\mathcal{J}(s_a,s_b) = \texttt{A}] = \Pr[\mathcal{J}(s_b,s_a) = \texttt{B}] 
    \quad \forall (s_a,s_b)
    \label{eq:invariance}
\end{equation}

We observe empirically a first-option bias of up to 15\%, motivating randomized 
presentation order. This \textbf{performance dichotomy}, high pointwise scores 
coexisting with near-random pairwise accuracy ($\approx$50\%), demands a new 
generation of \textit{Temporally-Aware Evaluation} paradigms.

\vspace{2mm}
\noindent\textbf{Structural Origins of Bias.}
Positional asymmetries in transformer-based judges emerge from the interplay of two 
architectural mechanisms~\cite{wu2025emergencepositionbiastransformers, 
wang2025eliminatingpositionbiaslanguage}. First, causal masking induces a 
\textbf{primacy bias}; as network depth $L$ increases, hidden representations tend to 
converge exponentially toward the first token:
\begin{equation}
    \bigl\|\mathbf{h}_t^{(L)} - \mathbf{h}_1^{(L)}\bigr\| \leq C\lambda^{L}, 
    \quad \lambda \in (0,1)
    \label{eq:primacy}
\end{equation}
rendering early content structurally more salient~\cite{wu2025emergencepositionbiastransformers}. 
This creates a theoretical disadvantage for bidirectional narrative reasoning, where 
the logic of a middle frame depends equally on what precedes and what follows it. 
Conversely, RoPE~\cite{su2023roformerenhancedtransformerrotary} exert a \textbf{recency bias} 
within attention heads by suppressing interactions between distant tokens, leaving the 
logical core of a narrative, in particular the middle steps of a sequence, theoretically 
invisible to the judge~\cite{liu2023lostmiddlelanguagemodels}.

The resulting depth-dependent competition between global primacy and local 
recency~\cite{guo2025serial} is further \textbf{amplified} by 
instruction tuning and RLHF~\cite{itzhak-etal-2024-instructed, tjuatja-etal-2024-llms}. 
This theoretical framework directly predicts the asymmetries observed in our 
experiments: a dominant primacy effect in semantic anomaly detection and a recency 
pattern in temporal swap detection. Crucially, as these biases operate at the 
representation level and are not surfaced during CoT reasoning, they 
remain resistant to systematic mitigation via reasoning 
augmentation~\cite{ye2024justiceprejudicequantifyingbiases}. Building on these 
theoretical foundations, we now present our experimental setup and the resulting 
performance stratification.

\section{The Sequence-Judge Framework}
\label{sec:framework}

Building on the theoretical constraints identified in
Section~\ref{sec:theory}, the \textit{Sequence-Judge} framework
translates each diagnosed failure mode into a concrete design choice.
Three operative pillars instantiate this translation.

\vspace{2mm}
\noindent\textbf{From Pointwise Bias to Pairwise Robustness.}
Rather than asking the judge to assign an absolute score (to avoid the collapse into distributional
bias~\cite{chen2024mllm}), we cast evaluation as a \textbf{relative
discrimination} task. Given a gold-standard sequence and a perturbed
variant, the model must identify which respects the causal order
described in the textual prompt.
Presentation order is randomised across all pairs to neutralise the first-option bias quantified in Equation~\ref{eq:invariance}~\cite{zheng2024largelanguagemodelsrobust,pezeshkpour-hruschka-2024-large}.

\vspace{2mm}
\noindent\textbf{Reasoning-Augmented Distillation.}

To operationalize the \textit{Analyze-then-Judge} paradigm, we distil step-level rationales from \textsc{Gemini-2.5-Flash}~\cite{comanici2025gemini25} and use them as a supervision signal alongside the scalar score. 
Adaptation is applied to the
\textsc{LLaVA-OneVision} base, whose native multi-image architecture supports the interleaved regime our task requires; the resulting fine-tuned variants are described in Section~\ref{sec:sft}.

This step is necessary as \textsc{LLaVA-Critic}~\cite{xiong2025llavacriticlearningevaluatemultimodal},
despite its evaluation focus, is specialized on the single-image 
dataset, making it less effective for our 
composite and sequential inputs. The objective is defined as:
\begin{equation}
  \mathcal{L}_{\text{total}} = \alpha\,\mathcal{L}_{\text{score}}
                              + \beta\,\mathcal{L}_{\text{CoT}}
  \label{eq:loss}
\end{equation}
where $\mathcal{L}_{\text{CoT}}$ penalises the model when it cannot
articulate the logical transition between consecutive frames before
committing to a score. 
Each prompt configuration P0--P6 and P$\star$ differs in the amount of 
procedural context and the requested output structure; full templates 
and input layouts are provided in the extended version of this paper.

\vspace{2mm}
\noindent\textbf{Bayesian Hyperparameter Optimisation.}
To prevent performance gains from being artefacts of sub-optimal
tuning, we employ \textbf{Optuna}~\cite{10.1145/3292500.3330701} for
automated Bayesian optimisation over 50 trials with search space
$\text{lr} \in [2{\cdot}10^{-6},\,5{\cdot}10^{-5}]$ and
$\text{batch size} \in [4, 16]$. Crucially, the optimisation target
is macro-$F_1$ on \textsc{Prism}-Temporal, the subset most exposed to
the lost-in-the-middle effect, ensuring that hyperparameter selection
directly rewards progress on the hardest structural
bottleneck~\cite{liu2023lostmiddlelanguagemodels,
wu2025emergencepositionbiastransformers,
su2023roformerenhancedtransformerrotary}.

\section{Temporal-Visual CoT Benchmark}
\label{sec:datasets}

In this section, we introduce temporal-visual reasoning tasks as a resource to assess the quality of LLM-judges on well-grounded samples. The first set of samples introduce known perturbations to the sequences to elicit out-of-order reasoning capabilities \textsc{Prism}. The second set of samples, called \textsc{Mirage}, are generated with image sequence generation methods and were manually annotated by humans. 
Visual examples are provided in Figure~\ref{fig:all_sequences}.

\paragraph{\textbf{\textsc{PRISM}}}
\label{subsec:prism}
The \textit{Perturbation-based Reasoning samples for Image Sequence Modeling} is designed to evaluate temporal and logical consistency through a multi-level perturbation strategy. Using cooking procedures as a canonical testbed for procedural reasoning~\cite{tang2019coinlargescaledatasetcomprehensive}, we generate thousands of evaluation instances where violations are objectively verifiable. 
Negatives are organized into two subsets: 
(i) \textbf{\textsc{PRISM}-Semantic} targets semantic coherence, where 1--3 frames are replaced with out-of-context images; 
(ii) \textbf{\textsc{PRISM}-Temporal} targets temporal logic, where frames are reordered via consecutive or non-consecutive swaps. 
This design follows the philosophy of \textsc{VBench}~\cite{Huang_2024_CVPR}, where perturbation diversity drives evaluation coverage. For pairwise comparison, \textsc{PRISM} provides 600 gold-standard instances, yielding over 4,620 model evaluations across all prompt configurations.

\paragraph{\textbf{\textsc{MIRAGE}}}
\label{subsec:mirage}
The \textit{Model-generated Image sequences with Real And hallucinated Generated Examples} benchmark provides a realistic evaluation setting grounded in actual generative model outputs. Positives are human-validated, while negatives are produced by seven distinct pipelines (\textit{e.g.}, \textsc{GILL}, \textsc{StoryDiffusion}, \textsc{BeamDiffusion}~\cite{fernandes2025latentbeamdiffusionmodels, zhou2024storydiffusionconsistentselfattentionlongrange, liu2025onepromptonestoryfreelunchconsistenttexttoimage, menon2024generatingillustratedinstructions}) across FLUX.1 and SD\,2.1 backbones. This ensures that judge limitations identified on \textsc{PRISM} are validated against authentic generative failure modes~\cite{parthasarathy2025makesgoodgeneratedimage}. \textsc{MIRAGE} spans two domains: \textbf{\textsc{MIRAGE}-Recipes} (procedural) and \textbf{\textsc{MIRAGE}-Vist} (narrative), totaling 672 pairwise comparisons.

\begin{figure}[h] 
  \centering
  \includegraphics[width=\columnwidth]{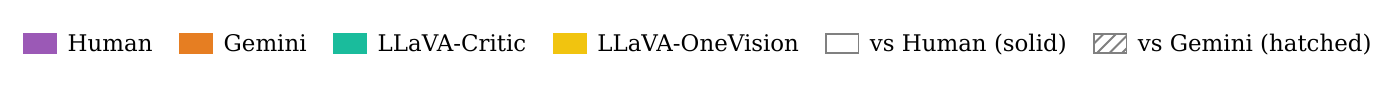}

  \Description{Two side-by-side plots: the left panel shows score distributions; the right panel shows alignment.}

  \begin{subfigure}{0.48\columnwidth}
    \centering
    \includegraphics[width=\linewidth]{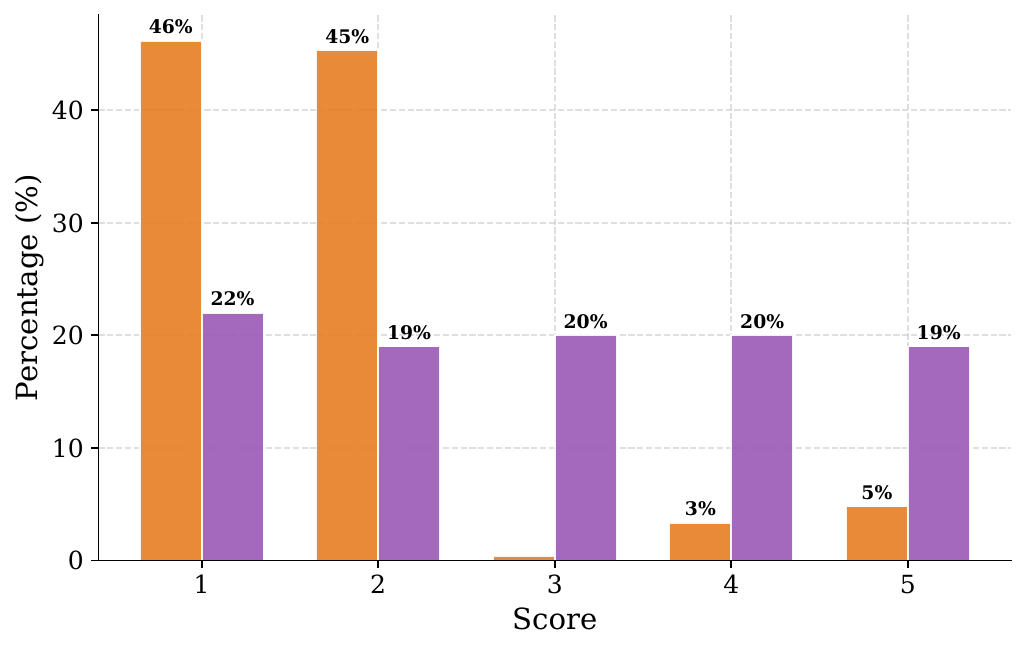}
    \caption{Score dist.}
    \label{fig:gemini_dist}
  \end{subfigure}
  \hfill 
  \begin{subfigure}{0.48\columnwidth}
    \centering
    \includegraphics[width=\linewidth]{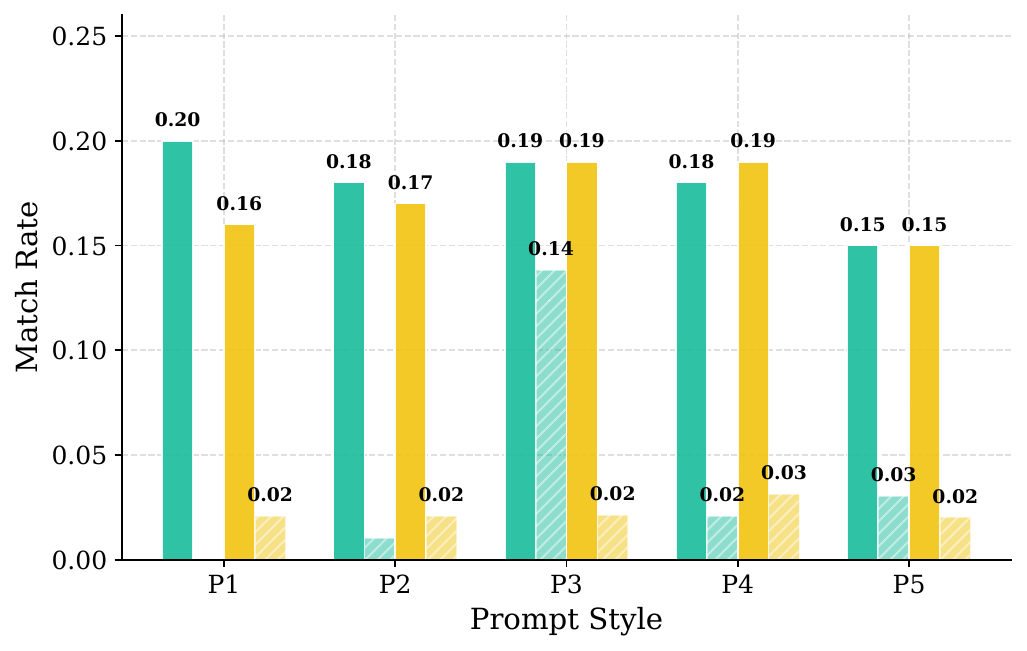}
    \caption{Alignment.}
    \label{fig:match_rate}
  \end{subfigure}
  \caption{\textbf{Automatic judge analysis and calibration.} 
    (\textit{a}) \textsc{Gemini} exhibits a severe \textbf{binary collapse} compared to humans. 
    (\textit{b}) Models demonstrate higher alignment with \textbf{human consensus} than with \textsc{Gemini}.}
  \label{fig:gemini_human_dist}
\end{figure}

\begin{figure*}[t]
  \centering
  \Description{Visual sequence comparison between MIRAGE and PRISM benchmarks.}
  \newcommand{\seqimg}[2]{%
    \setlength{\fboxsep}{0pt}\setlength{\fboxrule}{1.5pt}%
    \fcolorbox{#1}{white}{%
      \includegraphics[width=\linewidth, height=2.0cm, keepaspectratio=false]{#2}}}

  \newcommand{\dname}[1]{%
    \vspace{4pt} 
    \noindent\raggedright{\small\scshape\bfseries #1}\par
    \vspace{-4pt} 
  }

  \newcommand{\pairanno}[6]{%
    \begin{minipage}[t]{0.70\textwidth}
      \vspace{0pt}
      \seqimg{#1}{#2}\par
      \vspace{0pt} 
      \seqimg{#3}{#4}
    \end{minipage}%
    \hfill%
    \begin{minipage}[t]{0.25\textwidth}
      \vspace{0pt}
      \vspace{#6} 
      \raggedright\small\itshape #5
    \end{minipage}}

  
  \dname{PRISM-Semantic}
  \pairanno{cok}{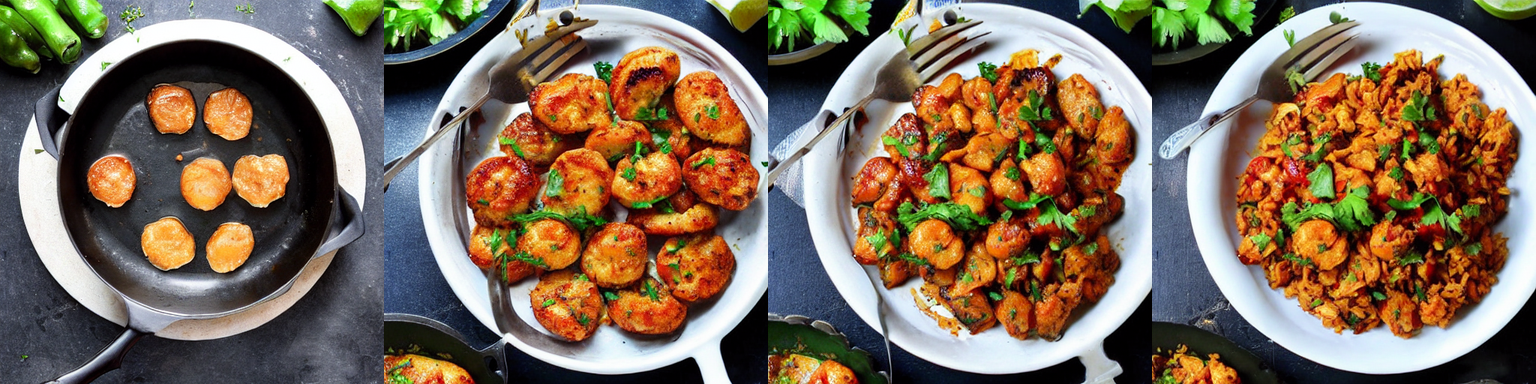}%
           {cerr}{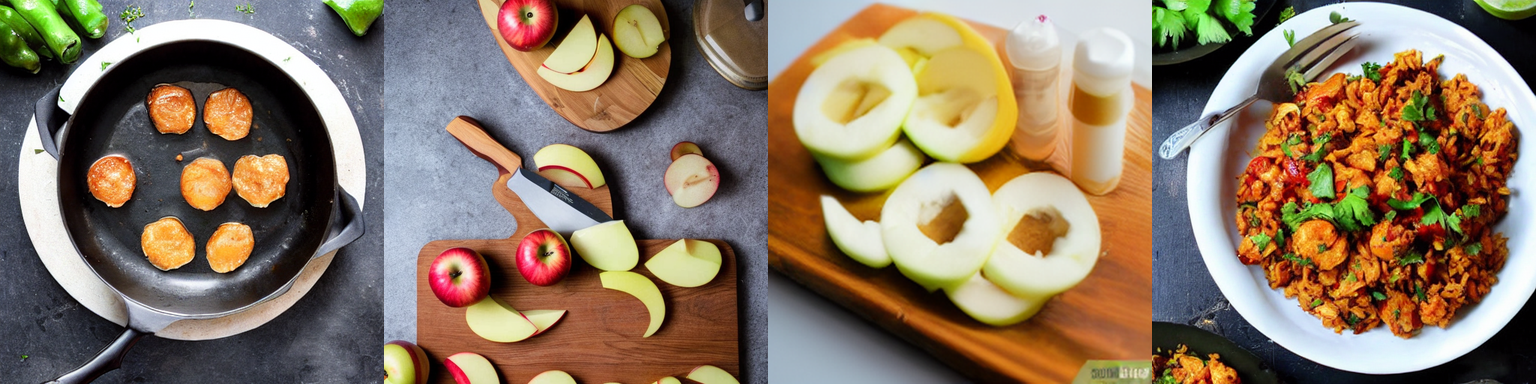}%
           {\textbf{LVLM-judge explanation:}\\Frames 2--3 are replaced with images from an unrelated recipe, breaking semantic coherence.}
           {2.5cm} 

  \dname{PRISM-Temporal}
  \pairanno{cok}{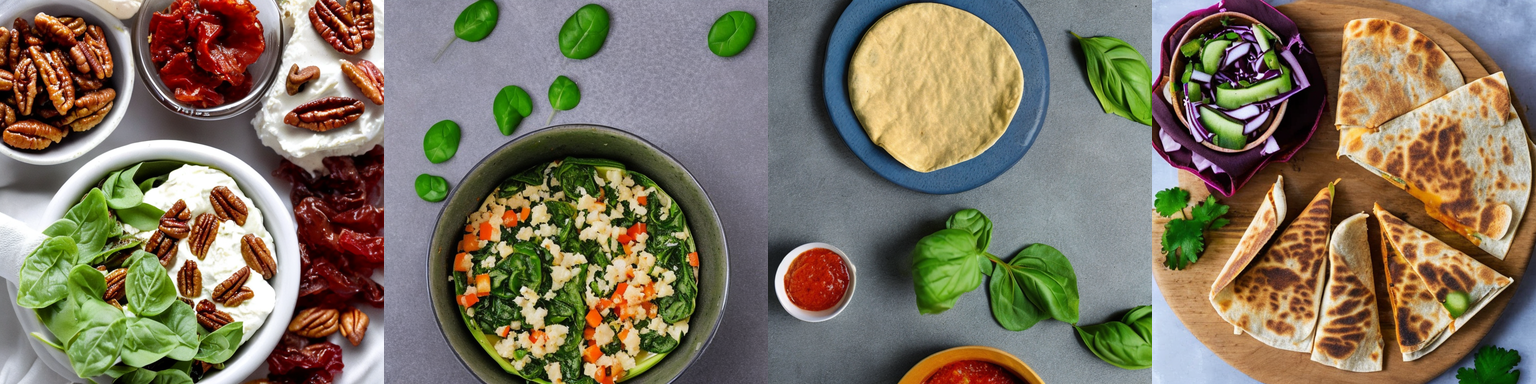}%
           {cerr}{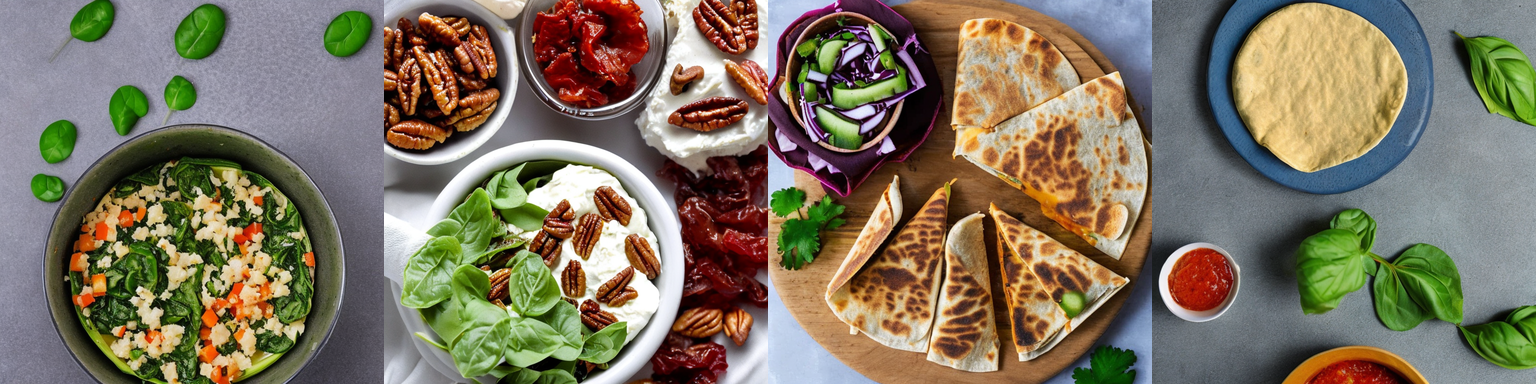}%
           {\textbf{LVLM-judge explanation:}\\Two consecutive steps are swapped, inverting the causal order of the procedure.}
           {2.5cm} 

  \dname{MIRAGE-Vist}
  \pairanno{cok}{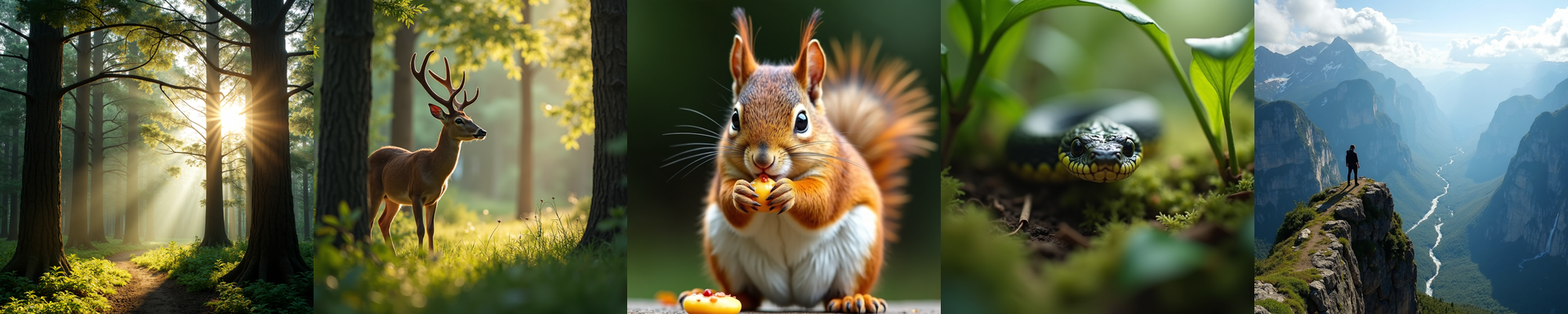}%
           {cerr}{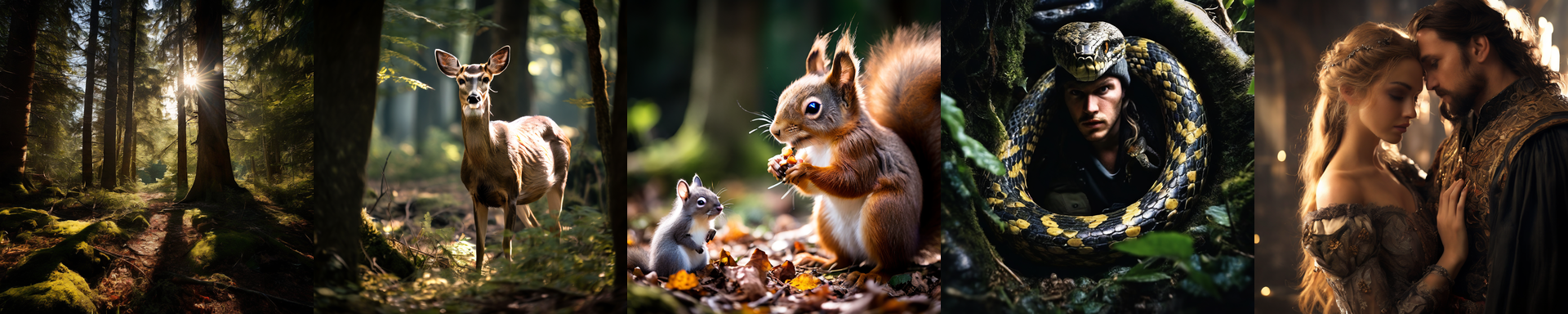}%
           {\textbf{LVLM-judge explanation:}\\Hallucinated frames introduce inconsistent characters and scenes, disrupting narrative continuity.}
           {2.5cm}

  \dname{MIRAGE-Recipes}
  \pairanno{cok}{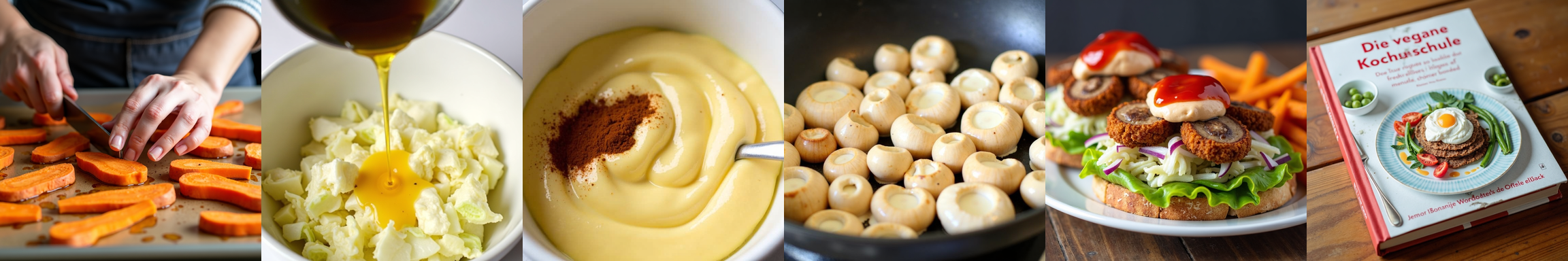}%
           {cerr}{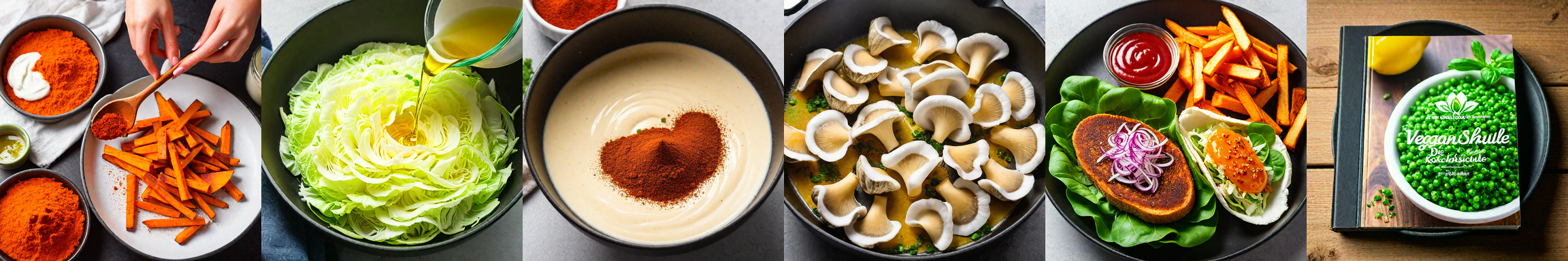}%
           {\textbf{LVLM-judge explanation:}\\Generated frames depict the wrong ingredients at the wrong procedural step.}
           {2.5cm}

   \caption{\textbf{Visual sequence comparisons across benchmarks.}
    \textcolor{cok}{\textbf{Green border}} indicates the correct sequence;
    \textcolor{cerr}{\textbf{Red border}} indicates reordered or hallucinated sequences.
    \textit{\textsc{MIRAGE}} features real-world generative failures.
    \textit{\textsc{PRISM}} utilizes controlled perturbations:
    \textsc{PRISM}-Semantic introduces out-of-context distractors, while
    \textsc{PRISM}-Temporal swaps frames to test ordering sensitivity.}
  \label{fig:all_sequences}
\end{figure*}

\paragraph{Temporal-Visual CoT Explanations}
Both PRISM perturbations and MIRAGE model outputs are augmented with CoT rationales automatically generated by \textsc{Gemini-2.5-Flash}. These explanations serve as a supervision signal for Supervised Fine-Tuning (SFT) and as a reference for assessing reasoning quality. The observed distributional skew in \textsc{Gemini} scores (91\% in $\{1,2\}$) is treated as a characteristic \textbf{binary collapse} of automated judges~\cite{adewumi2024fairnessbiasmultimodalai}, as visually quantified in Figure~\ref{fig:gemini_human_dist} (left). All splits follow a 70/20/10 ratio.

\section{Experiments}
\label{sec:setup}
The experimental framework is designed to evaluate the \textit{Temporal-visual judgment capabilities} of state-of-the-art LVLMs. We move beyond simple classification to probe the performance dichotomy between a model's ability to assign a score and its ability to reason through temporal contradictions. Each instance is systematically perturbed via temporal shuffling to create a rigorous blindness test for the models.

\subsection{Implementation Details}
\label{sec:sft}

\vspace{2mm}
\noindent\textbf{Model Details.}
\label{sec:models}
We evaluate two 7B-class LVLMs as primary judge baselines.
\textbf{\textsc{LLaVA-OneVision}}~\cite{li2024llavaonevision} serves 
as our general-purpose base, utilizing a \texttt{SigLIP-SO400M} vision 
encoder with sliding window attention, natively supporting the 
interleaved multi-image reasoning regime our task requires.
\textbf{\textsc{LLaVA-Critic}}~\cite{xiong2025llavacriticlearningevaluatemultimodal} 
shares the same architecture but is fine-tuned on single-image 
evaluation data; we assess whether this generalizes to composite 
multi-image inputs. 
We refer to our fine-tuned variants as \textbf{LLaVA-OneVision-SFT} (hereafter \textbf{OneVision$^{\dagger}$} , single-prompt) and \textbf{LLaVA-OneVision-SFT+} (hereafter \textbf{OneVision$^\ddagger$} , multi-prompt).

\vspace{2mm}
\noindent\textbf{Training Data.}
We use \textsc{Gemini-2.5-Flash} 
as a reasoning oracle to generate CoT rationales for supervision.
We characterize this skew in Figure~\ref{fig:gemini_human_dist}. For pairwise tasks, a unified 4,620-sample dataset was constructed, where \textit{MIRAGE} samples exhibit higher explanation complexity (2,227 chars) than \textsc{PRISM} (2,022 chars).
Following~\cite{itzhak-etal-2024-instructed, tjuatja-etal-2024-llms}, we use SFT not 
only to improve performance but to \textit{quantify} whether adaptation consolidates 
or mitigates inherent \textbf{primacy/recency biases}, a diagnostic perspective often 
missing in current MLLM-as-a-judge literature~\cite{Qi_2025_CVPR}.

\vspace{2mm}
\noindent\textbf{Optimization.}
To investigate whether structural biases can be mitigated through targeted training, 
adaptation is performed via LoRA ~\cite{hu2021loralowrankadaptationlarge}, 
with rank $r=128$ and $\alpha=256$ to ensure parameter efficiency while maximizing 
reasoning capacity. We employ \textsc{Optuna}~\cite{10.1145/3292500.3330701} for 
hyperparameter optimization minimizing validation loss.
Models are trained for 1 epoch with a global batch size of 128, a learning rate of 
$2\times10^{-5}$, and a cosine learning rate scheduler. We employ the AdamW optimizer 
with a weight decay of 0.1.
Training and inference are conducted using DeepSpeed ZeRO-3 on 
2$\times$H200 GPUs in \texttt{bfloat16}.

\vspace{2mm}
\noindent\textbf{Evaluation Metrics.}
\label{sec:judging_metrics}
Our evaluation protocol is dual-faceted to capture the nuance of the identified 
performance dichotomy~\cite{zheng2023judgingllmasajudgemtbenchchatbot, chen2024mllm}.
\begin{itemize}
    \item \noindent\textit{Pairwise Accuracy.}
This serves as our primary metric for reasoning. Models must select the logically 
coherent sequence over a temporally permuted version. We report macro-averaged 
\textbf{Accuracy, Precision, Recall, and $F_1$-score} to evaluate the model's 
discriminative power between competing sequences, determining whether a model performs 
better than random chance (50\%).
    \item \noindent\textit{Pointwise Correlation.}
For ratings on a 1--5 scale, we measure human alignment via \textbf{Pearson 
correlation ($r$)}, computed against \textsc{Gemini} ($r_G$) and human raters 
($r_H$). Error magnitudes are captured via \textbf{MAE} and \textbf{RMSE}; the latter 
specifically penalizes large outliers to ensure a robust assessment of scoring 
consistency.
    \item \noindent\textit{Explanation Quality.}
Rationales are evaluated against \textsc{Gemini} references via \textbf{METEOR} and 
\textbf{Cosine Similarity} over \textsf{all-MiniLM-L6-v2} embeddings, computed on 
valid non-empty explanations only.
\end{itemize}

\vspace{2mm}
\noindent\textbf{User Study.}
\label{sec:human}
To establish a human gold standard, sequences were manually 
annotated by ranking logical flow and temporal consistency via 
majority voting. This confirms that our perturbations are 
perceptible to human observers, validating model failures as 
genuine structural blindness rather than annotation noise.


\begin{table*}[t]
  \centering
\caption{\small \textbf{Pointwise scoring on \textsc{PRISM}.} 7B models vs.\
\textsc{Gemini} (G) and Human (H) references ($n{=}100$). To save space, only the
two best-performing prompt variants per model are shown; full evaluations and
averages are reported in the extended version. \textbf{Bold}: best value per column.
$r_{G/H}$: Pearson corr.; $F_1$: macro score; MAE/RMS: residuals; CS: Cosine
Similarity. \textbf{OneVision$^{\dagger}$}: single-prompt LoRA SFT on
\textsc{P}$_{\text{w\_R\_P5}}$.}
  \label{tab:pointwise_prism}
  
  \setlength{\tabcolsep}{0pt} 
  \small 
  \renewcommand{\arraystretch}{1.1}

  \begin{tabular*}{\textwidth}{@{\extracolsep{\fill}} llc rrrr rrrr r @{}}
    \toprule
    & & & \multicolumn{4}{c}{\textbf{vs.\ Gemini}} & \multicolumn{4}{c}{\textbf{vs.\ Human}} & \\
    \cmidrule(lr){5-8}\cmidrule(lr){9-12}
    \textbf{Model} & \textbf{SFT} & \textbf{Expl}
      & $\mathbf{F_1}$\,$\uparrow$
      & $\mathbf{r_G}$\,$\uparrow$
      & \textbf{MAE}\,$\downarrow$
      & \textbf{RMS}\,$\downarrow$
      & $\mathbf{F_1}$\,$\uparrow$
      & $\mathbf{r_H}$\,$\uparrow$
      & \textbf{MAE}\,$\downarrow$
      & \textbf{RMS}\,$\downarrow$
      & \textbf{CS}\,$\uparrow$ \\
    \midrule

    \multirow{2}{*}{Critic}
      & 0-shot & no
        & \textbf{0.24} & \textbf{0.48} & \textbf{0.93} & \textbf{1.22}
        & \textbf{0.20} & 0.42 & 1.19 & 1.54
        & -- \\
      & 0-shot & yes
        & 0.03 & 0.29 & 1.74 & 1.86
        & 0.15 & \textbf{0.46} & \textbf{1.10} & \textbf{1.35}
        & 0.72 \\
    \midrule
    \multirow{2}{*}{OneVision}
      & 0-shot & no 
        & 0.04 & 0.41 & 2.32 & 2.47
        & 0.18 & 0.28 & 1.33 & 1.71
        & -- \\
      & 0-shot & yes
        & 0.02 & 0.31 & 2.21 & 2.34
        & 0.15 & 0.28 & 1.30 & 1.61
        & 0.72 \\
    \midrule
    \multirow{2}{*}{OneVision$^{\dagger}$} 
      & \textsc{P}$_{\text{w\_R\_P5}}$ & yes
        & 0.02 & 0.32 & 1.96 & 2.08
        & 0.12 & 0.29 & 1.22 & 1.48
        & 0.72 \\
      & \textsc{P}$_{\text{w\_R\_P5}}$ & yes
        & 0.01 & 0.17 & 2.06 & 2.17
        & 0.17 & 0.34 & 1.24 & 1.56
        & 0.72 \\

    \bottomrule
  \end{tabular*}
\end{table*}

\begin{table*}[t]
\centering
\caption{\textbf{Pairwise evaluation results.} Comparison of judge performance across
\textsc{Mirage} and \textsc{Prism}. \textbf{Bold}: best value per column within each
benchmark. Metric groups follow the definitions in Sec.~\ref{sec:judging_metrics}.
\textbf{OneVision$^{\dagger}$}: single-prompt LoRA SFT;
\textbf{OneVision$^\ddagger$}: multi-prompt LoRA SFT. Prompt details in the extended
version.}
\label{tab:pairwise_final}

\setlength{\tabcolsep}{0pt} 
\small 

\begin{tabular*}{\textwidth}{@{\extracolsep{\fill}} llcc cccc cc @{}}
\toprule
& & & & \multicolumn{4}{c}{\textbf{Pair Comparison}} & \multicolumn{2}{c}{\textbf{Explanation Quality}} \\
\cmidrule(lr){5-8} \cmidrule(lr){9-10}
\textbf{Benchmark} & \textbf{Model} & \textbf{SFT} & \textbf{Expl} & \textbf{Acc} $\uparrow$ & \textbf{Pr} $\uparrow$ & \textbf{Rec} $\uparrow$ & $\mathbf{F_1}$ $\uparrow$ & \textbf{METEOR} $\uparrow$ & \textbf{CS} $\uparrow$ \\
\midrule

\multirow{6}{*}{\shortstack[l]{\textbf{\textsc{Mirage}} \\ \small (Real-world) \\ \small $n=672$}} 
  & \multirow{2}{*}{Critic \cite{xiong2025llavacriticlearningevaluatemultimodal}} 
    & 0-shot & no & 0.80 & 0.60 & 0.55 & 0.56 & -- & -- \\
  & & 0-shot & yes & \textbf{0.94} & \textbf{0.63} & \textbf{0.62} & \textbf{0.62} & \textbf{0.18} & \textbf{0.80} \\
  \cmidrule(lr){2-10}
  & \multirow{2}{*}{OneVision \cite{li2024llavaonevision}} 
    & 0-shot & no & 0.83 & 0.57 & 0.57 & 0.56 & -- & -- \\
  & & 0-shot & yes & 0.80 & 0.55 & 0.54 & 0.53 & 0.14 & 0.77 \\
  \cmidrule(lr){2-10}
  & \multirow{2}{*}{OneVision$^{\dagger}$}  
    & M$_{\text{P6}}$ & no & 0.90 & 0.60 & 0.60 & 0.60 & -- & -- \\
  & & M$_{\text{w\_R\_P6}}$ & yes & 0.84 & 0.57 & 0.57 & 0.56 & 0.15 & 0.78 \\

\midrule

\multirow{7}{*}{\shortstack[l]{\textbf{\textsc{Prism}} \\ \small (Synthetic) \\ \small $n=4,620$}} 
  & \multirow{2}{*}{Critic \cite{xiong2025llavacriticlearningevaluatemultimodal}} 
    & 0-shot & no & 0.43 & \textbf{0.50} & 0.29 & 0.36 & -- & -- \\
  & & 0-shot & yes & 0.63 & 0.47 & 0.42 & 0.44 & \textbf{0.20} & \textbf{0.81} \\
  \cmidrule(lr){2-10}
  & \multirow{2}{*}{OneVision \cite{li2024llavaonevision}} 
    & 0-shot & no & 0.64 & 0.44 & 0.43 & 0.42 & -- & -- \\
  & & 0-shot & yes & 0.65 & 0.46 & 0.43 & 0.42 & 0.17 & 0.80 \\
  \cmidrule(lr){2-10}
  & OneVision$^\dagger$ & P$_{\text{w\_R\_P6}}$ & yes & 0.67 & 0.49 & 0.45 & 0.43 & 0.17 & 0.80 \\
  \cmidrule(lr){2-10}
  & \multirow{2}{*}{OneVision$^\ddagger$} 
    & P$_{\text{w\_R\_P}\star}$ & no & 0.70 & 0.50 & 0.47 & 0.46 & -- & -- \\
  & & P$_{\text{w\_R\_P}\star}$ & yes & \textbf{0.72} & \textbf{0.51} & \textbf{0.48} & \textbf{0.47} & 0.16 & \textbf{0.81} \\
\bottomrule
\end{tabular*}
\end{table*}

\subsection{Main Results}
\label{sec:results_main}

\noindent\textbf{The Scoring Illusion.}
Table~\ref{tab:pointwise_prism} shows the two judges sitting on opposite sides of the
reference distributions. \textsc{LLaVA-Critic} tracks the \textsc{Gemini} oracle most
closely but diverges from human raters, while \textsc{LLaVA-OneVision} shows the
reverse pattern across every prompt. This is a property of the references rather than
evidence of better judgment: \textsc{Gemini} concentrates 91\% of its scores in the
$\{1,2\}$ range (Figure~\ref{fig:gemini_human_dist}), a \textbf{pessimistic collapse}
that any mid-range judge will appear to miss while sitting closer to the near-uniform
human distribution by construction. Task-specific synthetic supervision can thus
induce a skew opposite to the high-score bias usually reported for MLLM
judges~\cite{chen2024mllm}, and agreement with either reference alone remains a weak
proxy for discriminative ability.

\vspace{2mm}
\noindent\textbf{Human Agreement as a Performance Ceiling.}
Human inter-rater correlation on this task is low ($r_H{=}0.34$), reflecting the
intrinsic subjectivity of narrative coherence assessment. Judges can nonetheless
exceed this figure, as \textsc{LLaVA-Critic} does at P5: models are scored against the
majority-vote consensus, which is less noisy than any individual annotator, so
inter-rater agreement bounds the reliability of the labels rather than the attainable
performance. Figure~\ref{fig:gemini_human_dist} shows that both judges align more
closely with human consensus than with the \textsc{Gemini} oracle, though the absolute
match rates remain low throughout. The \textit{Analyze-then-Judge} paradigm moderately
improves alignment for both models; \textsc{LLaVA-OneVision} produces rationales
closer to the reference explanations while lagging in scoring correlation, indicating
a decoupling between explanation fluency and metric calibration.

\vspace{2mm}
\noindent\textbf{Explanation Quality and the Cost of Rationale Supervision.}
Across Tables~\ref{tab:pointwise_prism} and~\ref{tab:pairwise_final}, cosine
similarity remains consistently high while METEOR stays markedly lower, indicating
that generated rationales are semantically aligned with the \textsc{Gemini}
references at the embedding level yet diverge substantially in surface form.
Critically, fine-tuning \emph{with} \textsc{Gemini} rationales leaves explanation
quality essentially unchanged: on both benchmarks and under both protocols,
OneVision$^\dagger$ matches rather than exceeds its zero-shot counterpart. Rationale
supervision therefore improves scoring calibration without inducing richer
explanations, reinforcing the view that \textit{verbal reasoning quality and judgment
reliability are orthogonal axes} in current LVLM judge architectures.

\vspace{2mm}
\noindent\textbf{SFT and Calibration Gains.}
Fine-tuning 
\textsc{LLaVA-OneVision} (OneVision
$^\dagger$) yields consistent gains across 
all Gemini-referenced metrics, reducing $\text{MAE}_G$ from $2.21$ to $1.96$. Against human judgments, OneVision$^\dagger$ reaches $r_H{=}0.34$, perfectly recovering the human 
inter-rater ceiling while narrowing the gap with zero-shot \textsc{Critic}. Furthermore, 
SFT stabilizes generation: the variance in explanation length decreases significantly, 
reflecting more controlled outputs compared to zero-shot counterparts.

\vspace{2mm}
\noindent\textbf{The Accuracy Collapse.}
On \textsc{MIRAGE}, P6 emerges as the optimal prompt for both models (Table~\ref{tab:pairwise_final}). 
\textsc{LLaVA-Critic} and \textsc{LLaVA-OneVision} reach comparable $F_1$ scores 
($0.56$), despite a massive response-length disparity ($825$ vs.\ $19$ tokens). This 
confirms a localized \textbf{verbosity bias}~\cite{chen2024mllm}, where length does 
not confer a discriminative advantage on procedural tasks. Conversely, on the harder 
synthetic perturbations of \textsc{PRISM}, \textsc{LLaVA-OneVision} takes a clear lead 
(Acc $0.64$ vs.\ $0.43$), indicating superior sensitivity to controlled visual 
inconsistencies, yet both remain far below human-level discrimination.

\vspace{2mm}
\noindent\textbf{Effect of Structured Reasoning.}
The adoption of an \textit{Analyze-then-Judge} approach benefits \textsc{LLaVA-Critic} 
more than \textsc{OneVision}: \textsc{Critic} at P6 achieves the highest zero-shot 
accuracy overall ($0.94$ on \textsc{MIRAGE}). On \textsc{PRISM}, the inclusion of 
explanations improves \textsc{Critic}'s $F_1$ from $0.36$ to $0.44$. While reasoning 
quality remains moderate, the process introduces a throughput cost, highlighting a 
trade-off between reasoning depth and evaluation efficiency.

\vspace{2mm}
\noindent\textbf{SFT Scaling and the Limits of Mitigation.}
SFT yields the most pronounced gains in the pairwise setting, see Table~\ref{tab:pairwise_final}. The 
multi-prompt variant (OneVision$^\ddagger$) outperforms single-prompt training by $+0.03$ 
$F_1$, achieving the benchmark peak (Acc $0.72$, $F_1$ $0.47$). This confirms that 
exposure to stylistic diversity during training prevents over-specialization. However, 
fine-tuning does not fully close the gap with human performance, suggesting that the 
underlying architectural biases, primacy and recency effects rooted in causal masking 
and positional embeddings~\cite{wu2025emergencepositionbiastransformers, 
su2023roformerenhancedtransformerrotary}, remain a fundamental bottleneck that 
requires new model designs rather than simply larger datasets.

\noindent\textbf{Structural Blindness: Primacy and Recency Effects.} Beyond aggregate metrics, our diagnostic probes reveal that model failures are  rooted in positional asymmetries inherent to transformer  architectures~\cite{liu2023lostmiddlelanguagemodels, guo2025serial, zheng2024largelanguagemodelsrobust}. Many models show disproportionate reliance on  the first frame to determine overall quality, ignoring subsequent logical contradictions (\textbf{primacy bias}~\cite{wu2025emergencepositionbiastransformers}); conversely, middle-sequence information is often invisible to the judge, (\textbf{recency bias} and \textit{lost-in-the-middle}  effect~\cite{liu2023lostmiddlelanguagemodels}), artifacts of causal masking and RoPE~\cite{su2023roformerenhancedtransformerrotary} not optimized for the bidirectional logic required by interleaved storytelling~\cite{shi-etal-2025-judging, itzhak-etal-2024-instructed}. Critically, these asymmetries persist even after fine-tuning, confirming that SFT consolidates rather than mitigates structural bias~\cite{tjuatja-etal-2024-llms}.

\subsection{Ablation Study}
\label{sec:ablation}

\begin{table}[t]
  \centering
  \caption{\textbf{Per-subset ablation analysis (LLaVA-OneVision judge, eval P6).}
  Breakdown across domains: Recipes and Vist (\textsc{Mirage}); Semantic and Temporal
  (\textsc{Prism}). \textbf{Bold}: best value per column within each subset.
  \textbf{OneVision$^{\dagger}$}: single-prompt LoRA FT;
  \textbf{OneVision$^\ddagger$}: multi-prompt LoRA FT; \textit{w\_R}: trained with
  \textsc{Gemini} rationales.}
  \label{tab:ablation_summary}
  \vspace{-2pt}
  \setlength{\tabcolsep}{1.2pt}
  \fontsize{7.0}{8.5}\selectfont
  \renewcommand{\arraystretch}{1.1}

  \begin{tabular}{@{}lllcc cccc cc@{}}
    \toprule
    & & & & & \multicolumn{4}{c}{\textbf{Pair Comparison}} & \multicolumn{2}{c}{\textbf{Expl. Qual.}} \\
    \cmidrule(lr){6-9} \cmidrule(lr){10-11}
    \textbf{Benchmark} & \textbf{Subset} & \textbf{Model} & \textbf{Conf.} & \textbf{Exp.} & \textbf{Acc} $\uparrow$ & \textbf{Pr} $\uparrow$ & \textbf{Rec} $\uparrow$ & $\mathbf{F_1}$ $\uparrow$ & \textbf{MET.} $\uparrow$ & \textbf{CS} $\uparrow$ \\
    \midrule
    \multirow{8}{*}{\shortstack[l]{\textbf{\textsc{Mirage}} \\ \tiny $n=672$}}
      & \multirow{4}{*}{\shortstack[l]{\textsc{Recipes} \\ \tiny (Procedural)}}
        & OneVision$^\dagger$ & {\tiny M$_{\text{P0}}$}       & \no  & .85 & .56 & .56 & .56 & -- & -- \\
      & & OneVision$^\dagger$ & {\tiny M$_{\text{P6}}$}       & \no  & \textbf{.87} & \textbf{.57} & \textbf{.57} & \textbf{.57} & -- & -- \\
      & & OneVision$^\dagger$ & {\tiny M$_{\text{w\_R\_P0}}$} & \yes & .78 & .53 & .53 & .52 & \textbf{.16} & \textbf{.80} \\
      & & OneVision$^\dagger$ & {\tiny M$_{\text{w\_R\_P6}}$} & \yes & .78 & .53 & .53 & .52 & .14 & .78 \\
      \cmidrule(lr){2-11}
      & \multirow{4}{*}{\shortstack[l]{\textsc{Vist} \\ \tiny (Narrative)}}
        & OneVision$^\dagger$ & {\tiny M$_{\text{P0}}$}       & \no  & \textbf{.92} & \textbf{.62} & \textbf{.61} & \textbf{.61} & -- & -- \\
      & & OneVision$^\dagger$ & {\tiny M$_{\text{P6}}$}       & \no  & \textbf{.92} & \textbf{.62} & \textbf{.61} & \textbf{.61} & -- & -- \\
      & & OneVision$^\dagger$ & {\tiny M$_{\text{w\_R\_P0}}$} & \yes & .90 & .60 & .60 & .60 & \textbf{.15} & \textbf{.77} \\
      & & OneVision$^\dagger$ & {\tiny M$_{\text{w\_R\_P6}}$} & \yes & .90 & .60 & .60 & .60 & .13 & .76 \\
    \midrule
    \multirow{7}{*}{\shortstack[l]{\textbf{\textsc{Prism}} \\ \tiny $n = 4{,}620$}}
      & \multirow{4}{*}{\shortstack[l]{\textsc{Semantic} \\ \tiny (Coherence)}}
        & OneVision$^\dagger$  & {\tiny P$_{\text{P6}}$}            & \no  & .75 & .52 & .50 & .50 & -- & -- \\
      & & OneVision$^\dagger$  & {\tiny P$_{\text{w\_R\_P6}}$}      & \yes & .75 & .52 & .50 & .50 & \textbf{.17} & .80 \\
      & & OneVision$^\ddagger$ & {\tiny P$_{\text{P}\star}$}        & \no  & \textbf{.82} & \textbf{.56} & \textbf{.55} & \textbf{.55} & -- & -- \\
      & & OneVision$^\ddagger$ & {\tiny P$_{\text{w\_R\_P}\star}$}  & \yes & .75 & .52 & .50 & .50 & \textbf{.17} & \textbf{.81} \\
      \cmidrule(lr){2-11}
      & \multirow{3}{*}{\shortstack[l]{\textsc{Temporal} \\ \tiny (Logic)}}
        & OneVision$^\dagger$  & {\tiny P$_{\text{w\_R\_P6}}$}      & \yes & .59 & .49 & .40 & .35 & .17 & .81 \\
      & & OneVision$^\ddagger$ & {\tiny P$_{\text{P}\star}$}        & \no  & \textbf{.62} & .47 & \textbf{.42} & \textbf{.39} & -- & -- \\
      & & OneVision$^\ddagger$ & {\tiny P$_{\text{w\_R\_P}\star}$}  & \yes & \textbf{.62} & \textbf{.50} & \textbf{.42} & .38 & \textbf{.17} & \textbf{.81} \\
    \bottomrule
  \end{tabular}
\end{table}

\begin{table}[t]
  \centering
  \caption{\textbf{Positional sensitivity on \textsc{Prism} ($n{=}4,620$).} Accuracy
  and macro $F_1$ per position. $W_{\text{avg}}$ denotes sample-weighted mean.
  \textbf{Bold}: best value per column. The vertical line separates Semantic from
  Temporal subsets. \textbf{OneVision$^{\dagger}$}: single-prompt LoRA SFT;
  \textbf{OneVision$^\ddagger$}: multi-prompt LoRA SFT. \textit{Full per-position
  results in the extended version.}}
  \label{tab:positional_main}

  \setlength{\tabcolsep}{2.8pt}
  \footnotesize
  \renewcommand{\arraystretch}{1.2}

  \begin{tabular}{@{} l cccccc | c @{}}
    \toprule
    & \multicolumn{6}{c}{\textbf{\textsc{Semantic}} (Acc / $F_1$)} & \textbf{\textsc{Temp.}} \\
    \cmidrule(lr){2-7} \cmidrule(lr){8-8}
    \textbf{Model} & \textbf{p0} & \textbf{p1} & \textbf{p2} & \textbf{p3} & \textbf{p4} & \textbf{p5} & $\mathbf{W_{\text{avg}}}$ \\
    \midrule
    OneVision \cite{li2024llavaonevision} & .53/.49 & .52/.49 & .51/.48 & .50/.48 & .49/.45 & \textbf{.59}/\textbf{.52} & \textbf{.49}/\textbf{.46} \\
    Critic \cite{xiong2025llavacriticlearningevaluatemultimodal} & .54/.54 & .47/.49 & .48/.49 & .45/.46 & .48/.48 & .45/.41 & .43/.44 \\
    \midrule
    OneVision$^\dagger$ P$_{\text{P6}}$ & .65/.65 & .52/.52 & .50/.50 & \textbf{.51}/\textbf{.51} & .52/\textbf{.52} & .54/\textbf{.52} & .46/\textbf{.46} \\
    OneVision$^\ddagger$ P$_{\text{P}\star}$ & \textbf{.70}/\textbf{.69} & \textbf{.53}/\textbf{.52} & \textbf{.55}/\textbf{.53} & \textbf{.51}/.50 & \textbf{.53}/\textbf{.52} & .55/.50 & .47/.45 \\
    \bottomrule
  \end{tabular}
\end{table}

\vspace{2mm}
\noindent\textbf{Sensitivity to Evaluation Protocol: Pointwise vs.\ Pairwise.}
We investigated how the choice of evaluation protocol masks or reveals the model's 
structural blindness. In pointwise experiments, models tend to regress toward a 
``safe'' high score, making it impossible to distinguish between a 
coherent narrative and a shuffled one~\cite{chen2024mllm}. When transitioned to a 
pairwise protocol, the same models exhibit a catastrophic performance collapse. This 
ablation confirms that the off-the-shelf use of LVLMs as scoring judges provides a 
false sense of reliability, hiding a fundamental inability to perform logical 
discrimination~\cite{zheng2023judgingllmasajudgemtbenchchatbot}.

\vspace{2mm}
\noindent\textbf{Semantic Robustness vs.\ Temporal Fragility.}
As detailed in Table~\ref{tab:ablation_summary}, per-subset analysis on \textsc{PRISM} 
reveals a stark contrast between perturbation types. While adaptation through 
multi-prompt training (OneVision$^\ddagger$) significantly bolsters \textbf{semantic outlier 
detection} ($F_1{=}0.55$, Acc${=}0.82$), the \textsc{PRISM}-Temporal subset remains 
structurally challenging. All model variants fall below the $F_1{=}0.40$ threshold, 
with SFT providing only marginal gains over zero-shot baselines. The failure 
of the \textit{Analyze-then-Judge} paradigm to shift results on this subset confirms 
that temporal ordering sensitivity is a qualitatively different capability, largely 
decoupled from standard reasoning supervision~\cite{Huang_2024_CVPR}.

\vspace{2mm}
\noindent\textbf{SFT-Induced Shifts in Domain Tractability.}
Fine-tuning on \textsc{MIRAGE} effectively inverts the zero-shot tractability ranking 
of sub-domains (Table~\ref{tab:ablation_summary}). Without adaptation, \textit{MIRAGE-Recipes} appears more tractable; however, following SFT, 
\textit{MIRAGE-Vist} emerges as the superior subset (Acc $0.92$ vs.\ $0.87$). This 
suggests that adaptation is more effective at resolving narrative ambiguity in visual 
storytelling than at overcoming the rigid procedural constraints of recipe-based 
hallucinations~\cite{parthasarathy2025makesgoodgeneratedimage}.

\vspace{2mm}
\noindent\textbf{Positional Bias: Primacy vs.\ Recency Effects.}
Positional sensitivity (Table~\ref{tab:positional_main}, 
Figure~\ref{fig:positional_sensitivity} in the Appendix) reveals a significant first-option bias: models disproportionately select the first sequence regardless of logical consistency.
On \textsc{PRISM}-Semantic, we observe a consistent 
\textbf{primacy effect}~\cite{guo2025serial, 
wu2025emergencepositionbiastransformers}: performance peaks at the initial position 
($pos_0$, Acc${=}0.70$, see Table~\ref{tab:positional_main}) and degrades monotonically toward later positions 
, with anomalies disrupting the initial context proving most 
salient to the judge. Conversely, \textsc{PRISM}-Temporal exhibits a \textbf{recency 
effect}, where swaps at later positions are marginally easier to detect, errors buried 
in the middle of a 5--6 image sequence are frequently invisible to the 
judge~\cite{liu2023lostmiddlelanguagemodels}. However, even the best-tuned variant (OneVision$^\ddagger$) fails to exceed a $W_{\text{avg}}$ of $0.49$, remaining near chance level (Table~\ref{tab:positional_main}, last column).

\vspace{2mm}
\noindent\textbf{Scaling vs.\ Reasoning: Does Size Solve Blindness?}
The persistence of positional asymmetries across both zero-shot and SFT models 
indicates that these biases reflect \textbf{structural attention constraints}, 
rooted in causal masking and RoPE ~\cite{su2023roformerenhancedtransformerrotary} 
not optimized for bidirectional logic, rather than artifacts of supervision 
data~\cite{itzhak-etal-2024-instructed, tjuatja-etal-2024-llms}. 
The persistent gap across all model configurations suggests that \textit{temporal 
illiteracy} is an architectural bottleneck rather than a data or scaling issue, 
reinforcing the necessity of moving toward \textit{Temporally-Aware Evaluation} 
paradigms that explicitly model sequential 
dependencies~\cite{Qi_2025_CVPR, adewumi2024fairnessbiasmultimodalai}.

\section{Conclusion: Beyond the Beauty of the Frame}
\label{sec:conclusion}

We have challenged the prevailing ``pointwise'' evaluation paradigm in generative
multimedia: by treating multi-image sequences as a ``bag of frames,'' the community
prioritises visual fidelity while remaining blind to the temporal failures that
undermine narrative coherence~\cite{huang2024surveyevaluationmultimodallarge,
Huang_2024_CVPR}.

Through \textsc{Prism} and \textsc{Mirage} we show that judges
collapse toward chance when discriminating temporal order, not for lack of data,
but because of primacy and recency biases rooted in architectural
constraints~\cite{wu2025emergencepositionbiastransformers,
su2023roformerenhancedtransformerrotary}. Scale does not resolve the bottleneck:
zero-shot evaluations on \textsc{Gemma4}-31B~\cite{gemmateam2026gemma4technicalreport}
and \textsc{Qwen3.6}-27B~\cite{qwen3.6-27b} reproduce every failure mode observed
at 7B, from severe answer invalidity to the same RoPE-consistent positional
signature.

\paragraph{A Brave New Direction for Generative Evaluation.}
We call for \textbf{Temporally-Aware Evaluation}: metrics that assess the
interleaved logic of a sequence as a whole~\cite{chen2024mllm}, judges that ground
decisions in per-step causal rationales rather than visual similarity, and
architectural redesign rather than data scaling to mitigate positional drift.
The stakes extend beyond benchmarking: ordering clinical scans by disease progression
or reconstructing timelines from archival footage both assume a judge that tracks
temporal structure rather than position.
The era of judging generative AI by its
beauty must give way to judging it by its logic.

\begin{acks}
The authors would like to thank SACMI for supporting this work.  This work has been partially funded NOVA LINCS project Ref. UIDP/04516/2020.

\end{acks}

\bibliographystyle{ACM-Reference-Format}
\bibliography{sample-base}

@String(CVPR  = {IEEE Conf. Comput. Vis. Pattern Recog.})

@String(CVPR  = {CVPR})

@article{li2024llavaonevision,
	title={LLaVA-OneVision: Easy Visual Task Transfer},
	author={Li, Bo and Zhang, Yuanhan and Guo, Dong and Zhang, Renrui and Li, Feng and Zhang, Hao and Zhang, Kaichen and Li, Yanwei and Liu, Ziwei and Li, Chunyuan},
	journal={arXiv preprint arXiv:2408.03326},
	year={2024}
}

@misc{xiong2025llavacriticlearningevaluatemultimodal,
	title={LLaVA-Critic: Learning to Evaluate Multimodal Models}, 
	author={Tianyi Xiong and Xiyao Wang and Dong Guo and Qinghao Ye and Haoqi Fan and Quanquan Gu and Heng Huang and Chunyuan Li},
	year={2025},
	eprint={2410.02712},
	archivePrefix={arXiv},
	primaryClass={cs.CV},
	url={https://arxiv.org/abs/2410.02712}, 
}

@inproceedings{10.1145/3292500.3330701,
	author = {Akiba, Takuya and Sano, Shotaro and Yanase, Toshihiko and Ohta, Takeru and Koyama, Masanori},
	title = {Optuna: A Next-generation Hyperparameter Optimization Framework},
	year = {2019},
	isbn = {9781450362016},
	publisher = {Association for Computing Machinery},
	address = {New York, NY, USA},
	url = {https://doi.org/10.1145/3292500.3330701},
	doi = {10.1145/3292500.3330701},
	booktitle = {Proceedings of the 25th ACM SIGKDD International Conference on Knowledge Discovery \& Data Mining},
	pages = {2623–2631},
	numpages = {9},
	location = {Anchorage, AK, USA},
	series = {KDD '19}
}

@misc{liu2023lostmiddlelanguagemodels,
	title={Lost in the Middle: How Language Models Use Long Contexts}, 
	author={Nelson F. Liu and Kevin Lin and John Hewitt and Ashwin Paranjape and Michele Bevilacqua and Fabio Petroni and Percy Liang},
	year={2023},
	eprint={2307.03172},
	archivePrefix={arXiv},
	primaryClass={cs.CL},
	url={https://arxiv.org/abs/2307.03172}, 
}

@article{lin2024evaluating,
	title={Evaluating Text-to-Visual Generation with Image-to-Text Generation},
	author={Lin, Zhiqiu and Pathak, Deepak and Li, Baiqi and Li, Jiayao and Xia, Xide and Neubig, Graham and Zhang, Pengchuan and Ramanan, Deva},
	journal={arXiv preprint arXiv:2404.01291},
	year={2024}
}

@misc{gordon2025unblockingfinegrainedevaluationdetailed,
	title={Unblocking Fine-Grained Evaluation of Detailed Captions: An Explaining AutoRater and Critic-and-Revise Pipeline}, 
	author={Brian Gordon and Yonatan Bitton and Andreea Marzoca and Yasumasa Onoe and Xiao Wang and Daniel Cohen-Or and Idan Szpektor},
	year={2025},
	eprint={2506.07631},
	archivePrefix={arXiv},
	primaryClass={cs.CL},
	url={https://arxiv.org/abs/2506.07631}, 
}

@misc{fernandes2025latentbeamdiffusionmodels,
	title={Latent Beam Diffusion Models for Decoding Image Sequences}, 
	author={Guilherme Fernandes and Vasco Ramos and Regev Cohen and Idan Szpektor and João Magalhães},
	year={2025},
	eprint={2503.20429},
	archivePrefix={arXiv},
	primaryClass={cs.CV},
	url={https://arxiv.org/abs/2503.20429}, 
}

@misc{jiang2025bridgingmodelingcorrelationspairwise,
      title={Bridging and Modeling Correlations in Pairwise Data for Direct Preference Optimization}, 
      author={Yuxin Jiang and Bo Huang and Yufei Wang and Xingshan Zeng and Liangyou Li and Yasheng Wang and Xin Jiang and Lifeng Shang and Ruiming Tang and Wei Wang},
      year={2025},
      eprint={2408.07471},
      archivePrefix={arXiv},
      primaryClass={cs.CL},
      url={https://arxiv.org/abs/2408.07471}, 
}

@misc{wang2025unifiedmultimodalchainofthoughtreward,
      title={Unified Multimodal Chain-of-Thought Reward Model through Reinforcement Fine-Tuning}, 
      author={Yibin Wang and Zhimin Li and Yuhang Zang and Chunyu Wang and Qinglin Lu and Cheng Jin and Jiaqi Wang},
      year={2025},
      eprint={2505.03318},
      archivePrefix={arXiv},
      primaryClass={cs.CV},
      url={https://arxiv.org/abs/2505.03318}, 
}

@misc{zang2025internlmxcomposer25rewardsimpleeffectivemultimodal,
      title={InternLM-XComposer2.5-Reward: A Simple Yet Effective Multi-Modal Reward Model}, 
      author={Yuhang Zang and Xiaoyi Dong and Pan Zhang and Yuhang Cao and Ziyu Liu and Shengyuan Ding and Shenxi Wu and Yubo Ma and Haodong Duan and Wenwei Zhang and Kai Chen and Dahua Lin and Jiaqi Wang},
      year={2025},
      eprint={2501.12368},
      archivePrefix={arXiv},
      primaryClass={cs.CV},
      url={https://arxiv.org/abs/2501.12368}, 
}

@misc{dai2025captionsrewardscarevlleveraging,
      title={From Captions to Rewards (CAREVL): Leveraging Large Language Model Experts for Enhanced Reward Modeling in Large Vision-Language Models}, 
      author={Muzhi Dai and Jiashuo Sun and Zhiyuan Zhao and Shixuan Liu and Rui Li and Junyu Gao and Xuelong Li},
      year={2025},
      eprint={2503.06260},
      archivePrefix={arXiv},
      primaryClass={cs.CV},
      url={https://arxiv.org/abs/2503.06260}, 
}

@inproceedings{chen2024mllm,
  title={Mllm-as-a-judge: Assessing multimodal llm-as-a-judge with vision-language benchmark},
  author={Chen, Dongping and Chen, Ruoxi and Zhang, Shilin and Wang, Yaochen and Liu, Yinuo and Zhou, Huichi and Zhang, Qihui and Wan, Yao and Zhou, Pan and Sun, Lichao},
  booktitle={Forty-first International Conference on Machine Learning},
  year={2024}
}

@misc{parthasarathy2025makesgoodgeneratedimage,
      title={What Makes a Good Generated Image? Investigating Human and Multimodal LLM Image Preference Alignment}, 
      author={Rishab Parthasarathy and Jasmine Collins and Cory Stephenson},
      year={2025},
      eprint={2509.12750},
      archivePrefix={arXiv},
      primaryClass={cs.CV},
      url={https://arxiv.org/abs/2509.12750}, 
}

@article{yu2025aligning,
  title={Aligning multimodal llm with human preference: A survey},
  author={Yu, Tao and Zhang, Yi-Fan and Fu, Chaoyou and Wu, Junkang and Lu, Jinda and Wang, Kun and Lu, Xingyu and Shen, Yunhang and Zhang, Guibin and Song, Dingjie and others},
  journal={arXiv preprint arXiv:2503.14504},
  year={2025}
}

@InProceedings{Qi_2025_CVPR,
    author    = {Qi, Zelu and Shi, Ping and Zhang, Chaoyang and Wang, Shuqi and Zhao, Fei and Pan, Da and Ying, Zefeng},
    title     = {Towards Holistic Visual Quality Assessment of AI-Generated Videos: A LLM-Based Multi-Dimensional Evaluation Model},
    booktitle = {Proceedings of the IEEE/CVF Conference on Computer Vision and Pattern Recognition (CVPR) Workshops},
    month     = {June},
    year      = {2025},
    pages     = {1493-1502}
}

@misc{zheng2025multimodalspatialreasoninglarge,
      title={Multimodal Spatial Reasoning in the Large Model Era: A Survey and Benchmarks}, 
      author={Xu Zheng and Zihao Dongfang and Lutao Jiang and Boyuan Zheng and Yulong Guo and Zhenquan Zhang and Giuliano Albanese and Runyi Yang and Mengjiao Ma and Zixin Zhang and Chenfei Liao and Dingcheng Zhen and Yuanhuiyi Lyu and Yuqian Fu and Bin Ren and Linfeng Zhang and Danda Pani Paudel and Nicu Sebe and Luc Van Gool and Xuming Hu},
      year={2025},
      eprint={2510.25760},
      archivePrefix={arXiv},
      primaryClass={cs.CV},
      url={https://arxiv.org/abs/2510.25760}, 
}

@ARTICLE{11399654,

  author={Xing, Zheng and Zhao, Weibing},

  journal={IEEE Transactions on Image Processing}, 

  title={Temporal Visual Semantics-Induced Human Motion Understanding With Large Language Models}, 

  year={2026},

  volume={35},

  number={},

  pages={2182-2197},

  doi={10.1109/TIP.2026.3663857}}

@misc{lee2024prometheusvisionvisionlanguagemodeljudge,
      title={Prometheus-Vision: Vision-Language Model as a Judge for Fine-Grained Evaluation}, 
      author={Seongyun Lee and Seungone Kim and Sue Hyun Park and Geewook Kim and Minjoon Seo},
      year={2024},
      eprint={2401.06591},
      archivePrefix={arXiv},
      primaryClass={cs.CL},
      url={https://arxiv.org/abs/2401.06591}, 
}

@misc{li2023generativejudgeevaluatingalignment,
      title={Generative Judge for Evaluating Alignment}, 
      author={Junlong Li and Shichao Sun and Weizhe Yuan and Run-Ze Fan and Hai Zhao and Pengfei Liu},
      year={2023},
      eprint={2310.05470},
      archivePrefix={arXiv},
      primaryClass={cs.CL},
      url={https://arxiv.org/abs/2310.05470}, 
}

@misc{comanici2025gemini25,
      title={Gemini 2.5: Pushing the Frontier with Advanced Reasoning, 
             Multimodality, Long Context, and Next Generation Agentic 
             Capabilities}, 
      author={{Gemini Team}},
      year={2025},
      eprint={2507.06261},
      archivePrefix={arXiv},
      primaryClass={cs.CL},
      url={https://arxiv.org/abs/2507.06261}, 
}

@misc{huang2024surveyevaluationmultimodallarge,
      title={A Survey on Evaluation of Multimodal Large Language Models}, 
      author={Jiaxing Huang and Jingyi Zhang},
      year={2024},
      eprint={2408.15769},
      archivePrefix={arXiv},
      primaryClass={cs.CV},
      url={https://arxiv.org/abs/2408.15769}, 
}

@misc{hu2021loralowrankadaptationlarge,
      title={LoRA: Low-Rank Adaptation of Large Language Models}, 
      author={Edward J. Hu and Yelong Shen and Phillip Wallis and Zeyuan Allen-Zhu and Yuanzhi Li and Shean Wang and Lu Wang and Weizhu Chen},
      year={2021},
      eprint={2106.09685},
      archivePrefix={arXiv},
      primaryClass={cs.CL},
      url={https://arxiv.org/abs/2106.09685}, 
}

@techreport{OpenAI2023GPTV,
  title={GPT-4V(ision) System Card},
  author={{OpenAI}},
  year={2023},
  url={https://cdn.openai.com/papers/GPTV_System_Card.pdf}
}

@misc{wu2025emergencepositionbiastransformers,
      title={On the Emergence of Position Bias in Transformers}, 
      author={Xinyi Wu and Yifei Wang and Stefanie Jegelka and Ali Jadbabaie},
      year={2025},
      eprint={2502.01951},
      archivePrefix={arXiv},
      primaryClass={cs.LG},
      url={https://arxiv.org/abs/2502.01951}, 
}

@misc{lu2025artifactsattentionsinksstructured,
      title={Artifacts and Attention Sinks: Structured Approximations for Efficient Vision Transformers}, 
      author={Andrew Lu and Wentinn Liao and Liuhui Wang and Huzheng Yang and Jianbo Shi},
      year={2025},
      eprint={2507.16018},
      archivePrefix={arXiv},
      primaryClass={cs.CV},
      url={https://arxiv.org/abs/2507.16018}, 
}

@misc{sun2025rethinkingbradleyterrymodelspreferencebased,
      title={Rethinking Bradley-Terry Models in Preference-Based Reward Modeling: Foundations, Theory, and Alternatives}, 
      author={Hao Sun and Yunyi Shen and Jean-Francois Ton},
      year={2025},
      eprint={2411.04991},
      archivePrefix={arXiv},
      primaryClass={cs.AI},
      url={https://arxiv.org/abs/2411.04991}, 
}

@inproceedings{shi-etal-2025-judging,
    title = "Judging the Judges: A Systematic Study of Position Bias in {LLM}-as-a-Judge",
    author = "Shi, Lin  and
      Ma, Chiyu  and
      Liang, Wenhua  and
      Diao, Xingjian  and
      Ma, Weicheng  and
      Vosoughi, Soroush",
    editor = "Inui, Kentaro  and
      Sakti, Sakriani  and
      Wang, Haofen  and
      Wong, Derek F.  and
      Bhattacharyya, Pushpak  and
      Banerjee, Biplab  and
      Ekbal, Asif  and
      Chakraborty, Tanmoy  and
      Singh, Dhirendra Pratap",
    booktitle = "Proceedings of the 14th International Joint Conference on Natural Language Processing and the 4th Conference of the Asia-Pacific Chapter of the Association for Computational Linguistics",
    month = dec,
    year = "2025",
    address = "Mumbai, India",
    publisher = "The Asian Federation of Natural Language Processing and The Association for Computational Linguistics",
    url = "https://aclanthology.org/2025.ijcnlp-long.18/",
    pages = "292--314",
    ISBN = "979-8-89176-298-5"
}

@InProceedings{Chen_2025_CVPR,
    author    = {Chen, Wei and Li, Lin and Yang, Yongqi and Wen, Bin and Yang, Fan and Gao, Tingting and Wu, Yu and Chen, Long},
    title     = {CoMM: A Coherent Interleaved Image-Text Dataset for Multimodal Understanding and Generation},
    booktitle = {Proceedings of the IEEE/CVF Conference on Computer Vision and Pattern Recognition (CVPR)},
    month     = {June},
    year      = {2025},
    pages     = {8073-8082}
}

@InProceedings{Huang_2024_CVPR,
    author    = {Huang, Ziqi and He, Yinan and Yu, Jiashuo and Zhang, Fan and Si, Chenyang and Jiang, Yuming and Zhang, Yuanhan and Wu, Tianxing and Jin, Qingyang and Chanpaisit, Nattapol and Wang, Yaohui and Chen, Xinyuan and Wang, Limin and Lin, Dahua and Qiao, Yu and Liu, Ziwei},
    title     = {VBench: Comprehensive Benchmark Suite for Video Generative Models},
    booktitle = {Proceedings of the IEEE/CVF Conference on Computer Vision and Pattern Recognition (CVPR)},
    month     = {June},
    year      = {2024},
    pages     = {21807-21818}
}

@ARTICLE{11215765,

  author={Dou, Hongkun and Lu, Junzhe and Du, Jinyang and Fu, Chengwei and Yao, Wen and Li, Hongjue and Deng, Yue},

  journal={IEEE Transactions on Artificial Intelligence}, 

  title={Towards a Unified Framework for Consistency Generative Modeling}, 

  year={2025},

  volume={},

  number={},

  pages={1-13},

  doi={10.1109/TAI.2025.3624330}}

@misc{li2025judgesjudgellmjuryondemand,
      title={Who Judges the Judge? LLM Jury-on-Demand: Building Trustworthy LLM Evaluation Systems}, 
      author={Xiaochuan Li and Ke Wang and Girija Gouda and Shubham Choudhary and Yaqun Wang and Linwei Hu and Joel Vaughan and Freddy Lecue},
      year={2025},
      eprint={2512.01786},
      archivePrefix={arXiv},
      primaryClass={cs.AI},
      url={https://arxiv.org/abs/2512.01786}, 
}

@misc{ye2024justiceprejudicequantifyingbiases,
      title={Justice or Prejudice? Quantifying Biases in LLM-as-a-Judge}, 
      author={Jiayi Ye and Yanbo Wang and Yue Huang and Dongping Chen and Qihui Zhang and Nuno Moniz and Tian Gao and Werner Geyer and Chao Huang and Pin-Yu Chen and Nitesh V Chawla and Xiangliang Zhang},
      year={2024},
      eprint={2410.02736},
      archivePrefix={arXiv},
      primaryClass={cs.CL},
      url={https://arxiv.org/abs/2410.02736}, 
}

@misc{adewumi2024fairnessbiasmultimodalai,
      title={Fairness and Bias in Multimodal AI: A Survey}, 
      author={Tosin Adewumi and Lama Alkhaled and Namrata Gurung and Goya van Boven and Irene Pagliai},
      year={2024},
      eprint={2406.19097},
      archivePrefix={arXiv},
      primaryClass={cs.CL},
      url={https://arxiv.org/abs/2406.19097}, 
}

@misc{wang2025eliminatingpositionbiaslanguage,
      title={Eliminating Position Bias of Language Models: A Mechanistic Approach}, 
      author={Ziqi Wang and Hanlin Zhang and Xiner Li and Kuan-Hao Huang and Chi Han and Shuiwang Ji and Sham M. Kakade and Hao Peng and Heng Ji},
      year={2025},
      eprint={2407.01100},
      archivePrefix={arXiv},
      primaryClass={cs.CL},
      url={https://arxiv.org/abs/2407.01100}, 
}

@article{itzhak-etal-2024-instructed,
    title = "Instructed to Bias: Instruction-Tuned Language Models Exhibit Emergent Cognitive Bias",
    author = "Itzhak, Itay  and
      Stanovsky, Gabriel  and
      Rosenfeld, Nir  and
      Belinkov, Yonatan",
    journal = "Transactions of the Association for Computational Linguistics",
    volume = "12",
    year = "2024",
    address = "Cambridge, MA",
    publisher = "MIT Press",
    url = "https://aclanthology.org/2024.tacl-1.43/",
    doi = "10.1162/tacl_a_00673",
    pages = "771--785"
}

@article{tjuatja-etal-2024-llms,
    title = "Do {LLM}s Exhibit Human-like Response Biases? A Case Study in Survey Design",
    author = "Tjuatja, Lindia  and
      Chen, Valerie  and
      Wu, Tongshuang  and
      Talwalkwar, Ameet  and
      Neubig, Graham",
    journal = "Transactions of the Association for Computational Linguistics",
    volume = "12",
    year = "2024",
    address = "Cambridge, MA",
    publisher = "MIT Press",
    url = "https://aclanthology.org/2024.tacl-1.56/",
    doi = "10.1162/tacl_a_00685",
    pages = "1011--1026"
}

@inproceedings{pezeshkpour-hruschka-2024-large,
    title = "Large Language Models Sensitivity to The Order of Options in Multiple-Choice Questions",
    author = "Pezeshkpour, Pouya  and
      Hruschka, Estevam",
    editor = "Duh, Kevin  and
      Gomez, Helena  and
      Bethard, Steven",
    booktitle = "Findings of the Association for Computational Linguistics: NAACL 2024",
    month = jun,
    year = "2024",
    address = "Mexico City, Mexico",
    publisher = "Association for Computational Linguistics",
    url = "https://aclanthology.org/2024.findings-naacl.130/",
    doi = "10.18653/v1/2024.findings-naacl.130",
    pages = "2006--2017"
}

@misc{zheng2024largelanguagemodelsrobust,
      title={Large Language Models Are Not Robust Multiple Choice Selectors}, 
      author={Chujie Zheng and Hao Zhou and Fandong Meng and Jie Zhou and Minlie Huang},
      year={2024},
      eprint={2309.03882},
      archivePrefix={arXiv},
      primaryClass={cs.CL},
      url={https://arxiv.org/abs/2309.03882}, 
}

@misc{drissi2024lesssimulationbasedapproachdynamic,
      title={More is Less? A Simulation-Based Approach to Dynamic Interactions between Biases in Multimodal Models}, 
      author={Mounia Drissi},
      year={2024},
      eprint={2412.17505},
      archivePrefix={arXiv},
      primaryClass={stat.ML},
      url={https://arxiv.org/abs/2412.17505}, 
}

@misc{su2023roformerenhancedtransformerrotary,
      title={RoFormer: Enhanced Transformer with Rotary Position Embedding}, 
      author={Jianlin Su and Yu Lu and Shengfeng Pan and Ahmed Murtadha and Bo Wen and Yunfeng Liu},
      year={2023},
      eprint={2104.09864},
      archivePrefix={arXiv},
      primaryClass={cs.CL},
      url={https://arxiv.org/abs/2104.09864}, 
}

@misc{zheng2023judgingllmasajudgemtbenchchatbot,
      title={Judging LLM-as-a-Judge with MT-Bench and Chatbot Arena}, 
      author={Lianmin Zheng and Wei-Lin Chiang and Ying Sheng and Siyuan Zhuang and Zhanghao Wu and Yonghao Zhuang and Zi Lin and Zhuohan Li and Dacheng Li and Eric P. Xing and Hao Zhang and Joseph E. Gonzalez and Ion Stoica},
      year={2023},
      eprint={2306.05685},
      archivePrefix={arXiv},
      primaryClass={cs.CL},
      url={https://arxiv.org/abs/2306.05685}, 
}

@misc{tang2019coinlargescaledatasetcomprehensive,
      title={COIN: A Large-scale Dataset for Comprehensive Instructional Video Analysis}, 
      author={Yansong Tang and Dajun Ding and Yongming Rao and Yu Zheng and Danyang Zhang and Lili Zhao and Jiwen Lu and Jie Zhou},
      year={2019},
      eprint={1903.02874},
      archivePrefix={arXiv},
      primaryClass={cs.CV},
      url={https://arxiv.org/abs/1903.02874}, 
}

@misc{zhukov2019crosstaskweaklysupervisedlearning,
      title={Cross-task weakly supervised learning from instructional videos}, 
      author={Dimitri Zhukov and Jean-Baptiste Alayrac and Ramazan Gokberk Cinbis and David Fouhey and Ivan Laptev and Josef Sivic},
      year={2019},
      eprint={1903.08225},
      archivePrefix={arXiv},
      primaryClass={cs.CV},
      url={https://arxiv.org/abs/1903.08225}, 
}

@misc{sener2022assembly101largescalemultiviewvideo,
      title={Assembly101: A Large-Scale Multi-View Video Dataset for Understanding Procedural Activities}, 
      author={Fadime Sener and Dibyadip Chatterjee and Daniel Shelepov and Kun He and Dipika Singhania and Robert Wang and Angela Yao},
      year={2022},
      eprint={2203.14712},
      archivePrefix={arXiv},
      primaryClass={cs.CV},
      url={https://arxiv.org/abs/2203.14712}, 
}

@misc{koh2023generatingimagesmultimodallanguage,
      title={Generating Images with Multimodal Language Models}, 
      author={Jing Yu Koh and Daniel Fried and Ruslan Salakhutdinov},
      year={2023},
      eprint={2305.17216},
      archivePrefix={arXiv},
      primaryClass={cs.CL},
      url={https://arxiv.org/abs/2305.17216}, 
}

@misc{menon2024generatingillustratedinstructions,
      title={Generating Illustrated Instructions}, 
      author={Sachit Menon and Ishan Misra and Rohit Girdhar},
      year={2024},
      eprint={2312.04552},
      archivePrefix={arXiv},
      primaryClass={cs.CV},
      url={https://arxiv.org/abs/2312.04552}, 
}

@misc{zhou2024storydiffusionconsistentselfattentionlongrange,
      title={StoryDiffusion: Consistent Self-Attention for Long-Range Image and Video Generation}, 
      author={Yupeng Zhou and Daquan Zhou and Ming-Ming Cheng and Jiashi Feng and Qibin Hou},
      year={2024},
      eprint={2405.01434},
      archivePrefix={arXiv},
      primaryClass={cs.CV},
      url={https://arxiv.org/abs/2405.01434}, 
}

@misc{liu2025onepromptonestoryfreelunchconsistenttexttoimage,
      title={One-Prompt-One-Story: Free-Lunch Consistent Text-to-Image Generation Using a Single Prompt}, 
      author={Tao Liu and Kai Wang and Senmao Li and Joost van de Weijer and Fahad Shahbaz Khan and Shiqi Yang and Yaxing Wang and Jian Yang and Ming-Ming Cheng},
      year={2025},
      eprint={2501.13554},
      archivePrefix={arXiv},
      primaryClass={cs.CV},
      url={https://arxiv.org/abs/2501.13554}, 
}

@InProceedings{Chen_2024_CVPR,
    author    = {Chen, Tsai-Shien and Siarohin, Aliaksandr and Menapace, Willi and Deyneka, Ekaterina and Chao, Hsiang-wei and Jeon, Byung Eun and Fang, Yuwei and Lee, Hsin-Ying and Ren, Jian and Yang, Ming-Hsuan and Tulyakov, Sergey},
    title     = {Panda-70M: Captioning 70M Videos with Multiple Cross-Modality Teachers},
    booktitle = {Proceedings of the IEEE/CVF Conference on Computer Vision and Pattern Recognition (CVPR)},
    month     = {June},
    year      = {2024},
    pages     = {13320-13331}
}

@inproceedings{huang-etal-2025-empirical,
    title = "An Empirical Study of {LLM}-as-a-Judge for {LLM} Evaluation: Fine-tuned Judge Model is not a General Substitute for {GPT}-4",
    author = "Huang, Hui  and
      Bu, Xingyuan  and
      Zhou, Hongli  and
      Qu, Yingqi  and
      Liu, Jing  and
      Yang, Muyun  and
      Xu, Bing  and
      Zhao, Tiejun",
    editor = "Che, Wanxiang  and
      Nabende, Joyce  and
      Shutova, Ekaterina  and
      Pilehvar, Mohammad Taher",
    booktitle = "Findings of the Association for Computational Linguistics: ACL 2025",
    month = jul,
    year = "2025",
    address = "Vienna, Austria",
    publisher = "Association for Computational Linguistics",
    url = "https://aclanthology.org/2025.findings-acl.306/",
    doi = "10.18653/v1/2025.findings-acl.306",
    pages = "5880--5895",
    ISBN = "979-8-89176-256-5"
}

@misc{yu2025benchmarkinglargevisionlanguagemodels,
      title={Benchmarking Large Vision-Language Models on Fine-Grained Image Tasks: A Comprehensive Evaluation}, 
      author={Hong-Tao Yu and Xiu-Shen Wei and Yuxin Peng and Serge Belongie},
      year={2025},
      eprint={2504.14988},
      archivePrefix={arXiv},
      primaryClass={cs.CV},
      url={https://arxiv.org/abs/2504.14988}, 
}

@article{LI2026104270,
title = {A fine-grained evaluation framework for language models: Combining pointwise grading and pairwise comparison},
journal = {Information Processing \& Management},
volume = {63},
number = {1},
pages = {104270},
year = {2026},
issn = {0306-4573},
doi = {https://doi.org/10.1016/j.ipm.2025.104270},
url = {https://www.sciencedirect.com/science/article/pii/S0306457325002110},
author = {Yijie Li and Yuan Sun}
}

@article{chen2025beyond,
  title={Beyond Single-Point Judgment: Distribution Alignment for LLM-as-a-Judge},
  author={Chen, Luyu and Zhang, Zeyu and Tan, Haoran and Dai, Quanyu and Yang, Hao and Dong, Zhenhua and Chen, Xu},
  journal={arXiv preprint arXiv:2505.12301},
  year={2025}
}

@article{fang2024mmbench,
  title={Mmbench-video: A long-form multi-shot benchmark for holistic video understanding},
  author={Fang, Xinyu and Mao, Kangrui and Duan, Haodong and Zhao, Xiangyu and Li, Yining and Lin, Dahua and Chen, Kai},
  journal={Advances in Neural Information Processing Systems},
  volume={37},
  pages={89098--89124},
  year={2024}
}

@article{qi2025vcr,
  title={Vcr-bench: A comprehensive evaluation framework for video chain-of-thought reasoning},
  author={Qi, Yukun and Zhao, Yiming and Zeng, Yu and Bao, Xikun and Huang, Wenxuan and Chen, Lin and Chen, Zehui and Zhao, Jie and Qi, Zhongang and Zhao, Feng},
  journal={arXiv preprint arXiv:2504.07956},
  year={2025}
}

@article{he2024mmworld,
  title={Mmworld: Towards multi-discipline multi-faceted world model evaluation in videos},
  author={He, Xuehai and Feng, Weixi and Zheng, Kaizhi and Lu, Yujie and Zhu, Wanrong and Li, Jiachen and Fan, Yue and Wang, Jianfeng and Li, Linjie and Yang, Zhengyuan and others},
  journal={arXiv preprint arXiv:2406.08407},
  year={2024}
}

@inproceedings{liu2024tempcompass,
  title={Tempcompass: Do video llms really understand videos?},
  author={Liu, Yuanxin and Li, Shicheng and Liu, Yi and Wang, Yuxiang and Ren, Shuhuai and Li, Lei and Chen, Sishuo and Sun, Xu and Hou, Lu},
  booktitle={Findings of the Association for Computational Linguistics: ACL 2024},
  pages={8731--8772},
  year={2024}
}

@inproceedings{sun2025t2v,
  title={T2v-compbench: A comprehensive benchmark for compositional text-to-video generation},
  author={Sun, Kaiyue and Huang, Kaiyi and Liu, Xian and Wu, Yue and Xu, Zihan and Li, Zhenguo and Liu, Xihui},
  booktitle={Proceedings of the Computer Vision and Pattern Recognition Conference},
  pages={8406--8416},
  year={2025}
}

@inproceedings{qiu2025step,
  title={STEP: Enhancing Video-LLMs' Compositional Reasoning by Spatio-Temporal Graph-guided Self-Training},
  author={Qiu, Haiyi and Gao, Minghe and Qian, Long and Pan, Kaihang and Yu, Qifan and Li, Juncheng and Wang, Wenjie and Tang, Siliang and Zhuang, Yueting and Chua, Tat-Seng},
  booktitle={Proceedings of the IEEE/CVF Conference on Computer Vision and Pattern Recognition},
  pages={3284--3294},
  year={2025}
}

@inproceedings{loginova2025deep,
  title={Deep Temporal Reasoning in Video Language Models: A Cross-Linguistic Evaluation of Action Duration and Completion through Perfect Times},
  author={Loginova, Olga and Loguinova, Sof{\'\i}a Ortega},
  booktitle={Proceedings of the 63rd Annual Meeting of the Association for Computational Linguistics (Volume 1: Long Papers)},
  pages={20472--20502},
  year={2025}
}

@inproceedings{wu2025videoqa,
  title={VideoQA-TA: Temporal-Aware Multi-Modal Video Question Answering},
  author={Wu, Zhixuan and Cheng, Bo and Han, Jiale and Ma, Jiabao and Zhang, Shuhao and Chen, Yuli and Li, Changbo},
  booktitle={Proceedings of the 31st International Conference on Computational Linguistics},
  pages={7239--7252},
  year={2025}
}

@inproceedings{liang2025reasvqa,
  title={ReasVQA: Advancing VideoQA with imperfect reasoning process},
  author={Liang, Jianxin and Meng, Xiaojun and Zhang, Huishuai and Wang, Yueqian and Wei, Jiansheng and Zhao, Dongyan},
  booktitle={Proceedings of the 2025 Conference of the Nations of the Americas Chapter of the Association for Computational Linguistics: Human Language Technologies (Volume 1: Long Papers)},
  pages={1696--1709},
  year={2025}
}

@inproceedings{gao2025building,
  title={Building embodied evoagent: A brain-inspired paradigm for bridging multimodal large models and world models},
  author={Gao, Junyu and Yao, Xuan and Rui, Yong and Xu, Changsheng},
  booktitle={Proceedings of the 33rd ACM International Conference on Multimedia},
  pages={3280--3289},
  year={2025}
}

@inproceedings{kang2025calibclip,
  title={Calibclip: Contextual calibration of dominant semantics for text-driven image retrieval},
  author={Kang, Bin and Chen, Bin and Wang, Junjie and Li, Yulin and Zhao, Junzhi and Wang, Junle and Tian, Zhuotao},
  booktitle={Proceedings of the 33rd ACM International Conference on Multimedia},
  pages={5140--5149},
  year={2025}
}

@inproceedings{liu2025aster,
  title={Aster: Adaptive dynamic layer-skipping for efficient transformer inference via markov decision process},
  author={Liu, Fangxin and Wang, Junjie and Yang, Ning and Wang, Zongwu and Zhao, Junping and Jiang, Li and Guan, Haibing},
  booktitle={Proceedings of the 33rd ACM International Conference on Multimedia},
  pages={11853--11861},
  year={2025}
}

@inproceedings{guo2025serial,
  title={Serial position effects of large language models},
  author={Guo, Xiaobo and Vosoughi, Soroush},
  booktitle={Findings of the Association for Computational Linguistics: ACL 2025},
  pages={927--953},
  year={2025}
}

@article{rottman2025learning,
  title={Learning about causal relations that change over time: primacy and recency over long timeframes in causal judgments and memory},
  author={Rottman, Benjamin M and Zhang, Yiwen},
  journal={Cognitive Research: Principles and Implications},
  volume={10},
  number={1},
  pages={9},
  year={2025},
  publisher={Springer}
}

@article{zhang2024confronting,
  title={Confronting reward overoptimization for diffusion models: A perspective of inductive and primacy biases},
  author={Zhang, Ziyi and Zhang, Sen and Zhan, Yibing and Luo, Yong and Wen, Yonggang and Tao, Dacheng},
  journal={arXiv preprint arXiv:2402.08552},
  year={2024}
}

@article{sinclair2024first,
  title={First impressions or good endings? Preferences depend on when you ask.},
  author={Sinclair, Alyssa H and Wang, Yuxi C and Adcock, R Alison},
  journal={Journal of Experimental Psychology: General},
  year={2024},
  publisher={American Psychological Association}
}

@article{gu2024attention,
  title={When attention sink emerges in language models: An empirical view},
  author={Gu, Xiangming and Pang, Tianyu and Du, Chao and Liu, Qian and Zhang, Fengzhuo and Du, Cunxiao and Wang, Ye and Lin, Min},
  journal={arXiv preprint arXiv:2410.10781},
  year={2024}
}

@article{rulli2025attention,
  title={Attention sinks in diffusion language models},
  author={Rulli, Maximo Eduardo and Petruzzi, Simone and Michielon, Edoardo and Silvestri, Fabrizio and Scardapane, Simone and Devoto, Alessio},
  journal={arXiv preprint arXiv:2510.15731},
  year={2025}
}

@inproceedings{liu2026sinktrack,
  title={SinkTrack: Attention Sink based Context Anchoring for Large Language Models},
  author={Liu, Xu and Chen, Guikun and Wang, Wenguan},
  booktitle={The Fourteenth International Conference on Learning Representations},
  year={2026}
}

@article{allen1983maintaining,
  title   = {Maintaining knowledge about temporal intervals},
  author  = {Allen, James F.},
  journal = {Communications of the ACM},
  volume  = {26},
  number  = {11},
  pages   = {832--843},
  year    = {1983},
}

@misc{qwen3.6-27b,
    title  = {{Qwen3.6-27B}: Flagship-Level Coding in a {27B} Dense Model},
    author = {{Qwen Team}},
    month  = {April},
    year   = {2026},
    url    = {https://qwen.ai/blog?id=qwen3.6-27b}
}

@misc{gemmateam2026gemma4technicalreport,
      title={Gemma 4 Technical Report}, 
       author={{Gemma Team}},
      year={2026},
      eprint={2607.02770},
      archivePrefix={arXiv},
      primaryClass={cs.CL},
      url={https://arxiv.org/abs/2607.02770}, 
}

\clearpage
\appendix
\setcounter{figure}{0}
\renewcommand{\thefigure}{S\arabic{figure}}
\setcounter{table}{0}
\renewcommand{\thetable}{S\arabic{table}}
\setcounter{section}{0}
\renewcommand{\thesection}{\Alph{section}}

\twocolumn[
  \begin{center}
    \LARGE \textbf{Supplementary Material:\\Order Matters: LVLMs as Judges for Temporal Reasoning in Image Sequences}
    \vspace{1.5em}
  \end{center}
]

\section*{Overview}
This document provides the technical details, extended experimental
results, and qualitative analyses that complement the main paper.
The content is organised as follows.

\begin{figure}[t]
\centering
\newlength{\stepcolW}\setlength{\stepcolW}{0.235\columnwidth} 
\newlength{\stepimgH}\setlength{\stepimgH}{0.19\columnwidth}
\newlength{\stepboxH}\setlength{\stepboxH}{3.5cm} 

\centering
\begin{minipage}[t]{\stepcolW}%
  \begin{tcolorbox}[enhanced, sharp corners=all, colback=stepbg, colframe=stepframe, 
      boxrule=0.6pt, left=0pt, right=0pt, top=0pt, bottom=2pt, 
      width=\linewidth, height=\stepboxH, nobeforeafter, valign=top]
    \includegraphics[width=\linewidth,height=\stepimgH,keepaspectratio=false]{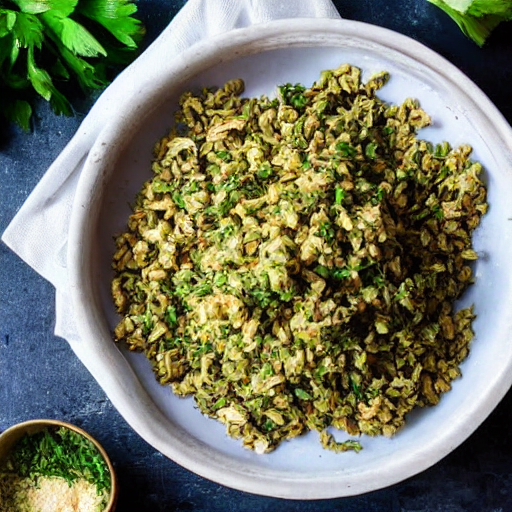}\\[3pt]
    \hspace{4pt}{\scriptsize\textbf{Step 1}}\\[1pt]
    \hspace{4pt}\begin{minipage}{\dimexpr\linewidth-8pt}
      \scriptsize\raggedright Marinade: garlic, paprika, lime zest, rosemary and buttermilk. \par
    \end{minipage}
  \end{tcolorbox}
\end{minipage}\hfill
\begin{minipage}[t]{\stepcolW}%
  \begin{tcolorbox}[enhanced, sharp corners=all, colback=stepbg, colframe=stepframe, 
      boxrule=0.6pt, left=0pt, right=0pt, top=0pt, bottom=2pt, 
      width=\linewidth, height=\stepboxH, nobeforeafter, valign=top]
    \includegraphics[width=\linewidth,height=\stepimgH,keepaspectratio=false]{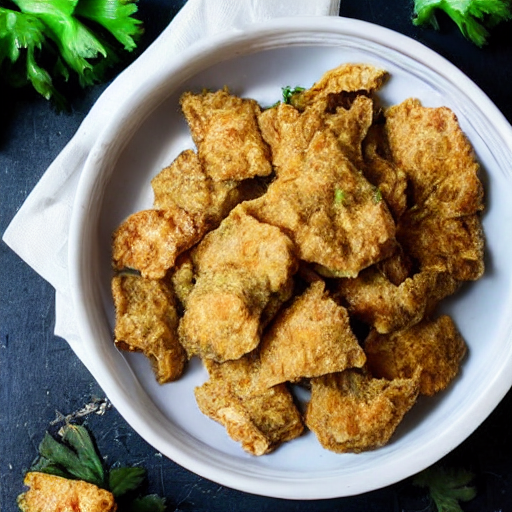}\\[3pt]
    \hspace{4pt}{\scriptsize\textbf{Step 2}}\\[1pt]
    \hspace{4pt}\begin{minipage}{\dimexpr\linewidth-8pt}
      \scriptsize\raggedright Chicken wings marinating in a mixture bowl.\par
    \end{minipage}
  \end{tcolorbox}
\end{minipage}\hfill
\begin{minipage}[t]{\stepcolW}%
  \begin{tcolorbox}[enhanced, sharp corners=all, colback=stepbg, colframe=stepframe, 
      boxrule=0.6pt, left=0pt, right=0pt, top=0pt, bottom=2pt, 
      width=\linewidth, height=\stepboxH, nobeforeafter, valign=top]
    \includegraphics[width=\linewidth,height=\stepimgH,keepaspectratio=false]{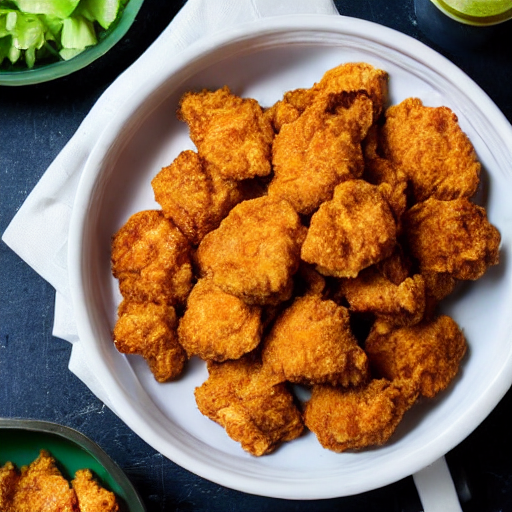}\\[3pt]
    \hspace{4pt}{\scriptsize\textbf{Step 3}}\\[1pt]
    \hspace{4pt}\begin{minipage}{\dimexpr\linewidth-8pt}
      \scriptsize\raggedright Chicken wings coated in breadcrumbs and spices in a bowl.\par
    \end{minipage}
  \end{tcolorbox}
\end{minipage}\hfill
\begin{minipage}[t]{\stepcolW}%
  \begin{tcolorbox}[enhanced, sharp corners=all, colback=stepbg, colframe=stepframe, 
      boxrule=0.6pt, left=0pt, right=0pt, top=0pt, bottom=2pt, 
      width=\linewidth, height=\stepboxH, nobeforeafter, valign=top]
    \includegraphics[width=\linewidth,height=\stepimgH,keepaspectratio=false]{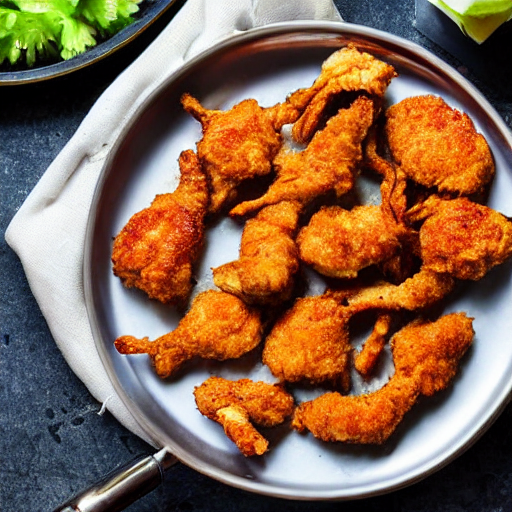}\\[3pt]
    \hspace{4pt}{\scriptsize\textbf{Step 4}}\\[1pt]
    \hspace{4pt}\begin{minipage}{\dimexpr\linewidth-8pt}
      \scriptsize\raggedright Chicken wings being fried in hot vegetable oil in a pan.\par
    \end{minipage}
  \end{tcolorbox}
\end{minipage}

\vspace{-4pt}
\caption{%
  \textbf{Example of a temporal sequence from our benchmarks (\textsc{Prism} and \textsc{Mirage}).}
  Each image in the procedural flow is integrated with its corresponding textual step description in a unified block. 
}
\label{fig:benchmark_example}
\end{figure}

\textbf{Section~\ref{sec:prompts} — Prompting Architectures.}  We
formalise all prompt configurations (P0–P6 and P$\star$) used for
both pointwise and pairwise evaluation.  We describe the exact
instruction templates fed to the LVLM judges and illustrate how the
relative ordering of textual steps and images changes across
configurations. Figure~\ref{fig:benchmark_example} provides a 
representative example of the interleaved visual-textual structure 
characterising our benchmarks.

\textbf{Section~\ref{sec:training} — Experimental Setup and Training
Details.}  We report the full set of LoRA hyperparameters selected
by Optuna for each fine-tuning run, together with the training dataset
statistics for \textsc{Prism} and \textsc{Mirage}.

\textbf{Section~\ref{sec:results} — Extended Experimental Results.}
We provide the complete per-prompt breakdown of pointwise scoring
and zero-shot/fine-tuned pairwise evaluation on both benchmarks,
including NLP overlap metrics omitted from the main paper for space.

\textbf{Section~\ref{sec:positional} — Positional Sensitivity Analysis.}
We report the full positional sensitivity tables and heatmaps that
motivate the primacy/recency discussion in Section~6.3 of the main paper.

\textbf{Section~\ref{sec:larger_models} — Evaluation on Larger Models.}
We report zero-shot pairwise results for two substantially larger
vision-language models, \textsc{Gemma-4}-31B and \textsc{Qwen3.6}-27B,
on the same benchmarks and prompt configurations as the main paper.

\textbf{Section~\ref{sec:sensitivity_larger} — Positional Sensitivity Analysis on Larger Models.}
We replicate the positional sensitivity analysis of Section~\ref{sec:positional}
on Gemma-4-31B and Qwen3.6-27B to assess whether primacy and recency effects
persist across model scale and architectural families.

\textbf{Section~\ref{sec:probe} — Visual Chronological Ordering Probe.}
We introduce a diagnostic task in which models must order four
historical images by date using visual content alone, with no textual
scaffolding.

\textbf{Section~\ref{sec:qualitative} — Qualitative Case Studies.}
We present representative examples comparing fine-tuned and
zero-shot model outputs on \textsc{Prism} and \textsc{Mirage}
sequences.

\section{Prompting Architectures and Layouts}
\label{sec:prompts}

A central experimental variable in our study is \emph{how} input
elements (textual step descriptions and images) are arranged within
the prompt.  We evaluate seven distinct configurations, labelled
P0--P6, plus a multi-prompt variant P$\star$ used for the
multi-prompt fine-tuning (OneVision$^\ddagger$).  These
configurations differ along two orthogonal axes: (i) whether steps
and images are \emph{interleaved} or \emph{grouped}, and (ii)
whether images from the two sequences (A and B) are presented
jointly or separately.

\subsection{Pointwise vs.\ Pairwise Task Definitions}

We evaluate temporal reasoning under two complementary protocols.

\textbf{Pointwise evaluation} asks the judge to score a single image
sequence on a 1–5 Likert scale reflecting logical and temporal
coherence.  The absolute nature of this task makes it susceptible to
distributional bias (Section~3 of the main paper), where a model can
assign consistently high scores without actually detecting ordering
violations.

\textbf{Pairwise evaluation} asks the judge to compare two sequences
and select which one better adheres to the described process.  By
design, this protocol forces fine-grained discrimination and is more
robust to score-range collapse, yet it introduces its own vulnerability
to positional biases (primacy/recency effects).

Both protocols are applied to all configurations P0--P6 with both
\textsc{LLaVA-OneVision} and \textsc{LLaVA-Critic} as judge
backbones.

\subsection{Configuration Structure}

Table~\ref{tab:input_prompt} summarises the seven prompt
configurations.  In the \emph{pointwise} column each prompt contains
steps and images from a single sequence; configurations~0 and~4
collapse to the same layout in this regime.  In the \emph{pairwise}
column the textual steps are held constant while images are drawn
from sequence~A and sequence~B.  The colour coding used throughout
the table is:
\textcolor{stepcolor}{$\blacksquare$}~textual steps,
\textcolor{imgAcolor}{$\blacksquare$}~Sequence~A images,
\textcolor{imgBcolor}{$\blacksquare$}~Sequence~B images.

\begin{table*}[htbp]
\centering
\caption{%
  \textbf{Prompt configuration structure.}
  Structure of the seven prompt configurations tested for pointwise
  and pairwise evaluation with \textsc{LLaVA-OneVision} and
  \textsc{LLaVA-Critic}.
  Each row describes a different organisation of the input elements.
  In the \emph{pointwise} column each prompt contains descriptive steps
  and images from a single sequence; note that configurations~0 and~4
  become identical in this setting.
  In the \emph{pairwise} column the descriptive steps remain consistent
  while images represent elements from sequence~A and sequence~B.
  \textcolor{stepcolor}{\rule{8pt}{8pt}}~Steps,
  \textcolor{imgAcolor}{\rule{8pt}{8pt}}~Sequence~A images,
  \textcolor{imgBcolor}{\rule{8pt}{8pt}}~Sequence~B images.%
}
\label{tab:input_prompt}
\scriptsize
\setlength{\tabcolsep}{6pt}
\begin{tabular}{c p{5.5cm} p{8cm}}
\toprule
\textbf{\#} & \textbf{Pointwise Input Structure} & \textbf{Pairwise Input Structure} \\
\midrule

0 &
\centering\itshape Pairwise-only &
\colorbox{stepcolor}{\texttt{[step1]}}
\colorbox{stepcolor}{\texttt{[step2]}}\\[2pt]
& &
\colorbox{imgAcolor}{\texttt{[img1A]}}
\colorbox{imgAcolor}{\texttt{[img2A]}}
\colorbox{imgBcolor}{\texttt{[img1B]}}
\colorbox{imgBcolor}{\texttt{[img2B]}} \\[4pt]

\midrule
1 &
\colorbox{stepcolor}{\texttt{[step1]}}
\colorbox{imgAcolor}{\texttt{[img1]}}
\colorbox{stepcolor}{\texttt{[step2]}}
\colorbox{imgAcolor}{\texttt{[img2]}} &
\colorbox{stepcolor}{\texttt{[step1]}}
\colorbox{imgAcolor}{\texttt{[img1A]}}
\colorbox{stepcolor}{\texttt{[step2]}}
\colorbox{imgAcolor}{\texttt{[img2A]}} \\[2pt]
& &
\colorbox{stepcolor}{\texttt{[step1]}}
\colorbox{imgBcolor}{\texttt{[img1B]}}
\colorbox{stepcolor}{\texttt{[step2]}}
\colorbox{imgBcolor}{\texttt{[img2B]}} \\[4pt]

\midrule
2 &
\colorbox{imgAcolor}{\texttt{[img1]}}
\colorbox{stepcolor}{\texttt{[step1]}}
\colorbox{imgAcolor}{\texttt{[img2]}}
\colorbox{stepcolor}{\texttt{[step2]}} &
\colorbox{imgAcolor}{\texttt{[img1A]}}
\colorbox{stepcolor}{\texttt{[step1]}}
\colorbox{imgAcolor}{\texttt{[img2A]}}
\colorbox{stepcolor}{\texttt{[step2]}} \\[2pt]
& &
\colorbox{imgBcolor}{\texttt{[img1B]}}
\colorbox{stepcolor}{\texttt{[step1]}}
\colorbox{imgBcolor}{\texttt{[img2B]}}
\colorbox{stepcolor}{\texttt{[step2]}} \\[4pt]

\midrule
3 &
\colorbox{imgAcolor}{\texttt{[img1]}}
\colorbox{imgAcolor}{\texttt{[img2]}}
\colorbox{stepcolor}{\texttt{[step1]}}
\colorbox{stepcolor}{\texttt{[step2]}} &
\colorbox{imgAcolor}{\texttt{[img1A]}}
\colorbox{imgAcolor}{\texttt{[img2A]}}
\colorbox{stepcolor}{\texttt{[step1]}}
\colorbox{stepcolor}{\texttt{[step2]}} \\[2pt]
& &
\colorbox{imgBcolor}{\texttt{[img1B]}}
\colorbox{imgBcolor}{\texttt{[img2B]}}
\colorbox{stepcolor}{\texttt{[step1]}}
\colorbox{stepcolor}{\texttt{[step2]}} \\[4pt]

\midrule
4 &
\colorbox{stepcolor}{\texttt{[step1]}}
\colorbox{stepcolor}{\texttt{[step2]}}
\colorbox{imgAcolor}{\texttt{[img1]}}
\colorbox{imgAcolor}{\texttt{[img2]}} &
\colorbox{stepcolor}{\texttt{[step1]}}
\colorbox{stepcolor}{\texttt{[step2]}}
\colorbox{imgAcolor}{\texttt{[img1A]}}
\colorbox{imgAcolor}{\texttt{[img2A]}} \\[2pt]
& &
\colorbox{stepcolor}{\texttt{[step1]}}
\colorbox{stepcolor}{\texttt{[step2]}}
\colorbox{imgBcolor}{\texttt{[img1B]}}
\colorbox{imgBcolor}{\texttt{[img2B]}} \\[4pt]

\midrule
5 &
\colorbox{stepcolor}{\texttt{[step1]}}
\colorbox{stepcolor}{\texttt{[step2]}}
\colorbox{imgAcolor}{\texttt{[img1\ img2]}} &
\colorbox{stepcolor}{\texttt{[step1]}}
\colorbox{stepcolor}{\texttt{[step2]}}
\colorbox{imgAcolor}{\texttt{[img1A\ img2A]}} \\[2pt]
& &
\colorbox{stepcolor}{\texttt{[step1]}}
\colorbox{stepcolor}{\texttt{[step2]}}
\colorbox{imgBcolor}{\texttt{[img1B\ img2B]}} \\[4pt]

\midrule
6 &
\centering\itshape Pairwise-only &
\colorbox{stepcolor}{\texttt{[step1\ step2]}} \\[2pt]
& &
\colorbox{imgAcolor}{\texttt{[img1A\ img2A]}}
\colorbox{imgBcolor}{\texttt{[img1B\ img2B]}} \\[2pt]

\bottomrule
\end{tabular}
\end{table*}

\subsection{Instruction Templates}

Figure~\ref{fig:evaluation_prompts} reports the exact system
instructions given to the LVLM judges for pointwise and pairwise
evaluation.  The pointwise prompt defines the 5-point Likert
rubric, anchoring each score to a specific failure mode
(score~3 penalises ordering violations; scores~1--2 target content
mismatches).  The pairwise prompt requires the model to commit to
either Sequence~A or Sequence~B before providing a rationale, a
design choice intended to limit post-hoc rationalization.

\begin{figure*}[t]
\centering
\caption{%
  \textbf{Evaluation prompt templates (Figure~S1).}
  System instructions for pointwise (\emph{left}) and pairwise
  (\emph{right}) protocols.  The pointwise template anchors each
  Likert score to a specific failure mode, while the pairwise
  template commits the model to a binary choice before reasoning.
  The placeholder \textit{[Input Layout]} refers to one of the
  configurations in Table~\ref{tab:input_prompt}.
}
\label{fig:evaluation_prompts}
\begin{minipage}[t]{0.48\textwidth}
\scriptsize\textbf{Pointwise Evaluation (System Instruction):}
\vspace{0.3em}
\begin{tcolorbox}[colback=blue!5,colframe=blue!75!black,arc=3pt,
  boxrule=0.7pt,left=5pt,right=5pt,top=5pt,bottom=5pt,width=\linewidth]
\footnotesize
Given a visual sequence and a textual description of a process,
serve as an unbiased judge to evaluate how well the visual sequence
represents the logical and coherent flow of the described process.
Rate the sequence from 1 to 5 and explain your reasoning with
specific details.

\medskip\textit{[Input Layout — see Table~\ref{tab:input_prompt}]}\\

\medskip\textbf{Rating Scale:}\\
\textbf{5}: Perfect visual representation\\
\textbf{4}: Good representation with minor imperfections\\
\textbf{3}: Correct images but incorrect temporal order\\
\textbf{2}: Mostly correct but one or more images do not match\\
\textbf{1}: Poor quality — images do not represent the process\\

\medskip\textbf{Response Format:}\\[1pt]
\texttt{Score: [X/5]. Reasoning: [...]. ASSISTANT:}
\end{tcolorbox}
\end{minipage}
\hfill
\begin{minipage}[t]{0.48\textwidth}
\scriptsize\textbf{Pairwise Evaluation (System Instruction):}
\vspace{0.3em}
\begin{tcolorbox}[colback=green!5,colframe=green!75!black,arc=3pt,
  boxrule=0.7pt,left=5pt,right=5pt,top=5pt,bottom=5pt,width=\linewidth]
\footnotesize
Given two visual sequences and a textual description of a process,
serve as an unbiased judge to evaluate which visual sequence better
represents the logical and coherent flow of the described process.
Determine which sequence (A or~B) is better and explain your
reasoning with specific details.

\medskip\textit{[Input Layout — see Table~\ref{tab:input_prompt}]}\\

\medskip\textbf{Options:}\\

\textbf{Sequence A}\\
\textbf{Sequence B}\\

\medskip\textbf{Response Format:}\\[2pt]
\texttt{Better Sequence: [A or B].}\\
\texttt{Reasoning: [...]. ASSISTANT:}
\end{tcolorbox}
\end{minipage}
\end{figure*}

\subsection{Taxonomy of Diagnostic Perturbations}

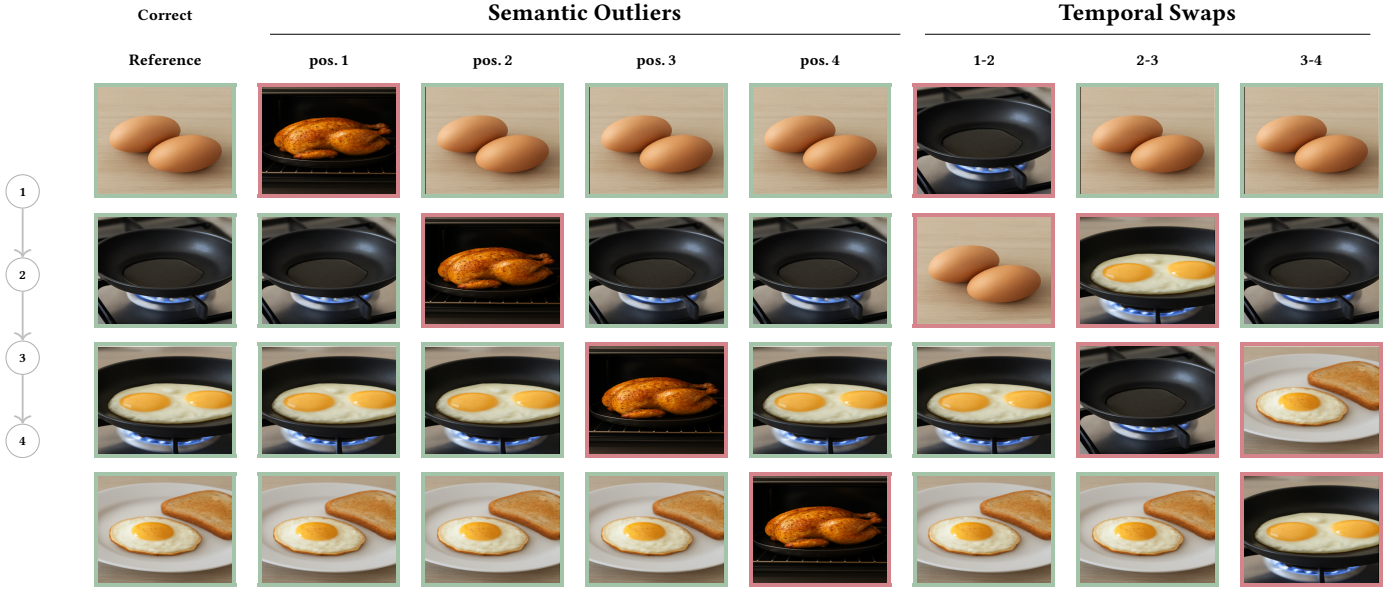
\begin{figure*}[t]
\centering
\definecolor{cok}{HTML}{A3C4A8}
\definecolor{cerr}{HTML}{D4848A}
\setlength{\tabcolsep}{4pt}
\setlength{\tW}{0.10\linewidth}
\setlength{\tH}{0.08\linewidth}
\setlength{\tlW}{0.05\linewidth}
\newcommand{\fr}[2]{%
  \setlength{\fboxsep}{0pt}\setlength{\fboxrule}{1.5pt}%
  \fcolorbox{#1}{white}{\includegraphics[width=\tW,height=\tH,
    keepaspectratio=false]{#2}}}
\newcommand{\collbl}[1]{\makebox[\tW]{\scriptsize\bfseries #1}}
\begin{tabular}{@{} p{\tlW} c @{\hspace{8pt}} cccc @{\hspace{8pt}} ccc @{}}
& \collbl{Correct} &
  \multicolumn{4}{c}{\textbf{Semantic Outliers}} &
  \multicolumn{3}{c}{\textbf{Temporal Swaps}} \\
\cmidrule(lr){3-6}\cmidrule(lr){7-9}
& \collbl{Reference} & \collbl{pos.\,1} & \collbl{pos.\,2} &
  \collbl{pos.\,3} & \collbl{pos.\,4} &
  \collbl{1-2} & \collbl{2-3} & \collbl{3-4} \\[3pt]
\multirow{4}{*}{\begin{tikzpicture}[baseline=(n1.center),
  nd/.style={circle,draw=gray!60,fill=white,font=\tiny\bfseries,minimum size=11pt}]
  \node[nd](n1)at(0,0){1};\node[nd](n2)at(0,-1.1){2};
  \node[nd](n3)at(0,-2.2){3};\node[nd](n4)at(0,-3.3){4};
  \draw[->,gray!50,line width=0.7pt](n1)--(n2);
  \draw[->,gray!50,line width=0.7pt](n2)--(n3);
  \draw[->,gray!50,line width=0.7pt](n3)--(n4);
\end{tikzpicture}}
& \fr{cok}{img/egg1} & \fr{cerr}{img/chicken_real} & \fr{cok}{img/egg1} & \fr{cok}{img/egg1} & \fr{cok}{img/egg1} & \fr{cerr}{img/egg2} & \fr{cok}{img/egg1} & \fr{cok}{img/egg1} \\[3pt]
& \fr{cok}{img/egg2} & \fr{cok}{img/egg2} & \fr{cerr}{img/chicken_real} & \fr{cok}{img/egg2} & \fr{cok}{img/egg2} & \fr{cerr}{img/egg1} & \fr{cerr}{img/egg3} & \fr{cok}{img/egg2} \\[3pt]
& \fr{cok}{img/egg3} & \fr{cok}{img/egg3} & \fr{cok}{img/egg3} & \fr{cerr}{img/chicken_real} & \fr{cok}{img/egg3} & \fr{cok}{img/egg3} & \fr{cerr}{img/egg2} & \fr{cerr}{img/egg4} \\[3pt]
& \fr{cok}{img/egg4} & \fr{cok}{img/egg4} & \fr{cok}{img/egg4} & \fr{cok}{img/egg4} & \fr{cerr}{img/chicken_real} & \fr{cok}{img/egg4} & \fr{cok}{img/egg4} & \fr{cerr}{img/egg3} \\
\end{tabular}
\caption{%
  \textbf{Complete taxonomy of diagnostic perturbations (Figure~S2).}
  Each column shows a distinct perturbation of a four-step cooking sequence.
  \emph{Semantic Outliers} (columns 2–5): a contextually irrelevant image
  (red border) is inserted at positions 1–4 in turn.
  \emph{Temporal Swaps} (columns 6–8): two consecutive frames are exchanged
  at positions 1–2, 2–3, and 3–4.
  Green borders denote correct frames; red borders denote perturbed ones.
  This figure extends Figure~1 of the main paper using real recipe images.
}
\label{fig:supp_taxonomy}
\end{figure*}

Figure~\ref{fig:supp_taxonomy} extends Figure~1 of the main paper
by showing the complete set of controlled perturbations applied to a
four-step cooking sequence using real recipe images.  Two perturbation
families are considered:

\textbf{Semantic Outliers.}  A contextually irrelevant image (red
border) replaces the correct frame at each of the four possible
positions, independently.  This yields four negative variants per
gold sequence and tests whether the judge can detect a content
mismatch at any point in the narrative.

\textbf{Temporal Swaps.}  Two consecutive frames are exchanged,
inverting the procedural order locally.  Swaps are applied at all
three adjacent-pair positions in a four-frame sequence (1–2, 2–3,
3–4), producing three negative variants per gold sequence.  These
perturbations are more subtle than semantic outliers because the
image content remains correct; only the causal ordering is violated.

Together, the seven negative types per gold sequence define the
diagnostic coverage of \textsc{Prism} (Section~5 of the main paper).

\begin{figure*}[t]
\centering
\scriptsize
\newlength{\imgW}\setlength{\imgW}{0.072\textwidth}
\newlength{\imgH}\setlength{\imgH}{0.072\textwidth}
\newtcolorbox{scorebox}[2]{enhanced,colback=#1!5,colframe=#1!75!black,
  fonttitle=\bfseries\scriptsize,title=\textsc{Score: #2},boxrule=0.7pt,
  arc=1mm,width=0.31\textwidth,nobeforeafter,top=1mm,bottom=1mm,left=1mm,right=1mm}
\makebox[0.31\textwidth]{\scriptsize\textbf{Correct Sequence}}
\hfill\makebox[0.31\textwidth]{\scriptsize\textbf{Temporal Swap}}
\hfill\makebox[0.31\textwidth]{\scriptsize\textbf{Semantic Outlier}}\\[3pt]
\begin{minipage}{0.31\textwidth}\centering
  \includegraphics[width=\imgW,height=\imgH,keepaspectratio=false]{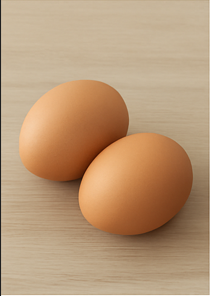}\hfill
  \includegraphics[width=\imgW,height=\imgH,keepaspectratio=false]{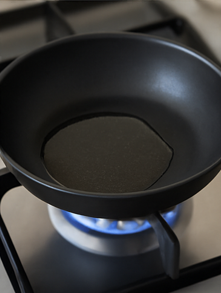}\hfill
  \includegraphics[width=\imgW,height=\imgH,keepaspectratio=false]{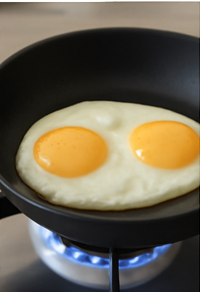}\hfill
  \includegraphics[width=\imgW,height=\imgH,keepaspectratio=false]{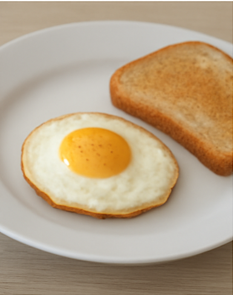}
\end{minipage}\hfill
\begin{minipage}{0.31\textwidth}\centering
  \includegraphics[width=\imgW,height=\imgH,keepaspectratio=false]{img/egg2.png}\hfill
  \includegraphics[width=\imgW,height=\imgH,keepaspectratio=false]{img/egg4.png}\hfill
  \includegraphics[width=\imgW,height=\imgH,keepaspectratio=false]{img/egg3.png}\hfill
  \includegraphics[width=\imgW,height=\imgH,keepaspectratio=false]{img/egg1.png}
\end{minipage}\hfill
\begin{minipage}{0.31\textwidth}\centering
  \includegraphics[width=\imgW,height=\imgH,keepaspectratio=false]{img/egg2.png}\hfill
  \includegraphics[width=\imgW,height=\imgH,keepaspectratio=false]{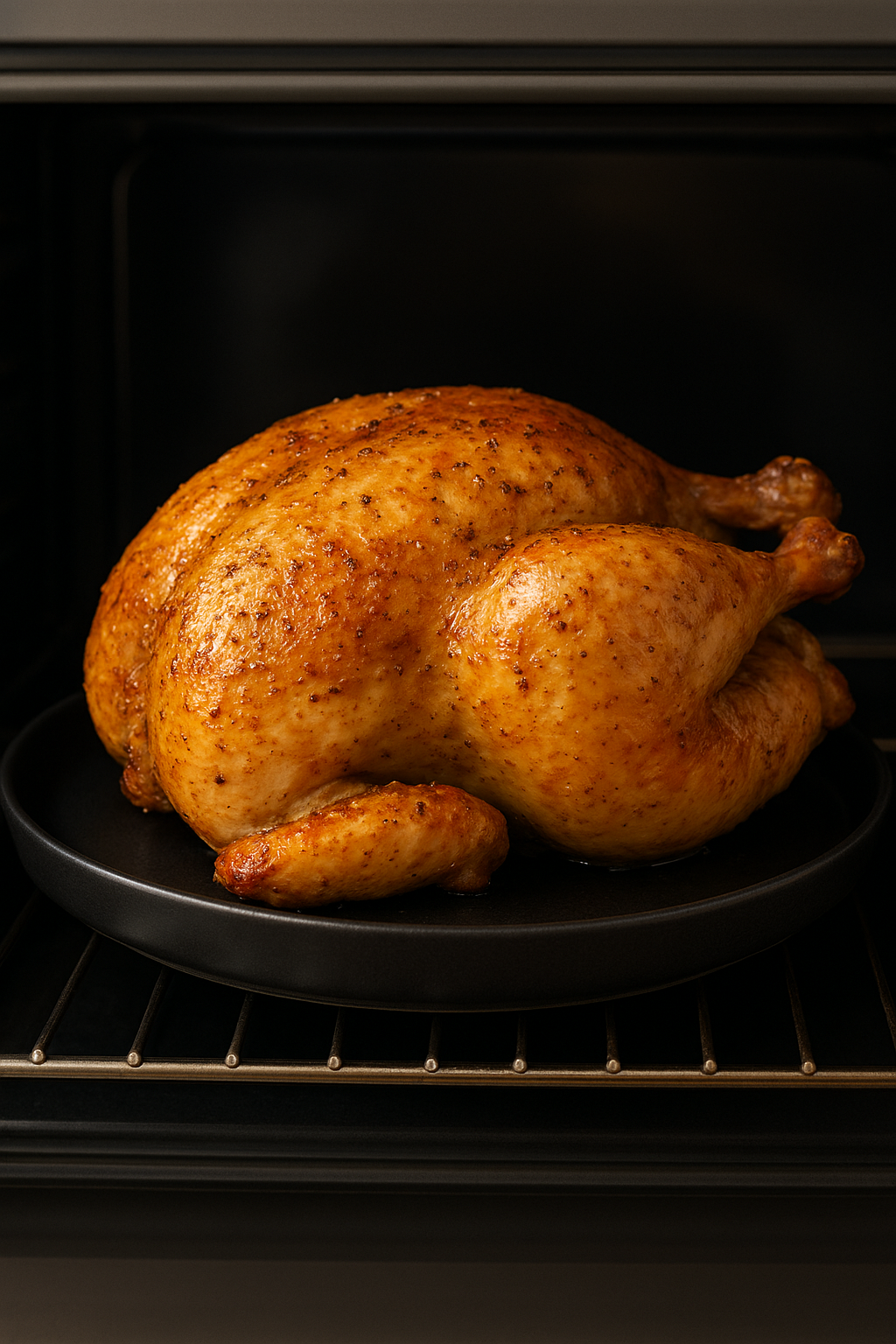}\hfill
  \includegraphics[width=\imgW,height=\imgH,keepaspectratio=false]{img/egg3.png}\hfill
  \includegraphics[width=\imgW,height=\imgH,keepaspectratio=false]{img/egg1.png}
\end{minipage}
\vspace{5pt}\\
\begin{scorebox}{teal}{5/5}
  \textbf{Coherent:} The sequence follows a logical culinary
  progression from raw ingredients to the completed dish.
\end{scorebox}\hfill
\begin{scorebox}{orange}{3/5}
  \textbf{Shuffled:} All correct frames are present but causal
  order is violated (scrambled temporal flow).
\end{scorebox}\hfill
\begin{scorebox}{red}{1/5}
  \textbf{Inconsistent:} An unrelated object (semantic outlier)
  is injected, breaking content and logical flow.
\end{scorebox}
\caption{%
  \textbf{Qualitative examples of pointwise scoring (Figure~S3).}
  Three variants of the same cooking sequence are scored by the
  LVLM judge under the pointwise rubric.
  The coherent sequence achieves 5/5; the temporally shuffled
  sequence is penalised to 3/5 despite correct frame content;
  the semantic outlier sequence receives 1/5 due to critical
  content inconsistency.  These examples illustrate the three
  principal failure modes targeted by \textsc{Prism}.
}
\label{fig:qualitative_scores}
\end{figure*}

Figure~\ref{fig:qualitative_scores} illustrates how the pointwise Likert
rubric maps onto the three perturbation outcomes for the same cooking
sequence.  A coherent sequence earns a perfect score~(5/5); a temporally
shuffled sequence is penalised for causal order violations despite
containing all correct frames~(3/5); and a sequence with a semantic
outlier receives the minimum score due to content inconsistency~(1/5).
These examples directly correspond to the three score anchors in the
pointwise prompt (Figure~\ref{fig:evaluation_prompts}).

\section{Experimental Setup and Training Details}
\label{sec:training}

\subsection{Hyperparameter Optimisation via Optuna}

All LoRA fine-tuning runs are configured through Bayesian
hyperparameter search with Optuna~\cite{10.1145/3292500.3330701} over 50
trials.  The search space covers LoRA rank $r \in \{16, 32, 64, 128\}$,
scaling factor $\alpha \in \{16, 32, 64, 128\}$, dropout rate
$\in [0.01, 0.10]$, learning rate $\in [2\!\times\!10^{-6},
5\!\times\!10^{-5}]$, maximum sequence length $\in \{4096, 8192,
12000\}$, and gradient clipping threshold.  The optimisation target is
macro-$F_1$ on \textsc{Prism}-Temporal, the subset most exposed to the
lost-in-the-middle effect (Section~3 of the main paper).

Table~\ref{tab:optuna-hparams} reports the winning configuration for
each of the nine fine-tuning runs that correspond to the model variants
reported in the main paper.  Across runs, higher-capacity LoRA
configurations ($r=128$, $\alpha=32$--$128$) tend to be selected for
reasoning-augmented (w/R) pairwise tasks, while lighter configurations
suffice for score-only objectives.  Gradient accumulation steps are
consistently high (4–12), reflecting the effective global batch size
of 128 used throughout.

\begin{table*}[htbp]
\centering
\caption{%
  \textbf{Optuna-selected hyperparameter configurations.}
  Best configuration per fine-tuning run as selected over 50 Bayesian
  trials.  \emph{Task}: \textsc{Pair}\,=\,pairwise discrimination;
  \textsc{Point}\,=\,pointwise scoring.
  \emph{Reasoning}: \cmark\,=\,CoT rationales included;
  \xmark\,=\,score label only.
  \emph{ID} maps to model variants in the extended results tables.
}
\label{tab:optuna-hparams}
\footnotesize
\renewcommand{\arraystretch}{1.15}
\setlength{\tabcolsep}{3.8pt}
\begin{tabular}{@{} l c c c  r r r  r r  r r r  r r  c @{}}
\toprule
\multicolumn{4}{c}{\textbf{Configuration}} &
\multicolumn{3}{c}{\textbf{LoRA}} &
\multicolumn{2}{c}{\textbf{Sequence}} &
\multicolumn{4}{c}{\textbf{Optimisation}} &
\multicolumn{1}{c}{\textbf{Training}} & \\
\cmidrule(lr){1-4}\cmidrule(lr){5-7}\cmidrule(lr){8-9}
\cmidrule(lr){10-13}\cmidrule(lr){14-15}
\textbf{Dataset} & \textbf{Reas.} & \textbf{Prompt} & \textbf{Task}
  & \textbf{\textit{r}} & $\boldsymbol{\alpha}$ & \textbf{Drop.}
  & \textbf{LR} & \textbf{Max Len}
  & \textbf{Grad.\ Clip} & \textbf{Warmup}
  & \textbf{W.\ Decay} & \textbf{Epochs}
  & \textbf{Grad.\ Acc.} & \textbf{ID} \\
\midrule
\rowcolor{gray!10}
\textsc{Mirage} & \xmark & P0 & \textsc{Pair}
  & 128 & 128 & 0.10 & 2e-5 & 12{,}000 & 0.3 & 0.05 & 0.00 & 6 & 12 & 1 \\
\textsc{Mirage} & \xmark & P6 & \textsc{Pair}
  & 32  & 16  & 0.10 & 1e-5 & 4{,}096  & 0.3 & 0.05 & 0.10 & 3 & 12 & 2 \\
\rowcolor{gray!10}
\textsc{Mirage} & \cmark & P0 & \textsc{Pair}
  & 128 & 32  & 0.05 & 5e-6 & 12{,}000 & 0.7 & 0.10 & 0.01 & 4 & 4  & 3 \\
\textsc{Prism}  & \cmark & P6 & \textsc{Pair}
  & 128 & 32  & 0.05 & 5e-6 & 12{,}000 & 0.7 & 0.10 & 0.01 & 4 & 4  & 4 \\
\rowcolor{gray!10}
\textsc{Prism}  & \xmark & P6 & \textsc{Pair}
  & 16  & 64  & 0.01 & 2e-5 & 4{,}096  & 0.5 & 0.01 & 0.05 & 3 & 12 & 5 \\
\textsc{Prism}  & \cmark & P5 & \textsc{Point}
  & 16  & 16  & 0.03 & 5e-6 & 4{,}096  & 1.0 & 0.03 & 0.00 & 4 & 4  & 6 \\
\rowcolor{gray!10}
\textsc{Prism}  & \cmark & P0 & \textsc{Pair}
  & 128 & 32  & 0.05 & 5e-6 & 12{,}000 & 0.7 & 0.10 & 0.01 & 4 & 4  & 7 \\
\textsc{Prism}  & \xmark & P$\star$ & \textsc{Pair}
  & 32  & 128 & 0.03 & 2e-5 & 12{,}000 & 0.3 & 0.05 & 0.05 & 5 & 8  & 8 \\
\rowcolor{gray!10}
\textsc{Prism}  & \cmark & P$\star$ & \textsc{Pair}
  & 32  & 128 & 0.10 & 5e-6 & 8{,}192  & 0.3 & 0.03 & 0.05 & 5 & 12 & 9 \\
\bottomrule
\end{tabular}
\end{table*}

\subsection{Training and Evaluation Dataset Statistics}
\label{sec:dataset_stats}

In this section, we provide a comprehensive breakdown of the datasets used for Supervised Fine-Tuning (SFT) and evaluation. Table~\ref{tab:comprehensive_stats} summarizes the record counts across splits and the character-level statistics of the Chain-of-Thought (CoT) rationales.

\begin{table*}[htbp]
\centering
\caption{%
  \textbf{Comprehensive Dataset Statistics for \textsc{Prism} and \textsc{Mirage}.} 
  Record counts per split and detailed CoT rationale length statistics (characters). 
  The \textsc{Prism}-Pointwise subset focuses on Likert-scale quality, while Pairwise subsets 
  target semantic and temporal violations.
}
\label{tab:comprehensive_stats}
\small
\renewcommand{\arraystretch}{1.2}
\begin{tabular}{l ccc c p{0.2cm} ccccc}
\toprule
\textbf{Benchmark / Subset} & \textbf{Train} & \textbf{Val} & \textbf{Test} & \textbf{Total} && \textbf{Mean} & \textbf{Med.} & \textbf{Std.} & \textbf{Min} & \textbf{Max} \\
\midrule
\multicolumn{11}{c}{\textit{\textsc{Prism} — Pointwise Evaluation (Likert 1--5)}} \\
\midrule
\textsc{Prism}-Pointwise (P1--P5) & 350 & 100 & 50 & 500 && 1761 & 1795 & 517 & 209 & 3291 \\
\midrule
\multicolumn{11}{c}{\textit{\textsc{Prism} — Pairwise Evaluation (A vs B)}} \\
\midrule
\textsc{Prism}-Semantic (Strong) & 1,625 & 450 & 235 & 2,310 && 1971 & 1903 & 686 & 493 & 4,570 \\
\textsc{Prism}-Temporal (Weak)   & 1,602 & 474 & 234 & 2,310 && 2073 & 2009 & 800 & 290 & 11,155 \\
\rowcolor{gray!10}
\textbf{Total \textsc{Prism} Pairwise} & \textbf{3,227} & \textbf{924} & \textbf{469} & \textbf{4,620} && \textbf{2022} & \textbf{1954} & \textbf{747} & \textbf{290} & \textbf{11,155} \\
\midrule
\multicolumn{11}{c}{\textit{\textsc{Mirage} — Pairwise Evaluation (A vs B)}} \\
\midrule
\textsc{Mirage}-Recipes (Procedural) & 230 & 64 & 28 & 322 && 2219 & 2108 & 701 & 827 & 4,850 \\
\textsc{Mirage}-Vist (Narrative)     & 239 & 69 & 42 & 350 && 2234 & 2096 & 720 & 761 & 4,514 \\
\rowcolor{gray!10}
\textbf{Total \textsc{Mirage} Pairwise} & \textbf{469} & \textbf{133} & \textbf{70} & \textbf{672} && \textbf{2227} & \textbf{2106} & \textbf{711} & \textbf{761} & \textbf{4,850} \\
\bottomrule
\end{tabular}
\end{table*}

\paragraph{Binary Collapse in Pointwise Scores}
As visually quantified in the main manuscript, the \textsc{Prism}-Pointwise subset exhibits a significant ``binary collapse'' in the zero-shot judge's distribution: 91\% of scores are concentrated in the $\{1, 2\}$ range, with only a marginal fraction (4.6\%) achieving a Score 5. This justifies our SFT objective to refine the model's ability to distinguish subtle quality tiers.

\paragraph{\textsc{Prism} Perturbation Strategy}
To assess specific reasoning capabilities, \textsc{Prism} sequences are generated using a multi-level perturbation strategy on a base set of 100 human-annotated cooking sequences (average length 4.7 frames). 
As detailed in Table~\ref{tab:prism_strategy}, \textbf{Semantic Negatives} (Type 1--3) involve replacing 1 to 3 frames with out-of-context images from different tasks to test domain consistency. \textbf{Temporal Negatives} involve local swaps (consecutive frames), global disruptions (non-consecutive swaps), or total shuffling to test causal and chronological reasoning.

\begin{table}[h]
\centering
\caption{\textbf{\textsc{Prism} Pairwise Construction.} Breakdown of perturbation types used to generate negative sequences from the base procedural dataset.}
\label{tab:prism_strategy}
\small
\begin{tabular}{@{}llcl@{}}
\toprule
\textbf{Category} & \textbf{Perturbation Type} & \textbf{Subtotal} & \textbf{Target Reasoning} \\
\midrule
\textbf{Semantic} & 1--3 Image Substitutions & 300 & Semantic Coherence \\
\midrule
\textbf{Temporal} & Consecutive Swap & 100 & Local Causal Order \\
                  & Non-consecutive Swap & 100 & Global Flow \\
                  & Shuffled Sequence & 100 & Absolute Chronology \\
\midrule
\textbf{Total} & \textbf{Unique Base Pairs} & \textbf{600} & \\
\bottomrule
\end{tabular}
\end{table}

\paragraph{Positional Bias Mitigation}
To ensure that models do not learn a positional bias (e.g., always preferring the first sequence), we applied a random shuffle to the order of sequences in each pairwise comparison. Table~\ref{tab:shuffle_stats} shows the distribution of shuffled vs. non-shuffled samples across \textsc{Prism} and \textsc{Mirage}. The balanced distribution (near 50\%) ensures that the LVLM judge must rely on content rather than sequence position.

\begin{table}[h]
\centering
\caption{\textbf{Pairwise Bias Mitigation.} Distribution of shuffled vs. non-shuffled samples to prevent positional bias during fine-tuning and evaluation.}
\label{tab:shuffle_stats}
\small
\begin{tabular}{@{}lcccc@{}}
\toprule
\textbf{Benchmark} & \textbf{Samples} & \textbf{Shuffled} & \textbf{Not Shuffled} & \textbf{Shuffle \%} \\
\midrule
\textsc{Mirage} & 672 & 357 & 315 & 53.1\% \\
\textsc{Prism}  & 4,620 & 2,317 & 2,303 & 50.1\% \\
\bottomrule
\end{tabular}
\end{table}

\paragraph{Reasoning Stability across Prompts}
Table~\ref{tab:per_prompt_stats} details the stability of the reasoning augmentation across different prompt layouts ($P_0$--$P_6$). The generation success rate remains near 100\% for all configurations, confirming that the SFT signal is robust to visual layout variations.

\subsection{Per-Prompt Reasoning Analysis}
To ensure the robustness of our SFT signal, we analyzed the stability of CoT generation across the different prompt configurations ($P_0$--$P_6$).

\begin{table*}[h]
\centering
\caption{\textbf{Reasoning Stability across Prompts.} Mean character length ($\pm$ std) and generation success rate for \textsc{Mirage} and \textsc{Prism} pairwise subsets.}
\label{tab:per_prompt_stats}
\footnotesize
\begin{tabular}{@{}llccccccc@{}}
\toprule
\textbf{Benchmark} & \textbf{Metric} & \textbf{P0} & \textbf{P1} & \textbf{P2} & \textbf{P3} & \textbf{P4} & \textbf{P5} & \textbf{P6} \\
\midrule
\textbf{\textsc{Mirage}} 
& Length & $2407 \pm 819$ & $2286 \pm 707$ & $2244 \pm 718$ & $2206 \pm 701$ & $2162 \pm 664$ & $2070 \pm 635$ & $2206 \pm 667$ \\
& Success & 100\% & 100\% & 100\% & 100\% & 100\% & 97\% & 97\% \\
\midrule
\textbf{\textsc{Prism}} 
& Length & $1849 \pm 706$ & $2127 \pm 740$ & $2169 \pm 803$ & $2172 \pm 747$ & $2116 \pm 751$ & $1959 \pm 753$ & $1760 \pm 600$ \\
& Success & 100\% & 100\% & 100\% & 100\% & 100\% & 99\% & 100\% \\
\bottomrule
\end{tabular}
\end{table*}

Table~\ref{tab:per_prompt_stats} shows that the reasoning length remains consistent regardless of the prompt layout, with a near-100\% success rate in rationale generation. The slight decrease in length for $P_5$ and $P_6$ (interleaved and grid layouts) suggests that high-density visual contexts lead to slightly more concise justifications compared to sequential layouts.

\section{Extended Experimental Results}
\label{sec:results}

The following tables extend the summary results of Tables~1--2 in the
main paper with a full per-prompt breakdown.  Results are reported
using deterministic inference (temperature\,=\,0,
\texttt{do\_sample=False}) throughout.  We use the following shorthand
for model variants consistently across all tables:

\begin{itemize}
  \item \textbf{\textsc{Critic}}: LLaVA-Critic 7B (zero-shot).
  \item \textbf{\textsc{OneVision}}: LLaVA-OneVision 7B (zero-shot).
  \item \textbf{OneVision$^\dagger$}: LLaVA-OneVision-SFT —
    single-prompt LoRA fine-tuned.
  \item \textbf{OneVision$^\ddagger$}: LLaVA-OneVision-SFT+ —
    multi-prompt LoRA fine-tuned (P$\star$).
  \item \textbf{OneVision$^\star$}: LoRA fine-tuned with CoT
    reasoning on \textsc{Prism} P$\star$.
\end{itemize}

\noindent The suffix \textbf{w/R} indicates that CoT rationales from
Gemini-2.5-Flash are used as an additional supervision signal
(Eq.~4 of the main paper).  NLP overlap metrics (BLEU, ROUGE, METEOR,
Cosine Similarity) are reported only when rationales are generated
(\cmark~in the \emph{Expl} column); they are computed against
Gemini-2.5-Flash reference explanations using sentence embeddings
from \texttt{all-MiniLM-L6-v2}.  Bold values indicate the best result
within each model block.

\subsection{Pointwise Evaluation on \textsc{Prism}}

Table~\ref{tab:supp_pointwise_prism_full} reports the full pointwise
results on \textsc{Prism} for zero-shot and LoRA fine-tuned variants.
The key takeaway is that prompt layout P3 and P4 consistently yield
stronger correlation with the Gemini oracle ($r_G$) across both
\textsc{Critic} and \textsc{OneVision}, while P5 tends to offer
the best human alignment ($r_H$) on \textsc{Critic}.  Fine-tuning
OneVision$^\dagger$ on the P5 reasoning split reduces MAE against the
Gemini oracle from 2.21 to 1.96, yet does not consistently improve
explanation quality, supporting the claim in Section~6.2 that
rationale fluency and score calibration are orthogonal axes.

\begin{table*}[htbp]
\centering
\caption{%
  \textbf{Full pointwise evaluation on \textsc{Prism} — zero-shot and
  LoRA fine-tuned.}
  NLP overlap metrics (BL, R$_1$/R$_2$/R$_L$, MTR, CS) are reported
  only when CoT rationales are generated (\cmark).
  $F_1$: macro score; $r_{G/H}$: Pearson correlation vs.\
  Gemini\,/\,Human; MAE/RMSE: error residuals;
  Len$_\Delta$: average explanation length difference vs.\ reference;
  $n_\text{val}$: valid predictions.
  $\dagger$\,OneVision LoRA SFT on \textsc{Prism}$_\text{w/R/P5}$
  (\texttt{r=16, $\alpha$=16, drop=0.03, lr=5e-6, maxlen=4096}). \textbf{Bold}: best value per column within each model block.
}
\label{tab:supp_pointwise_prism_full}
\small
\renewcommand{\arraystretch}{1.15}
\setlength{\tabcolsep}{3pt}
\resizebox{\textwidth}{!}{%
\begin{tabular}{@{} ll c  cccccc  cccc  cccc  rr @{}}
\toprule
& & &
\multicolumn{6}{c}{\textbf{Explanation Quality}} &
\multicolumn{4}{c}{\textbf{vs.\ Gemini}} &
\multicolumn{4}{c}{\textbf{vs.\ Human}} & & \\
\cmidrule(lr){4-9}\cmidrule(lr){10-13}\cmidrule(lr){14-17}
\textbf{Model} & \textbf{Prompt} & \textbf{Expl}
  & \textbf{BL} & \textbf{R$_1$} & \textbf{R$_2$} & \textbf{R$_L$}
  & \textbf{MTR} & \textbf{CS}
  & $\mathbf{F_1}$\,$\uparrow$ & $\mathbf{r_G}$\,$\uparrow$
  & \textbf{MAE}\,$\downarrow$ & \textbf{RMSE}\,$\downarrow$
  & $\mathbf{F_1}$\,$\uparrow$ & $\mathbf{r_H}$\,$\uparrow$
  & \textbf{MAE}\,$\downarrow$ & \textbf{RMSE}\,$\downarrow$
  & \textbf{Len$_\Delta$} & \textbf{$n_\text{val}$} \\
\midrule
\multirow{10}{*}{\textsc{Critic}}
& P1 & \cmark & 0.02 & 0.18 & 0.08 & 0.12 & 0.08 & 0.37 & 0.06 & 0.11 & 1.53 & 1.66 & 0.11 & 0.15 & 1.25 & 1.49 & 244 & 96 \\
& P2 & \cmark & 0.02 & 0.26 & 0.11 & 0.16 & 0.11 & 0.51 & 0.06 & 0.31 & 1.52 & 1.63 & 0.12 & 0.25 & 1.17 & 1.42 & 204 & 96 \\
& P3 & \cmark & 0.02 & 0.28 & 0.13 & 0.18 & 0.11 & 0.55 & \textbf{0.21} & 0.39 & \textbf{1.11} & \textbf{1.34} & \textbf{0.25} & 0.38 & 1.12 & 1.44 & 218 & 94 \\
& P4 & \cmark & \textbf{0.04} & \textbf{0.35} & \textbf{0.15} & \textbf{0.22} & \textbf{0.17} & \textbf{0.72} & 0.09 & \textbf{0.43} & 1.46 & 1.61 & 0.14 & \textbf{0.34} & 1.13 & 1.35 & -- & -- \\
& P5 & \cmark & 0.03 & \textbf{0.37} & \textbf{0.16} & \textbf{0.23} & 0.15 & \textbf{0.72} & 0.03 & 0.29 & 1.74 & 1.86 & 0.15 & \textbf{0.46} & \textbf{1.10} & \textbf{1.35} & 196 & 99 \\
\cmidrule(lr){2-18}
& P1 & \xmark & -- & -- & -- & -- & -- & -- & 0.14 & 0.13 & 1.16 & 1.58 & 0.12 & 0.18 & 1.56 & 1.90 & 964  & 100 \\
& P2 & \xmark & -- & -- & -- & -- & -- & -- & 0.13 & 0.22 & 1.31 & 1.63 & 0.12 & 0.19 & 1.45 & 1.76 & 824  & 100 \\
& P3 & \xmark & -- & -- & -- & -- & -- & -- & \textbf{0.24} & \textbf{0.48} & \textbf{0.93} & \textbf{1.22} & \textbf{0.20} & \textbf{0.42} & \textbf{1.19} & \textbf{1.54} & 818  & 100 \\
& P4 & \xmark & -- & -- & -- & -- & -- & -- & 0.18 & 0.47 & 1.21 & 1.49 & 0.18 & 0.33 & 1.21 & 1.48 & \textbf{1642} & 100 \\
& P5 & \xmark & -- & -- & -- & -- & -- & -- & 0.15 & 0.19 & 1.25 & 1.60 & 0.26 & 0.32 & 1.24 & 1.67 & 897  & 100 \\
\midrule
\multirow{10}{*}{\textsc{OneVision}}
& P1 & \cmark & 0.03 & 0.38 & 0.13 & 0.22 & 0.16 & \textbf{0.72} & 0.06 & 0.20 & 2.28 & 2.42 & 0.16 & 0.19 & 1.36 & 1.71 & 183 & 96 \\
& P2 & \cmark & 0.02 & 0.35 & 0.12 & 0.21 & 0.15 & 0.70 & \textbf{0.07} & 0.25 & 2.35 & 2.54 & \textbf{0.16} & 0.13 & 1.45 & 1.85 & 179 & 96 \\
& P3 & \cmark & \textbf{0.04} & \textbf{0.39} & \textbf{0.16} & \textbf{0.24} & \textbf{0.17} & \textbf{0.72} & 0.04 & \textbf{0.46} & 2.24 & 2.37 & \textbf{0.16} & \textbf{0.39} & 1.28 & \textbf{1.58} & 175 & 94 \\
& P4 & \cmark & 0.02 & 0.37 & 0.13 & 0.21 & 0.16 & \textbf{0.72} & 0.02 & 0.31 & 2.21 & 2.34 & 0.15 & 0.28 & 1.30 & 1.61 & 214 & 96 \\
& P5 & \cmark & 0.02 & 0.35 & 0.12 & 0.21 & 0.14 & 0.69 & 0.01 & 0.18 & \textbf{2.05} & \textbf{2.16} & \textbf{0.16} & 0.29 & \textbf{1.25} & \textbf{1.58} & 199 & 99 \\
\cmidrule(lr){2-18}
& P1 & \xmark & -- & -- & -- & -- & -- & -- & 0.05 & 0.25 & 2.41 & 2.57 & 0.17 & 0.17 & 1.44 & 1.83 & 468 & 100 \\
& P2 & \xmark & -- & -- & -- & -- & -- & -- & 0.03 & 0.20 & 2.75 & 2.95 & 0.14 & 0.04 & 1.71 & 2.18 & 269 & 100 \\
& P3 & \xmark & -- & -- & -- & -- & -- & -- & 0.01 & 0.36 & 2.66 & 2.80 & 0.16 & 0.23 & 1.51 & 1.94 & 327 & 100 \\
& P4 & \xmark & -- & -- & -- & -- & -- & -- & 0.04 & \textbf{0.41} & 2.32 & 2.47 & 0.18 & 0.28 & \textbf{1.33} & \textbf{1.71} & 684 & 100 \\
& P5 & \xmark & -- & -- & -- & -- & -- & -- & 0.02 & 0.21 & \textbf{2.16} & \textbf{2.28} & 0.20 & 0.26 & 1.30 & 1.67 & 555 & 100 \\
\midrule
\multirow{6}{*}{\textsc{OneVision}$^\dagger$}
& P1 & \cmark & 0.03 & 0.37 & 0.12 & 0.22 & 0.16 & 0.71 & 0.01 & 0.04 & 2.17 & 2.31 & 0.14 & 0.08 & 1.35 & 1.68 & 193 & 96 \\
& P2 & \cmark & 0.02 & 0.37 & 0.13 & 0.22 & 0.16 & 0.71 & \textbf{0.07} & 0.23 & 2.23 & 2.39 & \textbf{0.17} & 0.17 & 1.37 & 1.73 & 169 & 96 \\
& P3 & \cmark & \textbf{0.04} & 0.39 & \textbf{0.15} & \textbf{0.24} & \textbf{0.17} & 0.71 & 0.01 & \textbf{0.39} & 2.10 & 2.21 & 0.12 & 0.33 & 1.26 & 1.53 & 176 & 94 \\
& P4 & \cmark & \textbf{0.04} & \textbf{0.40} & \textbf{0.15} & \textbf{0.24} & \textbf{0.17} & \textbf{0.72} & 0.02 & 0.32 & \textbf{1.96} & \textbf{2.08} & 0.12 & 0.29 & \textbf{1.22} & \textbf{1.48} & 172 & 96 \\
& P5 & \cmark & 0.02 & 0.36 & 0.13 & 0.21 & 0.15 & \textbf{0.72} & 0.01 & 0.17 & 2.06 & 2.17 & \textbf{0.17} & \textbf{0.34} & 1.24 & 1.56 & 196 & 99 \\
\cmidrule(lr){2-18}
& \multicolumn{2}{l}{\textit{Average}} & 0.03 & 0.38 & 0.14 & 0.22 & 0.16 & 0.71 & 0.04 & 0.22 & 2.10 & 2.24 & 0.15 & 0.23 & 1.29 & 1.60 & 181 & 481 \\
\bottomrule
\end{tabular}%
}
\end{table*}

\subsection{Pairwise Evaluation on \textsc{Mirage}}

\begin{table*}[htbp]
\centering
\caption{%
  \textbf{Full zero-shot pairwise evaluation on \textsc{Mirage} (Table~S5).}
  Deterministic inference; $n_\text{val}$\,=\,valid predictions.
  Acc: accuracy; Pr/Rec/$F_1$: macro precision, recall, F-score.
  NLP metrics and Len$_\Delta$ reported only when rationales are generated.
}
\label{tab:supp_pairwise_mirage_zshot}
\scriptsize
\renewcommand{\arraystretch}{1.15}
\setlength{\tabcolsep}{3pt}
\resizebox{\textwidth}{!}{%
\begin{tabular}{@{} ll c  rr  cccc  cccccc  rr @{}}
\toprule
& & &
\multicolumn{2}{c}{\textbf{Length}} &
\multicolumn{4}{c}{\textbf{Pair Comparison}} &
\multicolumn{6}{c}{\textbf{Explanation Quality}} & & \\
\cmidrule(lr){4-5}\cmidrule(lr){6-9}\cmidrule(lr){10-15}
\textbf{Model} & \textbf{Prompt} & \textbf{Expl}
  & \textbf{Len$_\text{avg}$} & \textbf{Len$_\Delta$}
  & \textbf{Acc}\,$\uparrow$ & \textbf{Pr}\,$\uparrow$
  & \textbf{Rec}\,$\uparrow$ & $\mathbf{F_1}$\,$\uparrow$
  & \textbf{BL} & \textbf{R$_1$} & \textbf{R$_2$} & \textbf{R$_L$}
  & \textbf{MTR} & \textbf{CS}
  & \textbf{Len$_\Delta$} & \textbf{$n_\text{val}$} \\
\midrule
\multirow{14}{*}{\textsc{Critic}}
& P0 & \xmark & 971 & -- & 0.66 & 0.43 & 0.43 & 0.43 & -- & -- & -- & -- & -- & -- & -- & -- \\
& P1 & \xmark & 925 & -- & 0.60 & 0.43 & 0.37 & 0.33 & -- & -- & -- & -- & -- & -- & -- & -- \\
& P2 & \xmark & 807 & -- & 0.52 & 0.32 & 0.32 & 0.30 & -- & -- & -- & -- & -- & -- & -- & -- \\
& P3 & \xmark & 873 & -- & 0.55 & 0.38 & 0.37 & 0.37 & -- & -- & -- & -- & -- & -- & -- & -- \\
& P4 & \xmark & 951 & -- & 0.58 & 0.39 & 0.39 & 0.39 & -- & -- & -- & -- & -- & -- & -- & -- \\
& P5 & \xmark & 923 & -- & 0.54 & 0.60 & 0.36 & 0.45 & -- & -- & -- & -- & -- & -- & -- & -- \\
& P6 & \xmark & 825 & -- & \textbf{0.80} & \textbf{0.60} & \textbf{0.55} & \textbf{0.56} & -- & -- & -- & -- & -- & -- & -- & -- \\
\cmidrule(lr){2-17}
& P0 & \cmark & 831 & 252 & 0.58 & 0.52 & 0.37 & 0.31 & 0.05 & \textbf{0.42} & \textbf{0.22} & 0.27 & 0.18 & 0.81 & 252 & 96 \\
& P1 & \cmark & 833 & 231 & 0.53 & 0.18 & 0.33 & 0.23 & 0.05 & 0.40 & 0.19 & 0.26 & 0.17 & \textbf{0.82} & 231 & 96 \\
& P2 & \cmark & 799 & 226 & 0.59 & 0.48 & 0.38 & 0.33 & 0.04 & 0.39 & 0.17 & 0.24 & 0.17 & 0.80 & 226 & 96 \\
& P3 & \cmark & 765 & 231 & 0.61 & 0.43 & 0.40 & 0.38 & 0.04 & 0.39 & 0.18 & 0.25 & 0.17 & 0.80 & 231 & 96 \\
& P4 & \cmark & 874 & 207 & 0.58 & 0.39 & 0.38 & 0.36 & 0.05 & \textbf{0.42} & 0.19 & 0.26 & \textbf{0.19} & 0.81 & 207 & 96 \\
& P5 & \cmark & 797 & 201 & 0.67 & 0.59 & 0.44 & 0.50 & 0.04 & 0.40 & 0.17 & 0.25 & 0.18 & 0.79 & 201 & 93 \\
& P6 & \cmark & 838 & 223 & \textbf{0.94} & \textbf{0.63} & \textbf{0.62} & \textbf{0.62} & \textbf{0.06} & \textbf{0.42} & \textbf{0.22} & \textbf{0.29} & 0.18 & 0.80 & 223 & 93 \\
\midrule
\multirow{14}{*}{\textsc{OneVision}}
& P0 & \xmark & 14  & -- & 0.61 & 0.46 & 0.38 & 0.34 & -- & -- & -- & -- & -- & -- & -- & -- \\
& P1 & \xmark & 16  & -- & 0.62 & 0.53 & 0.38 & 0.33 & -- & -- & -- & -- & -- & -- & -- & -- \\
& P2 & \xmark & 15  & -- & 0.57 & 0.37 & 0.35 & 0.31 & -- & -- & -- & -- & -- & -- & -- & -- \\
& P3 & \xmark & 12  & -- & 0.68 & 0.47 & 0.43 & 0.43 & -- & -- & -- & -- & -- & -- & -- & -- \\
& P4 & \xmark & 15  & -- & 0.61 & 0.42 & 0.38 & 0.36 & -- & -- & -- & -- & -- & -- & -- & -- \\
& P5 & \xmark & 22  & -- & 0.78 & 0.52 & 0.52 & 0.52 & -- & -- & -- & -- & -- & -- & -- & -- \\
& P6 & \xmark & 19  & -- & \textbf{0.83} & \textbf{0.57} & \textbf{0.57} & \textbf{0.56} & -- & -- & -- & -- & -- & -- & -- & -- \\
\cmidrule(lr){2-17}
& P0 & \cmark & 686 & 274 & 0.57 & 0.46 & 0.36 & 0.30 & \textbf{0.02} & 0.34 & 0.14 & 0.21 & 0.13 & 0.76 & 274 & 96 \\
& P1 & \cmark & 586 & 270 & 0.55 & 0.51 & 0.35 & 0.26 & 0.01 & 0.29 & 0.10 & 0.18 & 0.11 & 0.76 & 270 & 96 \\
& P2 & \cmark & 581 & 261 & 0.57 & 0.46 & 0.36 & 0.30 & 0.01 & 0.29 & 0.10 & 0.18 & 0.11 & 0.76 & 261 & 96 \\
& P3 & \cmark & 555 & 261 & 0.67 & 0.47 & 0.43 & 0.42 & 0.01 & 0.28 & 0.10 & 0.18 & 0.11 & 0.74 & 261 & 96 \\
& P4 & \cmark & 648 & 243 & 0.52 & 0.32 & 0.33 & 0.26 & \textbf{0.02} & 0.31 & 0.11 & 0.19 & 0.13 & 0.76 & 243 & 96 \\
& P5 & \cmark & 657 & 229 & 0.78 & 0.52 & 0.52 & 0.52 & \textbf{0.02} & 0.33 & 0.13 & 0.21 & 0.14 & 0.76 & 229 & 93 \\
& P6 & \cmark & 645 & 251 & \textbf{0.80} & \textbf{0.55} & \textbf{0.54} & \textbf{0.53} & \textbf{0.02} & 0.33 & 0.14 & 0.21 & 0.14 & \textbf{0.77} & 251 & 93 \\
\bottomrule
\end{tabular}%
}
\end{table*}

\begin{table*}[htbp]
\centering
\caption{%
  \textbf{Full LoRA fine-tuned pairwise evaluation on \textsc{Mirage} (Table~S6).}
  Same inference setting as Table~\ref{tab:supp_pairwise_mirage_zshot}.
  $\dagger$\,=\,SFT on \textsc{Mirage} pairwise without reasoning, trained on P0;
  $\ddagger$\,=\,SFT on \textsc{Mirage} pairwise without reasoning, trained on P6;
  $\star$\,=\,SFT on \textsc{Mirage} pairwise with CoT reasoning, trained on P0.
}
\label{tab:supp_pairwise_mirage_lora}
\scriptsize
\renewcommand{\arraystretch}{1.15}
\setlength{\tabcolsep}{3pt}
\resizebox{\textwidth}{!}{%
\begin{tabular}{@{} ll c  rr  cccc  cccccc  rr @{}}
\toprule
& & &
\multicolumn{2}{c}{\textbf{Length}} &
\multicolumn{4}{c}{\textbf{Pair Comparison}} &
\multicolumn{6}{c}{\textbf{Explanation Quality}} & & \\
\cmidrule(lr){4-5}\cmidrule(lr){6-9}\cmidrule(lr){10-15}
\textbf{Model} & \textbf{Prompt} & \textbf{Expl}
  & \textbf{Len$_\text{avg}$} & \textbf{Len$_\Delta$}
  & \textbf{Acc}\,$\uparrow$ & \textbf{Pr}\,$\uparrow$
  & \textbf{Rec}\,$\uparrow$ & $\mathbf{F_1}$\,$\uparrow$
  & \textbf{BL} & \textbf{R$_1$} & \textbf{R$_2$} & \textbf{R$_L$}
  & \textbf{MTR} & \textbf{CS}
  & \textbf{Len$_\Delta$} & \textbf{$n_\text{val}$} \\
\midrule
\multirow{8}{*}{\textsc{OneVision}$^\dagger$}
& P0 & \xmark & 18 & -- & 0.71 & 0.47 & 0.46 & 0.46 & -- & -- & -- & -- & -- & -- & -- & -- \\
& P1 & \xmark & 18 & -- & 0.62 & 0.43 & 0.39 & 0.37 & -- & -- & -- & -- & -- & -- & -- & -- \\
& P2 & \xmark & 18 & -- & 0.70 & 0.46 & 0.46 & 0.46 & -- & -- & -- & -- & -- & -- & -- & -- \\
& P3 & \xmark & 18 & -- & 0.68 & 0.45 & 0.45 & 0.45 & -- & -- & -- & -- & -- & -- & -- & -- \\
& P4 & \xmark & 18 & -- & 0.53 & 0.34 & 0.34 & 0.33 & -- & -- & -- & -- & -- & -- & -- & -- \\
& P5 & \xmark & 55 & -- & 0.78 & 0.52 & 0.52 & 0.52 & -- & -- & -- & -- & -- & -- & -- & -- \\
& P6 & \xmark & 18 & -- & \textbf{0.89} & \textbf{0.59} & \textbf{0.59} & \textbf{0.59} & -- & -- & -- & -- & -- & -- & -- & -- \\
\cmidrule(lr){2-17}
& \multicolumn{2}{l}{\textit{Average}} & 23 & -- & 0.70 & 0.47 & 0.46 & 0.46 & -- & -- & -- & -- & -- & -- & -- & -- \\
\midrule
\multirow{8}{*}{\textsc{OneVision}$^\ddagger$}
& P0 & \xmark & 18 & -- & 0.68 & 0.45 & 0.45 & 0.45 & -- & -- & -- & -- & -- & -- & -- & -- \\
& P1 & \xmark & 19 & -- & 0.61 & 0.42 & 0.38 & 0.36 & -- & -- & -- & -- & -- & -- & -- & -- \\
& P2 & \xmark & 19 & -- & 0.68 & 0.45 & 0.44 & 0.44 & -- & -- & -- & -- & -- & -- & -- & -- \\
& P3 & \xmark & 18 & -- & 0.62 & 0.42 & 0.42 & 0.42 & -- & -- & -- & -- & -- & -- & -- & -- \\
& P4 & \xmark & 18 & -- & 0.51 & 0.32 & 0.33 & 0.32 & -- & -- & -- & -- & -- & -- & -- & -- \\
& P5 & \xmark & 19 & -- & 0.77 & 0.51 & 0.51 & 0.51 & -- & -- & -- & -- & -- & -- & -- & -- \\
& P6 & \xmark & 19 & -- & \textbf{0.90} & \textbf{0.60} & \textbf{0.60} & \textbf{0.60} & -- & -- & -- & -- & -- & -- & -- & -- \\
\cmidrule(lr){2-17}
& \multicolumn{2}{l}{\textit{Average}} & 19 & -- & 0.68 & 0.45 & 0.45 & 0.45 & -- & -- & -- & -- & -- & -- & -- & -- \\
\midrule
\multirow{8}{*}{\textsc{OneVision}$^\dagger$ (w/R)}
& P0 & \cmark & 906 & 255 & 0.74 & 0.51 & 0.49 & 0.49 & \textbf{0.04} & 0.39 & 0.18 & 0.25 & 0.17 & 0.78 & 255 & 96 \\
& P1 & \cmark & 858 & 242 & 0.61 & 0.46 & 0.40 & 0.36 & 0.03 & 0.35 & 0.14 & 0.22 & 0.14 & 0.78 & 242 & 96 \\
& P2 & \cmark & 779 & 244 & 0.71 & 0.48 & 0.47 & 0.47 & 0.03 & 0.34 & 0.13 & 0.21 & 0.14 & 0.77 & 244 & 96 \\
& P3 & \cmark & 734 & 246 & 0.72 & 0.48 & 0.48 & 0.48 & 0.02 & 0.32 & 0.12 & 0.20 & 0.13 & 0.75 & 246 & 96 \\
& P4 & \cmark & 849 & 232 & 0.62 & 0.42 & 0.41 & 0.40 & 0.03 & 0.35 & 0.14 & 0.22 & 0.15 & 0.77 & 232 & 96 \\
& P5 & \cmark & 794 & 221 & 0.77 & 0.52 & 0.51 & 0.51 & 0.02 & 0.34 & 0.13 & 0.22 & 0.14 & 0.77 & 221 & 93 \\
& P6 & \cmark & 811 & 237 & \textbf{0.84} & \textbf{0.57} & \textbf{0.57} & \textbf{0.56} & 0.03 & 0.37 & 0.16 & 0.24 & 0.15 & 0.78 & 237 & 93 \\
\cmidrule(lr){2-17}
& \multicolumn{2}{l}{\textit{Average}} & 819 & 239 & 0.72 & 0.48 & 0.47 & 0.47 & 0.03 & 0.35 & 0.14 & 0.22 & 0.14 & 0.77 & 239 & 666 \\
\midrule
\multirow{8}{*}{\textsc{OneVision}$^\ddagger$ (w/R)}
& P0 & \cmark & 637 & 282 & 0.74 & 0.51 & 0.49 & 0.49 & 0.02 & 0.33 & 0.14 & 0.21 & 0.13 & 0.77 & 282 & 96 \\
& P1 & \cmark & 593 & 267 & 0.64 & 0.47 & 0.41 & 0.39 & 0.01 & 0.30 & 0.11 & 0.19 & 0.11 & 0.77 & 267 & 96 \\
& P2 & \cmark & 604 & 258 & 0.74 & 0.50 & 0.49 & 0.49 & 0.01 & 0.30 & 0.10 & 0.18 & 0.12 & 0.76 & 258 & 96 \\
& P3 & \cmark & 578 & 261 & 0.66 & 0.44 & 0.44 & 0.44 & 0.01 & 0.28 & 0.10 & 0.17 & 0.10 & 0.73 & 261 & 96 \\
& P4 & \cmark & 584 & 250 & 0.62 & 0.42 & 0.41 & 0.41 & 0.02 & 0.31 & 0.12 & 0.20 & 0.12 & 0.77 & 250 & 96 \\
& P5 & \cmark & 599 & 232 & 0.76 & 0.51 & 0.51 & 0.51 & 0.02 & 0.31 & 0.11 & 0.20 & 0.13 & 0.75 & 232 & 93 \\
& P6 & \cmark & 616 & 254 & \textbf{0.84} & \textbf{0.57} & \textbf{0.57} & \textbf{0.56} & 0.02 & 0.33 & 0.14 & 0.21 & 0.13 & 0.77 & 254 & 93 \\
\cmidrule(lr){2-17}
& \multicolumn{2}{l}{\textit{Average}} & 602 & 258 & 0.71 & 0.48 & 0.47 & 0.47 & 0.02 & 0.31 & 0.12 & 0.19 & 0.12 & 0.76 & 258 & 666 \\
\midrule
\multirow{8}{*}{\textsc{OneVision}$^\star$ (w/R)}
& P0 & \cmark & 672 & 277 & 0.74 & 0.51 & 0.49 & 0.49 & 0.02 & 0.34 & 0.15 & 0.22 & 0.14 & 0.77 & 277 & 96 \\
& P1 & \cmark & 610 & 265 & 0.62 & 0.46 & 0.40 & 0.38 & 0.01 & 0.30 & 0.11 & 0.19 & 0.12 & 0.77 & 265 & 96 \\
& P2 & \cmark & 605 & 257 & 0.75 & 0.51 & 0.49 & 0.50 & 0.01 & 0.30 & 0.11 & 0.19 & 0.12 & 0.76 & 257 & 96 \\
& P3 & \cmark & 593 & 259 & 0.66 & 0.44 & 0.44 & 0.44 & 0.01 & 0.28 & 0.10 & 0.17 & 0.11 & 0.72 & 259 & 96 \\
& P4 & \cmark & 586 & 250 & 0.62 & 0.42 & 0.41 & 0.41 & 0.02 & 0.31 & 0.12 & 0.20 & 0.12 & 0.76 & 250 & 96 \\
& P5 & \cmark & 585 & 233 & 0.74 & 0.49 & 0.49 & 0.49 & 0.02 & 0.31 & 0.11 & 0.20 & 0.12 & 0.75 & 233 & 93 \\
& P6 & \cmark & 593 & 258 & \textbf{0.85} & \textbf{0.57} & \textbf{0.57} & \textbf{0.57} & 0.02 & 0.32 & 0.13 & 0.21 & 0.13 & 0.77 & 258 & 93 \\
\cmidrule(lr){2-17}
& \multicolumn{2}{l}{\textit{Average}} & 606 & 257 & 0.71 & 0.48 & 0.47 & 0.47 & 0.02 & 0.31 & 0.12 & 0.20 & 0.12 & 0.76 & 257 & 666 \\
\bottomrule
\end{tabular}%
}
\end{table*}

Tables~\ref{tab:supp_pairwise_mirage_zshot} and~\ref{tab:supp_pairwise_mirage_lora}
extend the \textsc{Mirage} pairwise results from the main paper.
Configuration~P6 emerges as the consistently best prompt layout for both
models and all fine-tuning variants, reaching Acc\,=\,0.94 for
\textsc{Critic} and Acc\,=\,0.90 for OneVision$^\dagger$ in the most
favourable settings.  The large response-length gap between
\textsc{Critic} ($\approx$825 tokens) and \textsc{OneVision}
($\approx$19 tokens) at equivalent $F_1$ confirms the verbosity bias
noted in Section~6.2 of the main paper.  Fine-tuning with CoT
reasoning (w/R suffix) does not consistently improve $F_1$ over
score-only fine-tuning on \textsc{Mirage}, suggesting that the
structured, procedural nature of this benchmark is already well-handled
by score-level supervision.

\begin{table*}[t]
\centering
\caption{\textbf{Full zero-shot pairwise evaluation on \textsc{Mirage}.}
Deterministic inference; $n_{\text{val}}$ = valid predictions.
Acc: accuracy; Pr/Rec/$F_1$: macro precision, recall, F-score.
NLP metrics and Len$_\Delta$ reported only when rationales are generated.
\textbf{Bold}: best value per column within each model block.}
\label{tab:supp_pairwise_mirage_zs}
\small
\setlength{\tabcolsep}{4pt}
\begin{tabular}{@{}ll c cc cccc cccccc cc@{}}
\toprule
& & & \multicolumn{2}{c}{\textbf{Length}} & \multicolumn{4}{c}{\textbf{Pair Comparison}} & \multicolumn{6}{c}{\textbf{Explanation Quality}} & & \\
\cmidrule(lr){4-5}\cmidrule(lr){6-9}\cmidrule(lr){10-15}
\textbf{Model} & \textbf{Prompt} & \textbf{Expl}
& \textbf{Len}$_{\text{avg}}$ & \textbf{Len}$_\Delta$
& \textbf{Acc} $\uparrow$ & \textbf{Pr} $\uparrow$ & \textbf{Rec} $\uparrow$ & $\mathbf{F_1}$ $\uparrow$
& \textbf{BL} & \textbf{R}$_1$ & \textbf{R}$_2$ & \textbf{R}$_L$ & \textbf{MTR} & \textbf{CS}
& \textbf{Len}$_\Delta$ & $\mathbf{n_{\text{val}}}$ \\
\midrule
\multirow{14}{*}{\textsc{Critic}}
& P0 & \xmark & 971 & -- & 0.66 & 0.43 & 0.43 & 0.43 & -- & -- & -- & -- & -- & -- & -- & -- \\
& P1 & \xmark & 925 & -- & 0.60 & 0.43 & 0.37 & 0.33 & -- & -- & -- & -- & -- & -- & -- & -- \\
& P2 & \xmark & 807 & -- & 0.52 & 0.32 & 0.32 & 0.30 & -- & -- & -- & -- & -- & -- & -- & -- \\
& P3 & \xmark & 873 & -- & 0.55 & 0.38 & 0.37 & 0.37 & -- & -- & -- & -- & -- & -- & -- & -- \\
& P4 & \xmark & 951 & -- & 0.58 & 0.39 & 0.39 & 0.39 & -- & -- & -- & -- & -- & -- & -- & -- \\
& P5 & \xmark & 923 & -- & 0.54 & \textbf{0.60} & 0.36 & 0.45 & -- & -- & -- & -- & -- & -- & -- & -- \\
& P6 & \xmark & 825 & -- & \textbf{0.80} & \textbf{0.60} & \textbf{0.55} & \textbf{0.56} & -- & -- & -- & -- & -- & -- & -- & -- \\
\cmidrule(lr){2-17}
& P0 & \cmark & 831 & 252 & 0.58 & 0.52 & 0.37 & 0.31 & 0.05 & \textbf{0.42} & \textbf{0.22} & 0.27 & 0.18 & 0.81 & 252 & 96 \\
& P1 & \cmark & 833 & 231 & 0.53 & 0.18 & 0.33 & 0.23 & 0.05 & 0.40 & 0.19 & 0.26 & 0.17 & \textbf{0.82} & 231 & 96 \\
& P2 & \cmark & 799 & 226 & 0.59 & 0.48 & 0.38 & 0.33 & 0.04 & 0.39 & 0.17 & 0.24 & 0.17 & 0.80 & 226 & 96 \\
& P3 & \cmark & 765 & 231 & 0.61 & 0.43 & 0.40 & 0.38 & 0.04 & 0.39 & 0.18 & 0.25 & 0.17 & 0.80 & 231 & 94 \\
& P4 & \cmark & 874 & 207 & 0.58 & 0.39 & 0.38 & 0.36 & 0.05 & \textbf{0.42} & 0.19 & 0.26 & \textbf{0.19} & 0.81 & 207 & 96 \\
& P5 & \cmark & 797 & 201 & 0.67 & 0.59 & 0.44 & 0.50 & 0.04 & 0.40 & 0.17 & 0.25 & 0.18 & 0.79 & 201 & 93 \\
& P6 & \cmark & 838 & 223 & \textbf{0.94} & \textbf{0.63} & \textbf{0.62} & \textbf{0.62} & \textbf{0.06} & \textbf{0.42} & \textbf{0.22} & \textbf{0.29} & 0.18 & 0.80 & 223 & 93 \\
\midrule
\multirow{14}{*}{\textsc{OneVision}}
& P0 & \xmark & 14 & -- & 0.61 & 0.46 & 0.38 & 0.34 & -- & -- & -- & -- & -- & -- & -- & -- \\
& P1 & \xmark & 16 & -- & 0.62 & 0.53 & 0.38 & 0.33 & -- & -- & -- & -- & -- & -- & -- & -- \\
& P2 & \xmark & 15 & -- & 0.57 & 0.37 & 0.35 & 0.31 & -- & -- & -- & -- & -- & -- & -- & -- \\
& P3 & \xmark & 12 & -- & 0.68 & 0.47 & 0.43 & 0.43 & -- & -- & -- & -- & -- & -- & -- & -- \\
& P4 & \xmark & 15 & -- & 0.61 & 0.42 & 0.38 & 0.36 & -- & -- & -- & -- & -- & -- & -- & -- \\
& P5 & \xmark & 22 & -- & 0.78 & 0.52 & 0.52 & 0.52 & -- & -- & -- & -- & -- & -- & -- & -- \\
& P6 & \xmark & 19 & -- & \textbf{0.83} & \textbf{0.57} & \textbf{0.57} & \textbf{0.56} & -- & -- & -- & -- & -- & -- & -- & -- \\
\cmidrule(lr){2-17}
& P0 & \cmark & 686 & 274 & 0.57 & 0.46 & 0.36 & 0.30 & \textbf{0.02} & \textbf{0.34} & \textbf{0.14} & \textbf{0.21} & 0.13 & 0.76 & 274 & 96 \\
& P1 & \cmark & 586 & 270 & 0.55 & 0.51 & 0.35 & 0.26 & 0.01 & 0.29 & 0.10 & 0.18 & 0.11 & 0.76 & 270 & 96 \\
& P2 & \cmark & 581 & 261 & 0.57 & 0.46 & 0.36 & 0.30 & 0.01 & 0.29 & 0.10 & 0.18 & 0.11 & 0.76 & 261 & 96 \\
& P3 & \cmark & 555 & 261 & 0.67 & 0.47 & 0.43 & 0.42 & 0.01 & 0.28 & 0.10 & 0.18 & 0.11 & 0.74 & 261 & 94 \\
& P4 & \cmark & 648 & 243 & 0.52 & 0.32 & 0.35 & 0.26 & \textbf{0.02} & 0.31 & 0.11 & 0.19 & 0.13 & 0.76 & 243 & 96 \\
& P5 & \cmark & 657 & 229 & 0.78 & 0.52 & 0.52 & 0.52 & \textbf{0.02} & 0.33 & 0.13 & \textbf{0.21} & \textbf{0.14} & 0.76 & 229 & 96 \\
& P6 & \cmark & 645 & 251 & \textbf{0.80} & \textbf{0.55} & \textbf{0.54} & \textbf{0.53} & \textbf{0.02} & 0.33 & \textbf{0.14} & \textbf{0.21} & \textbf{0.14} & \textbf{0.77} & 251 & 93 \\
\bottomrule
\end{tabular}
\end{table*}

\begin{table*}[t]
\centering
\caption{\textbf{Full pointwise evaluation on \textsc{Prism}: zero-shot and LoRA fine-tuned.}
NLP overlap metrics (BL, R$_1$/R$_2$/R$_L$, MTR, CS) are reported only when CoT rationales
are generated (\cmark). $F_1$: macro score; $r_{G/H}$: Pearson correlation vs.\ Gemini / Human;
MAE/RMSE: error residuals; Len$_\Delta$: average explanation length difference vs.\ reference;
$n_{\text{val}}$: valid predictions. \textbf{Bold}: best value per column within each model block.
$\dagger$ OneVision LoRA SFT on \textsc{Prism}$_{\text{w/R/P5}}$
(\texttt{r=16}, $\alpha$\texttt{=16}, \texttt{drop=0.03}, \texttt{lr=5e-6}, \texttt{maxlen=4096}).}
\label{tab:supp_pointwise_full}
\small
\setlength{\tabcolsep}{4pt}
\begin{tabular}{@{}ll c cccccc cccc cccc cc@{}}
\toprule
& & & \multicolumn{6}{c}{\textbf{Explanation Quality}} & \multicolumn{4}{c}{\textbf{vs. Gemini}} & \multicolumn{4}{c}{\textbf{vs. Human}} & & \\
\cmidrule(lr){4-9}\cmidrule(lr){10-13}\cmidrule(lr){14-17}
\textbf{Model} & \textbf{Prompt} & \textbf{Expl}
& \textbf{BL} & \textbf{R}$_1$ & \textbf{R}$_2$ & \textbf{R}$_L$ & \textbf{MTR} & \textbf{CS}
& $\mathbf{F_1}$ $\uparrow$ & $\mathbf{r_G}$ $\uparrow$ & \textbf{MAE} $\downarrow$ & \textbf{RMSE} $\downarrow$
& $\mathbf{F_1}$ $\uparrow$ & $\mathbf{r_H}$ $\uparrow$ & \textbf{MAE} $\downarrow$ & \textbf{RMSE} $\downarrow$
& \textbf{Len}$_\Delta$ & $\mathbf{n_{\text{val}}}$ \\
\midrule
\multirow{10}{*}{\textsc{Critic}}
& P1 & \cmark & 0.02 & 0.18 & 0.08 & 0.12 & 0.08 & 0.37 & 0.06 & 0.11 & 1.53 & 1.66 & 0.11 & 0.15 & 1.25 & 1.49 & 244 & 96 \\
& P2 & \cmark & 0.02 & 0.26 & 0.11 & 0.16 & 0.11 & 0.51 & 0.06 & 0.31 & 1.52 & 1.63 & 0.12 & 0.25 & 1.17 & 1.42 & 204 & 96 \\
& P3 & \cmark & 0.02 & 0.28 & 0.13 & 0.18 & 0.11 & 0.55 & \textbf{0.21} & 0.39 & \textbf{1.11} & \textbf{1.34} & \textbf{0.25} & 0.38 & \textbf{1.12} & 1.44 & 218 & 94 \\
& P4 & \cmark & \textbf{0.04} & 0.35 & \textbf{0.16} & 0.22 & \textbf{0.17} & \textbf{0.72} & 0.09 & \textbf{0.43} & 1.46 & 1.61 & 0.14 & 0.34 & 1.13 & \textbf{1.35} & -- & -- \\
& P5 & \cmark & 0.03 & \textbf{0.37} & \textbf{0.16} & \textbf{0.23} & 0.15 & \textbf{0.72} & 0.03 & 0.29 & 1.74 & 1.86 & 0.15 & \textbf{0.46} & \textbf{1.10} & \textbf{1.35} & 196 & 99 \\
\cmidrule(lr){2-19}
& P1 & \xmark & -- & -- & -- & -- & -- & -- & 0.14 & 0.13 & 1.16 & 1.58 & 0.12 & 0.18 & 1.56 & 1.90 & 964 & 100 \\
& P2 & \xmark & -- & -- & -- & -- & -- & -- & 0.13 & 0.22 & 1.31 & 1.63 & 0.12 & 0.19 & 1.45 & 1.76 & 824 & 100 \\
& P3 & \xmark & -- & -- & -- & -- & -- & -- & \textbf{0.24} & \textbf{0.48} & \textbf{0.93} & \textbf{1.22} & \textbf{0.20} & \textbf{0.42} & 1.19 & \textbf{1.54} & 818 & 100 \\
& P4 & \xmark & -- & -- & -- & -- & -- & -- & 0.18 & 0.47 & 1.21 & 1.49 & 0.18 & 0.33 & 1.21 & 1.48 & 1642 & 100 \\
& P5 & \xmark & -- & -- & -- & -- & -- & -- & 0.15 & 0.19 & 1.25 & 1.60 & 0.26 & 0.32 & 1.24 & 1.67 & 897 & 100 \\
\midrule
\multirow{10}{*}{\textsc{OneVision}}
& P1 & \cmark & 0.03 & 0.38 & 0.13 & 0.22 & 0.16 & \textbf{0.72} & 0.06 & 0.20 & 2.28 & 2.42 & \textbf{0.16} & 0.19 & 1.36 & 1.71 & 183 & 96 \\
& P2 & \cmark & 0.02 & 0.35 & 0.12 & 0.21 & 0.15 & 0.70 & \textbf{0.07} & 0.23 & 2.35 & 2.54 & \textbf{0.16} & 0.13 & 1.45 & 1.85 & 179 & 96 \\
& P3 & \cmark & \textbf{0.04} & \textbf{0.39} & \textbf{0.16} & \textbf{0.24} & \textbf{0.17} & \textbf{0.72} & 0.04 & \textbf{0.46} & 2.24 & 2.37 & \textbf{0.16} & \textbf{0.39} & 1.28 & \textbf{1.58} & 175 & 94 \\
& P4 & \cmark & 0.02 & 0.37 & 0.13 & 0.21 & 0.16 & \textbf{0.72} & 0.02 & 0.31 & 2.21 & 2.34 & 0.15 & 0.28 & 1.30 & 1.61 & 214 & 96 \\
& P5 & \cmark & 0.02 & 0.35 & 0.12 & 0.21 & 0.14 & 0.69 & 0.01 & 0.18 & \textbf{2.05} & \textbf{2.16} & \textbf{0.16} & 0.29 & \textbf{1.25} & \textbf{1.58} & 199 & 99 \\
\cmidrule(lr){2-19}
& P1 & \xmark & -- & -- & -- & -- & -- & -- & 0.05 & 0.29 & 2.41 & 2.57 & 0.17 & 0.17 & 1.44 & 1.83 & 468 & 100 \\
& P2 & \xmark & -- & -- & -- & -- & -- & -- & 0.03 & 0.20 & 2.75 & 2.95 & 0.14 & 0.04 & 1.71 & 2.18 & 269 & 100 \\
& P3 & \xmark & -- & -- & -- & -- & -- & -- & 0.01 & 0.36 & 2.66 & 2.80 & 0.16 & 0.23 & 1.51 & 1.94 & 327 & 100 \\
& P4 & \xmark & -- & -- & -- & -- & -- & -- & 0.04 & \textbf{0.41} & 2.32 & 2.47 & 0.18 & \textbf{0.28} & \textbf{1.33} & \textbf{1.71} & 684 & 100 \\
& P5 & \xmark & -- & -- & -- & -- & -- & -- & 0.02 & 0.21 & \textbf{2.16} & \textbf{2.28} & \textbf{0.20} & 0.26 & 1.30 & 1.67 & 555 & 100 \\
\midrule
\multirow{5}{*}{\textsc{OneVision}$^{\dagger}$}
& P1 & \cmark & 0.03 & 0.37 & 0.12 & 0.22 & 0.16 & 0.71 & 0.01 & 0.04 & 2.17 & 2.31 & 0.14 & 0.08 & 1.35 & 1.68 & 193 & 96 \\
& P2 & \cmark & 0.02 & 0.37 & 0.13 & 0.22 & 0.16 & 0.71 & \textbf{0.07} & 0.23 & 2.23 & 2.39 & \textbf{0.17} & 0.11 & 1.37 & 1.73 & 169 & 96 \\
& P3 & \cmark & \textbf{0.04} & 0.39 & \textbf{0.15} & \textbf{0.24} & \textbf{0.17} & 0.71 & 0.01 & \textbf{0.39} & 2.10 & 2.21 & 0.12 & 0.33 & 1.26 & 1.53 & 176 & 94 \\
& P4 & \cmark & \textbf{0.04} & \textbf{0.40} & \textbf{0.15} & \textbf{0.24} & \textbf{0.17} & \textbf{0.72} & 0.02 & 0.32 & \textbf{1.96} & \textbf{2.08} & 0.11 & 0.29 & \textbf{1.22} & \textbf{1.48} & 172 & 96 \\
& P5 & \cmark & 0.02 & 0.36 & 0.13 & 0.21 & 0.15 & \textbf{0.72} & 0.01 & 0.17 & 2.06 & 2.17 & \textbf{0.17} & \textbf{0.34} & 1.24 & 1.56 & 196 & 99 \\
\midrule
\multicolumn{3}{l}{\textit{Average}}
& 0.03 & 0.38 & 0.14 & 0.22 & 0.16 & 0.71 & 0.04 & 0.22 & 2.10 & 2.24 & 0.15 & 0.23 & 1.29 & 1.60 & 181 & 481 \\
\bottomrule
\end{tabular}
\end{table*}

\subsection{Pairwise Evaluation on \textsc{Prism}}

Tables~\ref{tab:supp_pairwise_prism_zshot_noexpl}–\ref{tab:supp_pairwise_prism_lora_b}
report pairwise results on \textsc{Prism}, broken down by prompt,
perturbation type (\textsc{Semantic} vs.\ \textsc{Temporal}), and
fine-tuning variant.

A consistent pattern emerges across all settings: \textsc{Prism}-Semantic
is substantially easier than \textsc{Prism}-Temporal.  The best
zero-shot accuracy on the semantic subset reaches 0.72 (OneVision,
P6 with CoT), while temporal swap detection peaks at 0.56 even after
fine-tuning, confirming the structural fragility predicted by the
lost-in-the-middle analysis in Section~3 of the main paper.

\begin{table*}[htbp]
\centering
\caption{%
  \textbf{Zero-shot pairwise on \textsc{Prism} — score only.}
  Acc: accuracy; Pr: precision; $F_1$: macro F-score. Bold = best per-model.
}
\label{tab:supp_pairwise_prism_zshot_noexpl}
\scriptsize
\renewcommand{\arraystretch}{0.85}
\setlength{\tabcolsep}{2pt}
\resizebox{0.72\textwidth}{!}{%
\begin{tabular}{@{} l l r  rrrr  rrrr @{}}
\toprule
& & &
\multicolumn{4}{c}{\textbf{Critic}} &
\multicolumn{4}{c}{\textbf{OneVision}} \\
\cmidrule(lr){4-7}\cmidrule(lr){8-11}
\textbf{Sub.} & \textbf{P} & \textbf{$n$}
  & \textbf{Len} & \textbf{Acc} & \textbf{Pr} & $\mathbf{F_1}$
  & \textbf{Len} & \textbf{Acc} & \textbf{Pr} & $\mathbf{F_1}$ \\
\midrule
\multirow{8}{*}{\rotatebox{90}{\textit{All}}}
& P0 & 600 & 813 & .50 & .34 & .31 & 15 & .28 & .28 & .20 \\
& P1 & 600 & 941 & .52 & .36 & .31 & 17 & .43 & .26 & .22 \\
& P2 & 600 & 824 & .41 & .30 & .27 & 13 & .47 & .30 & .26 \\
& P3 & 600 & 852 & .42 & .31 & .29 & 10 & .51 & .34 & .29 \\
& P4 & 600 & \textbf{1112} & .44 & .31 & .30 & 16 & .29 & .27 & .21 \\
& P5 & 600 & 912 & .20 & .47 & .20 & 19 & .57 & .40 & .37 \\
& P6 & 600 & 859 & .43 & .50 & .36 & \textbf{19} & \textbf{.64} & \textbf{.44} & \textbf{.42} \\
\midrule
& \textit{Avg} & 4200 & 902 & .42 & .35 & .30 & 15 & .45 & .35 & .31 \\
\midrule
\multirow{8}{*}{\rotatebox{90}{\textit{Sem.}}}
& P0 & 300 &  820 & .54 & .38 & .35 & 15 & .31 & .36 & .23 \\
& P1 & 300 &  966 & .57 & .41 & .35 & 16 & .44 & .30 & .23 \\
& P2 & 300 &  825 & .44 & .33 & .30 & 12 & .51 & .38 & .29 \\
& P3 & 300 &  851 & .48 & .35 & .33 &  9 & .53 & .37 & .31 \\
& P4 & 300 & \textbf{1147} & .53 & .36 & .35 & 16 & .32 & .32 & .23 \\
& P5 & 300 &  921 & .35 & .48 & .32 & 19 & .61 & .42 & .40 \\
& P6 & 300 &  823 & \textbf{.66} & \textbf{.53} & \textbf{.48} & \textbf{19} & \textbf{.72} & \textbf{.48} & \textbf{.48} \\
\midrule
& \textit{Avg} & 2100 & 908 & .51 & .40 & .36 & 15 & .49 & .39 & .34 \\
\midrule
\multirow{8}{*}{\rotatebox{90}{\textit{Temp.}}}
& P0 & 300 &  806 & .46 & .28 & .26 & 15 & .24 & .21 & .17 \\
& P1 & 300 &  915 & .47 & .30 & .26 & 17 & .42 & .24 & .22 \\
& P2 & 300 &  823 & .39 & .26 & .24 & 13 & .43 & .24 & .23 \\
& P3 & 300 &  853 & .35 & .26 & .24 & 10 & .48 & .30 & .27 \\
& P4 & 300 & 1078 & .36 & .23 & .22 & 16 & .26 & .22 & .18 \\
& P5 & 300 &  903 & .04 & .20 & .04 & 19 & .53 & .37 & .35 \\
& P6 & 300 &  895 & .19 & .44 & .19 & \textbf{19} & \textbf{.55} & \textbf{.39} & \textbf{.35} \\
\midrule
& \textit{Avg} & 2100 & 896 & .32 & .28 & .23 & 16 & .42 & .31 & .29 \\
\bottomrule
\end{tabular}}
\end{table*}

\begin{table*}[htbp]
\centering
\caption{%
  \textbf{Zero-shot pairwise on \textsc{Prism} — CoT rationales,
  classification metrics.}
  Same setting as Table~\ref{tab:supp_pairwise_prism_zshot_noexpl}.
  See Table~\ref{tab:supp_pairwise_prism_zshot_expl_nlp} for NLP metrics.
}
\label{tab:supp_pairwise_prism_zshot_expl_cls}
\scriptsize
\renewcommand{\arraystretch}{0.85}
\setlength{\tabcolsep}{2pt}
\resizebox{0.65\textwidth}{!}{%
\begin{tabular}{@{} l l  rrrr  rrrr @{}}
\toprule
& &
\multicolumn{4}{c}{\textbf{Critic}} &
\multicolumn{4}{c}{\textbf{OneVision}} \\
\cmidrule(lr){3-6}\cmidrule(lr){7-10}
\textbf{Sub.} & \textbf{P}
  & \textbf{Acc} & \textbf{Pr} & \textbf{Rec} & $\mathbf{F_1}$
  & \textbf{Acc} & \textbf{Pr} & \textbf{Rec} & $\mathbf{F_1}$ \\
\midrule
\multirow{8}{*}{\rotatebox{90}{\textit{All}}}
& P0 & .50 & .32 & .33 & .26 & .47 & .26 & .32 & .23 \\
& P1 & .49 & .30 & .32 & .24 & .47 & .18 & .31 & .22 \\
& P2 & .46 & .28 & .31 & .25 & .46 & .25 & .30 & .24 \\
& P3 & .49 & .32 & .33 & .30 & .47 & .30 & .31 & .28 \\
& P4 & .47 & .30 & .31 & .26 & .44 & .25 & .29 & .24 \\
& P5 & .35 & .49 & .24 & .29 & .60 & .41 & .40 & .39 \\
& P6 & \textbf{.63} & \textbf{.47} & \textbf{.42} & \textbf{.44} & \textbf{.65} & \textbf{.46} & \textbf{.43} & \textbf{.42} \\
\midrule
& \textit{Avg} & .48 & .37 & .32 & .30 & .51 & .34 & .34 & .32 \\
\midrule
\multirow{8}{*}{\rotatebox{90}{\textit{Sem.}}}
& P0 & .52 & .39 & .35 & .29 & .50 & .38 & .34 & .26 \\
& P1 & .49 & .34 & .33 & .26 & .48 & .22 & .32 & .22 \\
& P2 & .47 & .30 & .32 & .26 & .46 & .28 & .32 & .24 \\
& P3 & .51 & .35 & .34 & .33 & .52 & .36 & .35 & .32 \\
& P4 & .47 & .32 & .32 & .28 & .46 & .30 & .31 & .27 \\
& P5 & .52 & .51 & .35 & .39 & .64 & .44 & .43 & .42 \\
& P6 & \textbf{.77} & \textbf{.53} & \textbf{.52} & \textbf{.52} & \textbf{.74} & \textbf{.50} & \textbf{.49} & \textbf{.49} \\
\midrule
& \textit{Avg} & .54 & .40 & .36 & .34 & .54 & .38 & .37 & .35 \\
\midrule
\multirow{8}{*}{\rotatebox{90}{\textit{Temp.}}}
& P0 & .47 & .21 & .30 & .22 & .45 & .16 & .29 & .21 \\
& P1 & .48 & .17 & .31 & .22 & .47 & .16 & .30 & .21 \\
& P2 & .46 & .24 & .30 & .23 & .45 & .23 & .29 & .23 \\
& P3 & .47 & .29 & .31 & .27 & .42 & .24 & .27 & .24 \\
& P4 & .46 & .21 & .30 & .22 & .42 & .17 & .27 & .20 \\
& P5 & .19 & .41 & .12 & .16 & .55 & .38 & .37 & \textbf{.35} \\
& P6 & .49 & .40 & .32 & .34 & \textbf{.56} & \textbf{.41} & \textbf{.38} & \textbf{.35} \\
\midrule
& \textit{Avg} & .43 & .30 & .28 & .25 & .47 & .31 & .31 & .30 \\
\bottomrule
\end{tabular}}
\end{table*}

\begin{table*}[htbp]
\centering
\caption{%
  \textbf{Zero-shot pairwise on \textsc{Prism} — with CoT rationales,
  NLP overlap metrics.}
  Metrics computed against Gemini-2.5-Flash reference rationales.
  BL\,=\,BLEU; R$_1$/R$_2$/R$_L$\,=\,ROUGE-1/2/L; MTR\,=\,METEOR;
  CS\,=\,Cosine Similarity; Len$_\Delta$\,=\,response length
  difference vs.\ reference.
}
\label{tab:supp_pairwise_prism_zshot_expl_nlp}
\scriptsize
\renewcommand{\arraystretch}{1.15}
\setlength{\tabcolsep}{3pt}
\resizebox{\textwidth}{!}{%
\begin{tabular}{@{} l l
  r ccccc c r
  r ccccc c r
@{}}
\toprule
& &
\multicolumn{8}{c}{\textbf{\textsc{LLaVA-Critic}}} &
\multicolumn{8}{c}{\textbf{\textsc{LLaVA-OneVision}}} \\
\cmidrule(lr){3-10}\cmidrule(lr){11-18}
\textbf{Subset} & \textbf{Prompt}
  & \textbf{Len} & \textbf{BL} & \textbf{R$_1$} & \textbf{R$_2$}
  & \textbf{R$_L$} & \textbf{MTR} & \textbf{CS} & \textbf{Len$_\Delta$}
  & \textbf{Len} & \textbf{BL} & \textbf{R$_1$} & \textbf{R$_2$}
  & \textbf{R$_L$} & \textbf{MTR} & \textbf{CS} & \textbf{Len$_\Delta$} \\
\midrule
\multirow{8}{*}{\textit{All}}
& P0 & 796 & 0.07 & \textbf{0.46} & 0.22 & 0.29 & 0.20 & \textbf{0.84} & 174 & 653 & 0.04 & 0.37 & 0.15 & 0.23 & 0.16 & 0.79 & 199 \\
& P1 & 860 & 0.06 & 0.43 & 0.19 & 0.27 & 0.19 & 0.82 & 206 & 538 & 0.01 & 0.30 & 0.11 & 0.19 & 0.11 & 0.76 & 256 \\
& P2 & 824 & 0.05 & 0.41 & 0.18 & 0.25 & 0.18 & 0.81 & 218 & 566 & 0.01 & 0.30 & 0.10 & 0.18 & 0.12 & 0.76 & 255 \\
& P3 & 804 & 0.05 & 0.41 & 0.19 & 0.26 & 0.17 & 0.81 & 225 & 555 & 0.01 & 0.29 & 0.10 & 0.18 & 0.11 & 0.75 & 261 \\
& P4 & 828 & 0.05 & 0.41 & 0.18 & 0.25 & 0.18 & 0.81 & 211 & 555 & 0.02 & 0.31 & 0.11 & 0.19 & 0.12 & 0.77 & 250 \\
& P5 & 956 & 0.07 & 0.44 & 0.20 & 0.27 & 0.20 & 0.81 & 167 & 614 & 0.03 & 0.35 & 0.13 & 0.22 & 0.14 & 0.79 & 217 \\
& P6 & 832 & 0.07 & 0.44 & 0.21 & 0.27 & 0.20 & 0.81 & 164 & 635 & \textbf{0.04} & \textbf{0.39} & \textbf{0.16} & \textbf{0.24} & \textbf{0.17} & \textbf{0.80} & 182 \\
\cmidrule(lr){2-18}
& \textit{Avg} & 843 & 0.06 & 0.43 & 0.19 & 0.27 & 0.19 & 0.81 & 195 & 588 & 0.02 & 0.33 & 0.13 & 0.21 & 0.13 & 0.78 & 232 \\
\midrule
\multirow{8}{*}{\textit{\textsc{Prism}-Semantic}}
& P0 & 798 & 0.06 & 0.45 & 0.21 & 0.28 & 0.20 & \textbf{0.84} & 177 & 672 & 0.03 & 0.37 & 0.15 & 0.23 & 0.15 & 0.79 & 203 \\
& P1 & 875 & 0.06 & 0.44 & 0.20 & 0.28 & 0.20 & 0.82 & 191 & 560 & 0.02 & 0.31 & 0.11 & 0.19 & 0.12 & 0.77 & 239 \\
& P2 & 842 & 0.05 & 0.42 & 0.19 & 0.26 & 0.19 & 0.82 & 199 & 583 & 0.02 & 0.31 & 0.11 & 0.19 & 0.13 & 0.77 & 235 \\
& P3 & 813 & 0.05 & 0.41 & 0.19 & 0.26 & 0.18 & 0.81 & 216 & 573 & 0.02 & 0.30 & 0.11 & 0.19 & 0.12 & 0.76 & 251 \\
& P4 & 827 & 0.05 & 0.41 & 0.18 & 0.25 & 0.18 & 0.81 & 203 & 553 & 0.02 & 0.31 & 0.11 & 0.20 & 0.12 & 0.78 & 241 \\
& P5 & 899 & 0.06 & 0.43 & 0.19 & 0.26 & 0.19 & 0.81 & 174 & 609 & 0.02 & 0.35 & 0.13 & 0.21 & 0.14 & 0.79 & 215 \\
& P6 & 823 & 0.07 & 0.44 & 0.22 & 0.28 & 0.20 & 0.81 & 167 & 626 & 0.03 & \textbf{0.37} & \textbf{0.15} & \textbf{0.23} & \textbf{0.16} & \textbf{0.80} & 187 \\
\cmidrule(lr){2-18}
& \textit{Avg} & 839 & 0.06 & 0.43 & 0.20 & 0.27 & 0.19 & 0.82 & 190 & 596 & 0.02 & 0.33 & 0.13 & 0.21 & 0.13 & 0.78 & 225 \\
\midrule
\multirow{8}{*}{\textit{\textsc{Prism}-Temporal}}
& P0 & 794 & \textbf{0.08} & \textbf{0.46} & \textbf{0.23} & \textbf{0.30} & \textbf{0.21} & \textbf{0.84} & 171 & 634 & 0.04 & 0.37 & 0.16 & 0.24 & 0.16 & 0.80 & 195 \\
& P1 & 846 & 0.05 & 0.42 & 0.18 & 0.26 & 0.18 & 0.81 & 220 & 517 & 0.01 & 0.28 & 0.10 & 0.18 & 0.10 & 0.75 & 273 \\
& P2 & 806 & 0.04 & 0.39 & 0.18 & 0.24 & 0.17 & 0.79 & 237 & 549 & 0.01 & 0.28 & 0.10 & 0.17 & 0.11 & 0.74 & 275 \\
& P3 & 795 & 0.04 & 0.40 & 0.18 & 0.25 & 0.17 & 0.81 & 234 & 536 & 0.01 & 0.28 & 0.10 & 0.17 & 0.11 & 0.75 & 271 \\
& P4 & 829 & 0.05 & 0.40 & 0.17 & 0.25 & 0.17 & 0.80 & 219 & 557 & 0.01 & 0.30 & 0.11 & 0.19 & 0.12 & 0.76 & 259 \\
& P5 & 1014 & 0.07 & 0.46 & 0.21 & 0.28 & 0.21 & 0.81 & 160 & 619 & 0.03 & 0.35 & 0.14 & 0.22 & 0.14 & 0.79 & 219 \\
& P6 & 841 & 0.06 & 0.43 & 0.19 & 0.26 & 0.20 & 0.80 & 161 & 645 & \textbf{0.05} & \textbf{0.40} & \textbf{0.18} & \textbf{0.26} & \textbf{0.18} & \textbf{0.81} & 178 \\
\cmidrule(lr){2-18}
& \textit{Avg} & 846 & 0.06 & 0.42 & 0.19 & 0.26 & 0.19 & 0.81 & 200 & 580 & 0.02 & 0.32 & 0.13 & 0.20 & 0.13 & 0.77 & 238 \\
\bottomrule
\end{tabular}%
}
\end{table*}

\begin{table*}[htbp]
\centering
\caption{%
  \textbf{LoRA fine-tuned pairwise on \textsc{Prism}, part~1/2.}
  Three OneVision fine-tuning variants:
  $\dagger$\,=\,P6, no reasoning (\texttt{r=16, $\alpha$=64});
  $\ddagger$\,=\,P0, with reasoning (\texttt{r=128, $\alpha$=32});
  $\P$\,=\,P$\star$, no reasoning (\texttt{r=32, $\alpha$=128}).
  See Table~\ref{tab:supp_pairwise_prism_lora_b} for $\S$ and $\star$.
  The markers used here and in Table~\ref{tab:supp_pairwise_prism_lora_b}
  index the five training configurations of this ablation and are
  independent of the $\dagger$/$\ddagger$ notation of the main paper.
  \textbf{Bold}: best value per column within each model block.
}
\label{tab:supp_pairwise_prism_lora_a}
\scriptsize
\renewcommand{\arraystretch}{1.15}
\setlength{\tabcolsep}{3pt}
\resizebox{\textwidth}{!}{%
\begin{tabular}{@{} ll c  r  cccc  cccccc  r @{}}
\toprule
& & & &
\multicolumn{4}{c}{\textbf{Pair Comparison}} &
\multicolumn{6}{c}{\textbf{Explanation Quality}} & \\
\cmidrule(lr){5-8}\cmidrule(lr){9-14}
\textbf{Model} & \textbf{Prompt} & \textbf{Expl} & \textbf{Len}
  & \textbf{Acc}\,$\uparrow$ & \textbf{Pr}\,$\uparrow$
  & \textbf{Rec}\,$\uparrow$ & $\mathbf{F_1}$\,$\uparrow$
  & \textbf{BL} & \textbf{R$_1$} & \textbf{R$_2$} & \textbf{R$_L$}
  & \textbf{MTR} & \textbf{CS} & \textbf{Len$_\Delta$} \\
\midrule
\multirow{8}{*}{\textsc{OneVision}$^\dagger$}
& P0 & \xmark & 18 & 0.49 & 0.32 & 0.33 & 0.31 & -- & -- & -- & -- & -- & -- & -- \\
& P1 & \xmark & 18 & 0.51 & 0.33 & 0.33 & 0.30 & -- & -- & -- & -- & -- & -- & -- \\
& P2 & \xmark & 18 & 0.48 & 0.32 & 0.32 & 0.32 & -- & -- & -- & -- & -- & -- & -- \\
& P3 & \xmark & 18 & 0.48 & 0.32 & 0.32 & 0.32 & -- & -- & -- & -- & -- & -- & -- \\
& P4 & \xmark & 18 & 0.45 & 0.29 & 0.30 & 0.29 & -- & -- & -- & -- & -- & -- & -- \\
& P5 & \xmark & 18 & 0.58 & 0.40 & 0.39 & 0.38 & -- & -- & -- & -- & -- & -- & -- \\
& P6 & \xmark & 18 & \textbf{0.67} & \textbf{0.49} & \textbf{0.45} & \textbf{0.43} & -- & -- & -- & -- & -- & -- & -- \\
\cmidrule(lr){2-15}
& \textit{Avg} & \xmark & 18 & 0.52 & 0.35 & 0.35 & 0.35 & -- & -- & -- & -- & -- & -- & -- \\
\midrule
\multirow{8}{*}{\textsc{OneVision}$^\ddagger$}
& P0 & \cmark & 593 & 0.45 & 0.29 & 0.30 & 0.28 & 0.03 & 0.37 & 0.16 & 0.23 & 0.15 & 0.80 & 204 \\
& P1 & \cmark & 610 & 0.50 & 0.33 & 0.33 & 0.29 & 0.02 & 0.33 & 0.13 & 0.20 & 0.13 & 0.78 & 244 \\
& P2 & \cmark & 595 & 0.46 & 0.31 & 0.31 & 0.31 & 0.02 & 0.31 & 0.12 & 0.19 & 0.12 & 0.77 & 251 \\
& P3 & \cmark & 563 & 0.46 & 0.31 & 0.31 & 0.31 & 0.02 & 0.30 & 0.11 & 0.18 & 0.12 & 0.75 & 259 \\
& P4 & \cmark & 552 & 0.42 & 0.27 & 0.28 & 0.27 & 0.02 & 0.32 & 0.13 & 0.20 & 0.13 & 0.78 & 250 \\
& P5 & \cmark & 546 & 0.57 & 0.40 & 0.38 & 0.36 & 0.02 & 0.32 & 0.12 & 0.20 & 0.13 & 0.78 & 224 \\
& P6 & \cmark & 612 & \textbf{0.67} & \textbf{0.49} & \textbf{0.45} & \textbf{0.43} & \textbf{0.04} & \textbf{0.39} & \textbf{0.17} & \textbf{0.25} & \textbf{0.17} & \textbf{0.80} & 184 \\
\cmidrule(lr){2-15}
& \textit{Avg} & \cmark & 581 & 0.51 & 0.34 & 0.34 & 0.34 & 0.02 & 0.33 & 0.13 & 0.21 & 0.14 & 0.78 & 231 \\
\midrule
\multirow{8}{*}{\textsc{OneVision}$^\P$}
& P0 & \xmark & 18 & 0.52 & 0.35 & 0.35 & 0.33 & -- & -- & -- & -- & -- & -- & -- \\
& P1 & \xmark & 18 & 0.54 & 0.37 & 0.35 & 0.31 & -- & -- & -- & -- & -- & -- & -- \\
& P2 & \xmark & 18 & 0.50 & 0.33 & 0.33 & 0.33 & -- & -- & -- & -- & -- & -- & -- \\
& P3 & \xmark & 18 & 0.50 & 0.33 & 0.33 & 0.33 & -- & -- & -- & -- & -- & -- & -- \\
& P4 & \xmark & 18 & 0.50 & 0.33 & 0.33 & 0.30 & -- & -- & -- & -- & -- & -- & -- \\
& P5 & \xmark & 18 & 0.62 & 0.41 & 0.41 & 0.41 & -- & -- & -- & -- & -- & -- & -- \\
& P6 & \xmark & 18 & \textbf{0.70} & \textbf{0.50} & \textbf{0.47} & \textbf{0.46} & -- & -- & -- & -- & -- & -- & -- \\
\cmidrule(lr){2-15}
& \textit{Avg} & \xmark & 18 & 0.55 & 0.37 & 0.37 & 0.37 & -- & -- & -- & -- & -- & -- & -- \\
\bottomrule
\end{tabular}%
}
\end{table*}

\begin{table*}[htbp]
\centering
\caption{%
  \textbf{LoRA fine-tuned pairwise on \textsc{Prism}, part~2/2.}
  Continuation of Table~\ref{tab:supp_pairwise_prism_lora_a}.
  $\S$\,=\,P6, with reasoning (\texttt{r=128, $\alpha$=32});
  $\star$\,=\,P$\star$, with reasoning (\texttt{r=32, $\alpha$=128}).
  Markers index the training configurations of this ablation and are
  independent of the $\dagger$/$\ddagger$ notation of the main paper.
  \textbf{Bold}: best value per column within each model block.
}
\label{tab:supp_pairwise_prism_lora_b}
\scriptsize
\renewcommand{\arraystretch}{1.15}
\setlength{\tabcolsep}{3pt}
\resizebox{\textwidth}{!}{%
\begin{tabular}{@{} ll c  r  cccc  cccccc  r @{}}
\toprule
& & & &
\multicolumn{4}{c}{\textbf{Pair Comparison}} &
\multicolumn{6}{c}{\textbf{Explanation Quality}} & \\
\cmidrule(lr){5-8}\cmidrule(lr){9-14}
\textbf{Model} & \textbf{Prompt} & \textbf{Expl} & \textbf{Len}
  & \textbf{Acc}\,$\uparrow$ & \textbf{Pr}\,$\uparrow$
  & \textbf{Rec}\,$\uparrow$ & $\mathbf{F_1}$\,$\uparrow$
  & \textbf{BL} & \textbf{R$_1$} & \textbf{R$_2$} & \textbf{R$_L$}
  & \textbf{MTR} & \textbf{CS} & \textbf{Len$_\Delta$} \\
\midrule
\multirow{8}{*}{\textsc{OneVision}$^\S$}
& P0 & \cmark & 595 & 0.46 & 0.29 & 0.30 & 0.28 & 0.04 & 0.37 & 0.16 & 0.23 & 0.16 & 0.80 & 204 \\
& P1 & \cmark & 613 & 0.50 & 0.33 & 0.33 & 0.29 & 0.02 & 0.33 & 0.13 & 0.20 & 0.13 & 0.78 & 243 \\
& P2 & \cmark & 588 & 0.46 & 0.30 & 0.30 & 0.30 & 0.02 & 0.31 & 0.12 & 0.19 & 0.12 & 0.76 & 252 \\
& P3 & \cmark & 558 & 0.45 & 0.30 & 0.30 & 0.30 & 0.01 & 0.30 & 0.11 & 0.18 & 0.11 & 0.75 & 259 \\
& P4 & \cmark & 553 & 0.42 & 0.27 & 0.28 & 0.27 & 0.02 & 0.32 & 0.13 & 0.20 & 0.13 & 0.78 & 250 \\
& P5 & \cmark & 549 & 0.57 & 0.40 & 0.38 & 0.36 & 0.02 & 0.33 & 0.12 & 0.20 & 0.13 & 0.78 & 224 \\
& P6 & \cmark & 618 & \textbf{0.67} & \textbf{0.49} & \textbf{0.45} & \textbf{0.43} & \textbf{0.04} & \textbf{0.39} & \textbf{0.17} & \textbf{0.25} & \textbf{0.17} & \textbf{0.80} & 184 \\
\cmidrule(lr){2-15}
& \textit{Avg} & \cmark & 582 & 0.50 & 0.34 & 0.34 & 0.34 & 0.02 & 0.33 & 0.13 & 0.21 & 0.14 & 0.78 & 231 \\
\midrule
\multirow{8}{*}{\textsc{OneVision}$^\star$}
& P0 & \cmark & 645 & 0.47 & 0.30 & 0.31 & 0.29 & 0.04 & 0.40 & 0.18 & 0.25 & 0.17 & 0.81 & 195 \\
& P1 & \cmark & 647 & 0.51 & 0.35 & 0.34 & 0.30 & 0.03 & 0.35 & 0.14 & 0.21 & 0.14 & 0.79 & 238 \\
& P2 & \cmark & 624 & 0.47 & 0.31 & 0.31 & 0.31 & 0.02 & 0.33 & 0.13 & 0.20 & 0.13 & 0.77 & 246 \\
& P3 & \cmark & 616 & 0.47 & 0.31 & 0.31 & 0.31 & 0.02 & 0.32 & 0.13 & 0.20 & 0.13 & 0.77 & 251 \\
& P4 & \cmark & 570 & 0.43 & 0.28 & 0.29 & 0.27 & 0.02 & 0.33 & 0.13 & 0.21 & 0.13 & 0.78 & 247 \\
& P5 & \cmark & 569 & 0.59 & 0.41 & 0.39 & 0.38 & 0.02 & 0.33 & 0.13 & 0.21 & 0.14 & 0.78 & 220 \\
& P6 & \cmark & 633 & \textbf{0.68} & \textbf{0.50} & \textbf{0.46} & \textbf{0.44} & \textbf{0.04} & \textbf{0.40} & \textbf{0.17} & \textbf{0.25} & \textbf{0.17} & \textbf{0.81} & 181 \\
\cmidrule(lr){2-15}
& \textit{Avg} & \cmark & 615 & 0.52 & 0.35 & 0.35 & 0.34 & 0.03 & 0.35 & 0.14 & 0.22 & 0.14 & 0.79 & 225 \\
\bottomrule
\end{tabular}%
}
\end{table*}

\section{Positional Sensitivity: 7B Judges}
\label{sec:positional}

Section~6.3 of the main paper reports opposite positional profiles on
the two \textsc{Prism} subsets: a primacy peak on
\textsc{Prism}-Semantic, where detection is easiest when the outlier
falls in the opening frame, and a recency pattern on
\textsc{Prism}-Temporal, where swaps become progressively easier toward
the end of the sequence.  Here we provide the full per-position
accuracy and $F_1$ breakdown that underlies Figure~4 of the main paper,
together with additional heatmap visualisations.

\paragraph{Semantic subset.}  Models exhibit a clear \emph{primacy
bias}: performance is highest when the semantic outlier appears at
position~0 and drops to a flat regime close to chance for all later
positions.  This is consistent with the theoretical prediction in
Section~3 that causal masking causes hidden representations to
converge toward the initial token as depth increases.  The best
fine-tuned variant (OneVision$^\ddagger$, P$\star$) reaches
Acc\,=\,0.70 / $F_1$\,=\,0.69 at \texttt{pos0} but falls to
0.51--0.55 at all interior positions.

\paragraph{Temporal subset.}  Temporal swap detection follows a
\emph{recency pattern}: swaps at later positions (adjacent to the
final frame) are marginally easier to detect, while middle-sequence
swaps remain near chance.  Even the best model ($W_\text{avg}$\,=\,0.47)
does not exceed chance level overall, confirming the lost-in-the-middle
structural bottleneck.

\begin{table*}[t]
\centering
\caption{%
  \textbf{Full positional sensitivity on \textsc{Prism} ($n=4{,}620$).}
  Accuracy / macro $F_1$ per position.
  $W_\text{avg}$: sample-weighted mean;
  $\dagger$~positions with $n=140$.
  \textbf{(a)}~Model comparison averaged over P0--P6.
  \textbf{(b)}~Detailed prompt breakdown for the best fine-tuned
  configuration (OneVision$^\ddagger$, P$\star$).
}
\label{tab:positional_sensitivity_full}
\renewcommand{\arraystretch}{1.6}
\setlength{\tabcolsep}{2.5pt}

\noindent\textbf{(a)} Model comparison, averaged over P0--P6.\\[3pt]
\resizebox{\textwidth}{!}{%
\begin{tabular}{@{}l cccccc c @{\hspace{8pt}} ccccc c@{}}
\toprule
& \multicolumn{7}{c}{\textbf{\textsc{Semantic}} (Acc\,$\uparrow$ / $F_1$\,$\uparrow$)}
& \multicolumn{6}{c}{\textbf{\textsc{Temporal}} (Acc\,$\uparrow$ / $F_1$\,$\uparrow$)} \\
\cmidrule(lr){2-8}\cmidrule(lr){9-14}
\textbf{Model} & \textbf{p0} & \textbf{p1} & \textbf{p2} & \textbf{p3}
  & \textbf{p4} & \textbf{p5}$^\dagger$ & $\mathbf{W_\text{avg}}$
  & \textbf{0-1} & \textbf{1-2} & \textbf{2-3} & \textbf{3-4}
  & \textbf{4-5}$^\dagger$ & $\mathbf{W_\text{avg}}$ \\
\midrule
OneVision \cite{li2024llavaonevision}
  & .53/.49 & .52/.49 & .51/.48 & .50/.48 & .49/.45 & .59/.52 & .52/.48
  & .45/.44 & .49/.47 & .49/.47 & .50/.46 & .54/.49 & .49/.46 \\
Critic \cite{xiong2025llavacriticlearningevaluatemultimodal}
  & .54/.54 & .47/.49 & .48/.49 & .45/.46 & .48/.48 & .45/.41 & .48/.49
  & .42/.44 & .41/.43 & .46/.47 & .45/.44 & .43/.41 & .43/.44 \\
\midrule
OneVision$^\dagger$ \textsc{Prism}$_\text{P6}$
  & .65/.65 & .52/.52 & .50/.50 & .51/.51 & .52/.52 & .54/.52 & .54/.54
  & .40/.39 & .45/.44 & .49/.49 & .50/.49 & .56/.55 & .46/.46 \\
OneVision$^\dagger$ \textsc{Prism}$_\text{w/R/P6}$
  & .63/.63 & .52/.52 & .49/.49 & .51/.51 & .53/.53 & .51/.49 & .54/.54
  & .41/.39 & .44/.44 & .50/.50 & .47/.47 & .54/.54 & .46/.45 \\
OneVision$^\ddagger$ \textsc{Prism}$_\text{w/R/P}\star$
  & .64/.64 & .51/.51 & .50/.50 & .52/.52 & .54/.53 & .51/.50 & .54/.54
  & .40/.39 & .44/.43 & .51/.51 & .48/.48 & .54/.53 & .46/.45 \\
OneVision$^\ddagger$ \textsc{Prism}$_\text{P}\star$
  & \textbf{.70/.69} & \textbf{.53/.52} & \textbf{.55/.53} & \textbf{.51/.50}
  & \textbf{.53/.52} & \textbf{.55/.50} & \textbf{.57/.55}
  & .41/.41 & \textbf{.46/.44} & \textbf{.49/.48} & \textbf{.51/.49}
  & .53/.51 & .47/.45 \\
\bottomrule
\end{tabular}}

\vspace{8pt}

\noindent\textbf{(b)} Detailed prompt breakdown for OneVision$^\ddagger$ \textsc{Prism}$_\text{P}\star$.\\[3pt]
\resizebox{\textwidth}{!}{%
\begin{tabular}{@{}l cccccc c @{\hspace{8pt}} ccccc c@{}}
\toprule
& \multicolumn{7}{c}{\textbf{\textsc{Semantic}} (Acc\,$\uparrow$ / $F_1$\,$\uparrow$)}
& \multicolumn{6}{c}{\textbf{\textsc{Temporal}} (Acc\,$\uparrow$ / $F_1$\,$\uparrow$)} \\
\cmidrule(lr){2-8}\cmidrule(lr){9-14}
\textbf{Prompt} & \textbf{p0} & \textbf{p1} & \textbf{p2} & \textbf{p3}
  & \textbf{p4} & \textbf{p5}$^\dagger$ & $\mathbf{W_\text{avg}}$
  & \textbf{0-1} & \textbf{1-2} & \textbf{2-3} & \textbf{3-4}
  & \textbf{4-5}$^\dagger$ & $\mathbf{W_\text{avg}}$ \\
\midrule
P0 & .82/.81 & .46/.45 & .50/.44 & .53/.48 & .55/.46 & .65/.39 & \textbf{.58/.53}
   & .41/.40 & .38/.33 & .48/.41 & .55/.44 & .50/.33 & .45/.38 \\
P1 & .71/.68 & .46/.38 & .49/.36 & .46/.31 & .50/.33 & .65/.39 & .53/.42
   & .42/.40 & .51/.36 & .48/.32 & .52/.34 & .55/.36 & .48/.36 \\
P2 & .73/.73 & .49/.45 & .48/.44 & .48/.43 & .55/.47 & .60/.47 & .55/.51
   & .34/.33 & .43/.36 & .43/.36 & .50/.38 & .60/.52 & .43/.36 \\
P3 & .71/.71 & .46/.42 & .47/.39 & .43/.35 & .53/.44 & .60/.52 & .53/.47
   & .40/.38 & .48/.39 & .48/.40 & .50/.38 & .60/.52 & .47/.40 \\
P4 & .66/.64 & .43/.39 & .51/.43 & .48/.37 & .58/.50 & .65/.39 & .53/.46
   & .30/.25 & .43/.34 & .49/.38 & .55/.44 & .50/.33 & .43/.34 \\
P5 & .57/.57 & .67/.67 & .65/.64 & .59/.53 & .48/.39 & .35/.31 & .59/.56
   & .47/.41 & .50/.46 & .52/.44 & .48/.41 & .50/.41 & .50/.43 \\
P6 & .67/.67 & .76/.75 & .72/.71 & .61/.55 & .53/.46 & .35/.26 & \textbf{.66/.63}
   & .53/.41 & .48/.36 & .53/.37 & .48/.35 & .45/.31 & .51/.37 \\
\bottomrule
\end{tabular}}
\end{table*}

\begin{figure*}[t]
\centering
\begin{subfigure}[t]{0.48\linewidth}
  \centering
  \includegraphics[width=\linewidth]{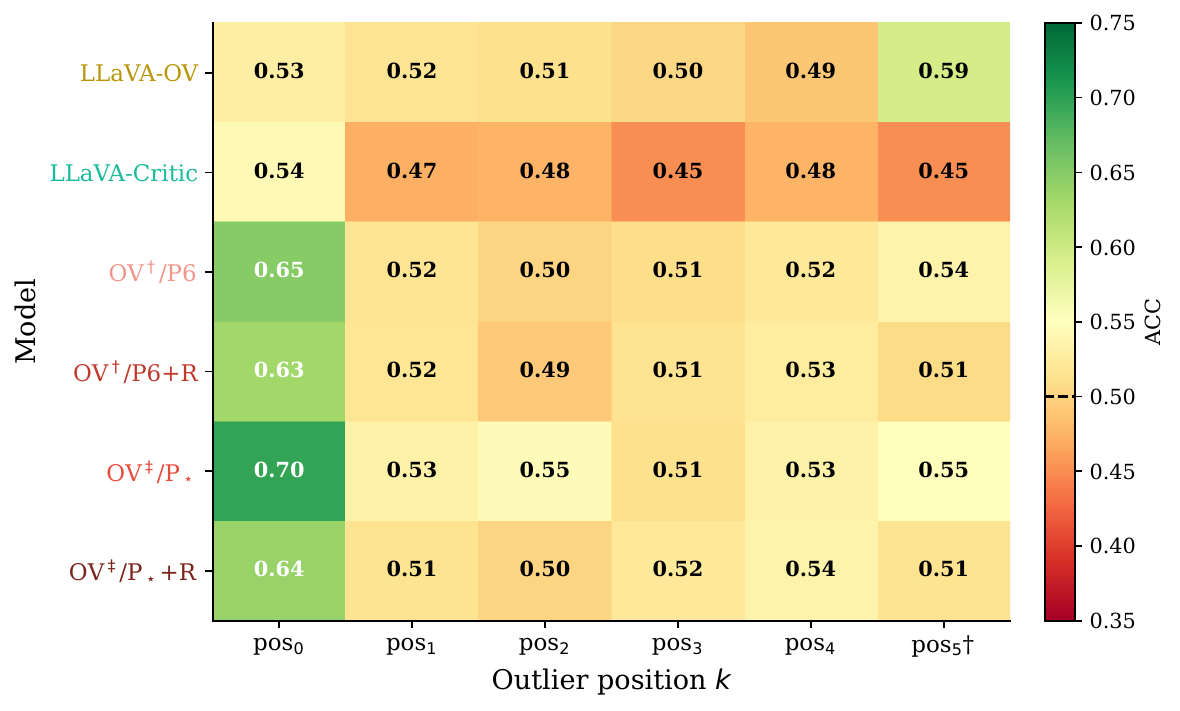}
  \caption{Semantic outlier position heatmap.}
\end{subfigure}
\hfill
\begin{subfigure}[t]{0.48\linewidth}
  \centering
  \includegraphics[width=\linewidth]{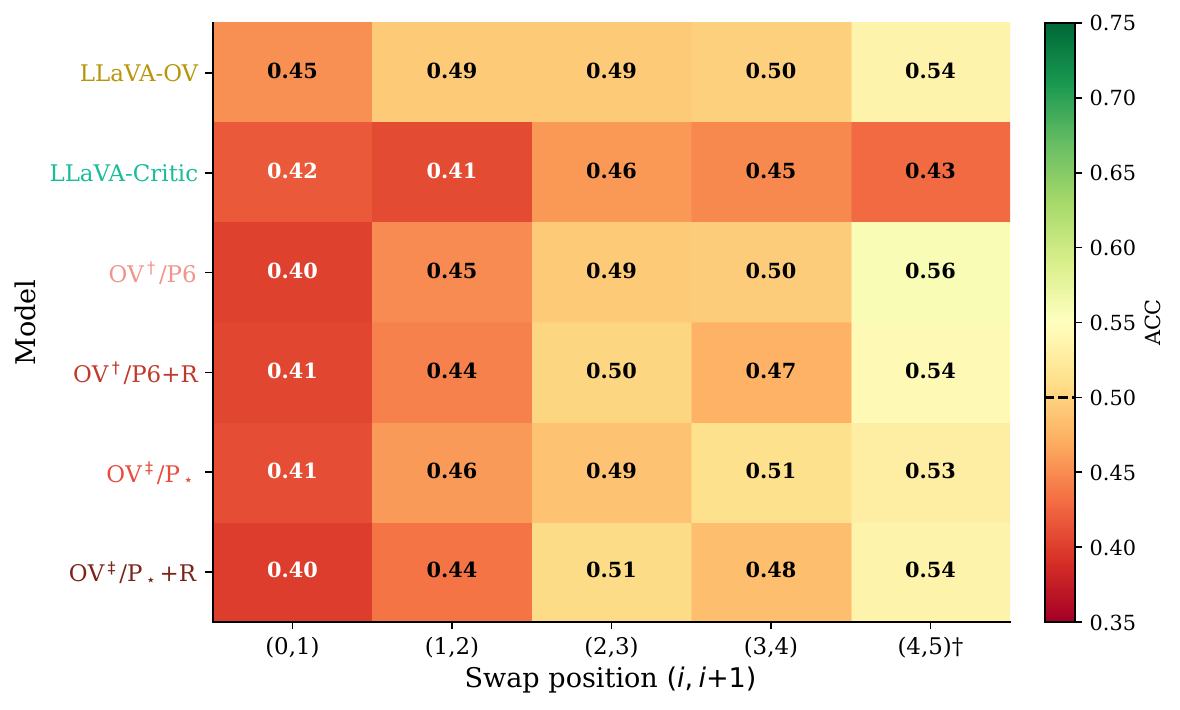}
  \caption{Temporal swap position heatmap.}
\end{subfigure}
\vspace{6pt}
\begin{subfigure}[t]{0.48\linewidth}
  \centering
  \includegraphics[width=0.85\linewidth]{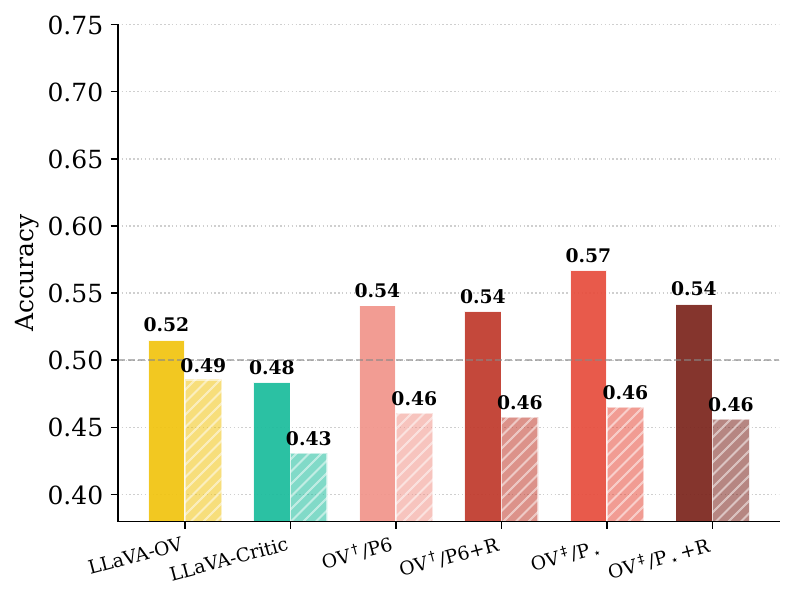}
  \caption{Sample-weighted mean accuracy ($W_\text{avg}$ Acc).}
\end{subfigure}
\hfill
\begin{subfigure}[t]{0.48\linewidth}
  \centering
  \includegraphics[width=0.85\linewidth]{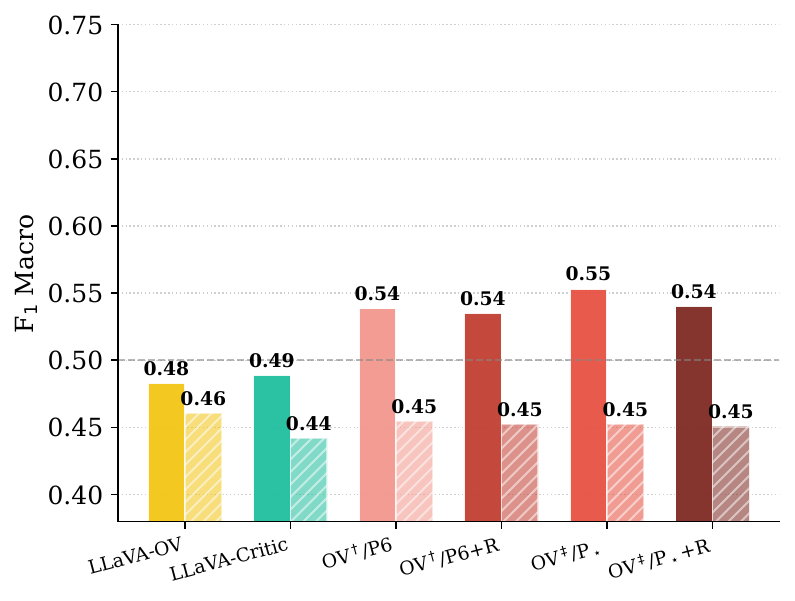}
  \caption{Sample-weighted mean $F_1$ ($W_\text{avg}$ $F_1$).}
\end{subfigure}
\vspace{4pt}
\includegraphics[width=0.7\linewidth]{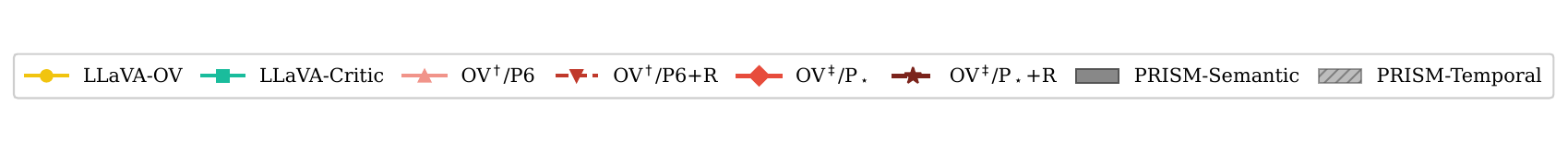}
\caption{%
  \textbf{Systematic positional sensitivity analysis.}
  Per-position accuracy heatmaps for the Semantic (a) and Temporal (b)
  subsets, and sample-weighted mean Accuracy (c) and $F_1$ (d) across
  all model variants.  The semantic heatmap exhibits a primacy peak
  at~\texttt{pos0} followed by a flat regime; the temporal heatmap
  shows detection improving toward the end of the sequence, leaving
  earlier and mid-sequence swaps hardest to identify.  No fine-tuned
  configuration escapes the positional floor.
}
\label{fig:positional_sensitivity_heatmap}
\end{figure*}

Figure~\ref{fig:positional_sensitivity_heatmap} presents the
corresponding heatmaps and weighted-average bar charts.  The heatmaps
make the two profiles directly visible: a single bright cell at the
opening position of the Semantic subset, and a gradient toward the end
of the sequence in the Temporal one.  The bar charts show that no
fine-tuning configuration overcomes the structural bottleneck imposed
by positional embeddings, supporting the conclusion in the main paper
that these asymmetries require architectural remediation beyond simple
data scaling.

\begin{figure*}[t]
\centering
\includegraphics[width=0.85\linewidth]{img/fig_pos_legend_fix.pdf}\\[4pt]
\begin{subfigure}[b]{0.245\linewidth}
    \centering
    \includegraphics[width=\linewidth]{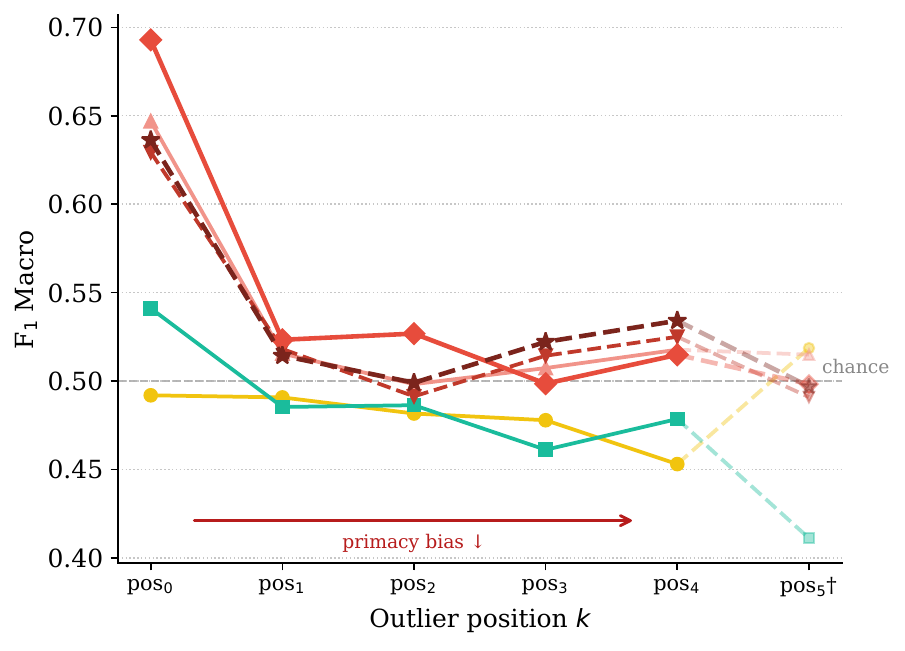}
    \caption{$\text{F}_1$ Semantic}
    \label{fig:pos_f1_semantic}
\end{subfigure}\hfill
\begin{subfigure}[b]{0.245\linewidth}
    \centering
    \includegraphics[width=\linewidth]{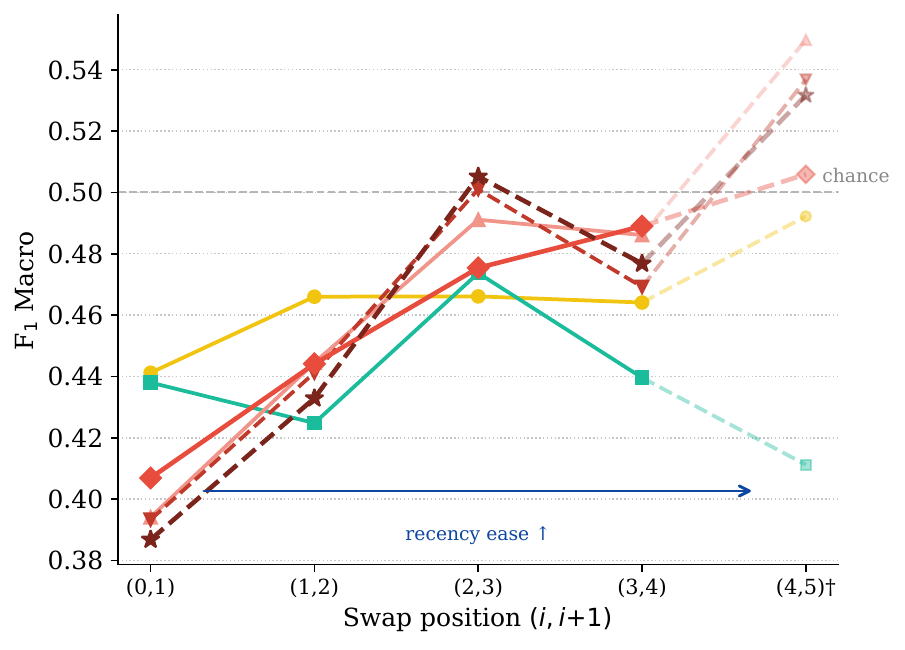}
    \caption{$\text{F}_1$ Temporal}
    \label{fig:pos_f1_temporal}
\end{subfigure}\hfill
\begin{subfigure}[b]{0.245\linewidth}
    \centering
    \includegraphics[width=\linewidth]{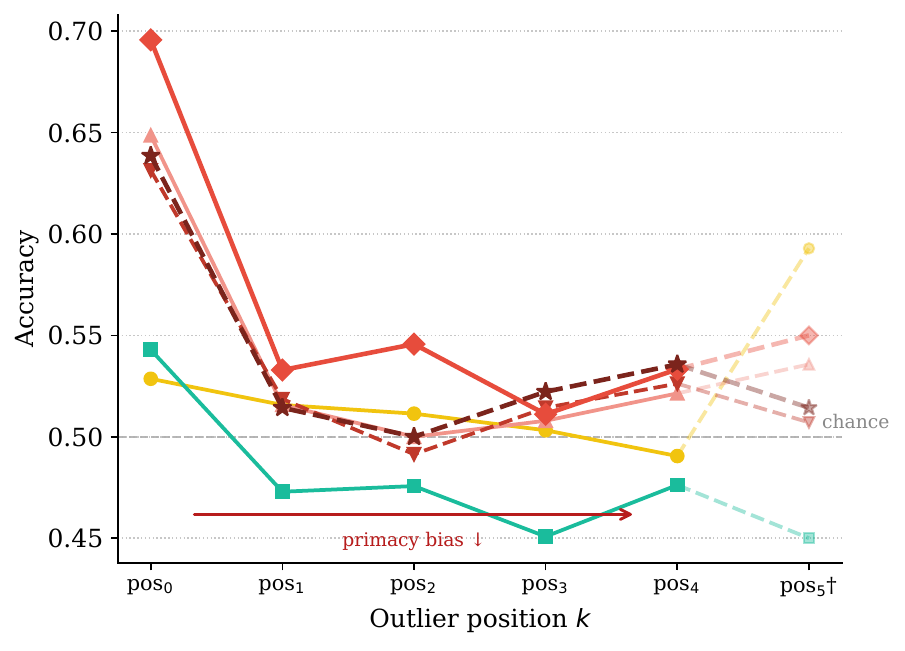}
    \caption{Acc Semantic}
    \label{fig:pos_acc_semantic}
\end{subfigure}\hfill
\begin{subfigure}[b]{0.245\linewidth}
    \centering
    \includegraphics[width=\linewidth]{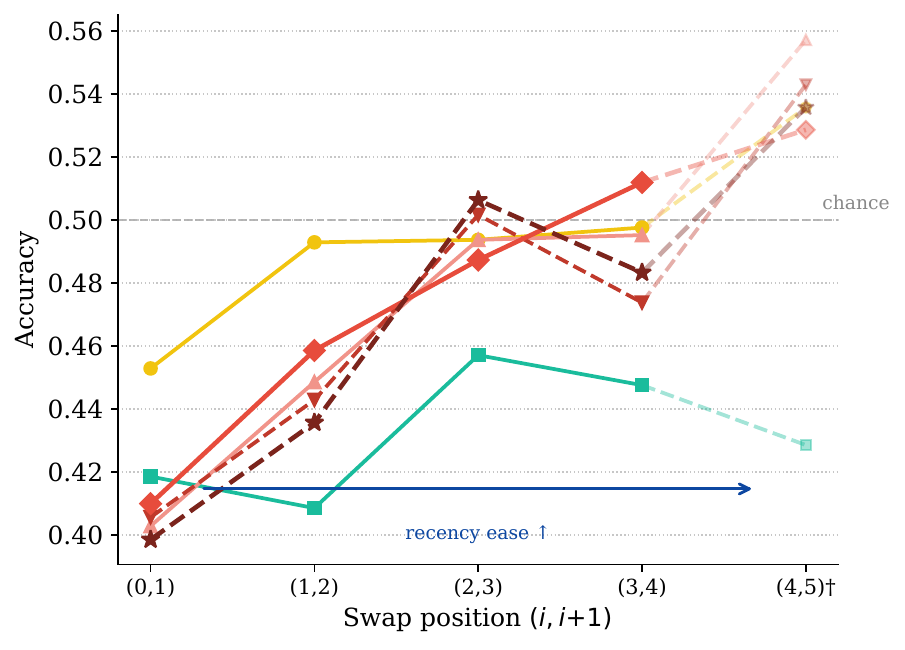}
    \caption{Acc Temporal}
    \label{fig:pos_acc_temporal}
\end{subfigure}
\caption{\textbf{Systematic Positional Sensitivity Analysis.} The top row reveals a clear ``U-shaped'' performance curve, exposing primacy and recency biases across both semantic and temporal tasks.}
\label{fig:positional_sensitivity}
\end{figure*}

\section{Evaluation on Larger and Architecturally Distinct Models}
\label{sec:larger_models}

A natural question is whether the structural blindness we identify is
specific to 7B-scale judges sharing a single model family.  To address
this, we ran zero-shot pairwise evaluation on two substantially larger
models from different architectural families, using the same benchmarks
(\textsc{Prism} and \textsc{Mirage}) and the same prompt configurations
(P0--P6) as the main paper.

\paragraph{Models.}
\textbf{Gemma-4-31B}~\cite{gemmateam2026gemma4technicalreport} is a 31B
dense vision-language model by Google DeepMind, accessed via API.
\textbf{Qwen3.6-27B}~\cite{qwen3.6-27b} is a 27B dense multimodal
model by Alibaba, accessed via API.
Both are evaluated zero-shot at temperature~0.

\paragraph{Scale improves aggregate accuracy.}
Table~\ref{tab:larger_models} reports accuracy and the number of valid
responses per prompt configuration.  Gemma-4-31B reaches
Acc\,=\,0.91 on \textsc{Prism}, well above the 0.43--0.72 range of our
7B judges: at roughly four times the parameter count, pairwise
discrimination improves substantially.  This is worth stating plainly,
since it bounds the scope of our claim, as the difficulty we diagnose
is not immune to capacity.

\paragraph{Yet the same failure modes persist.}
Three patterns nonetheless reproduce those observed at 7B.
\textit{(1) Answer invalidity remains severe.}  Gemma-4-31B produces no
extractable answer in 1{,}622 of 4{,}200 \textsc{Prism} inferences
(38.6\%).  Qwen3.6-27B produces 338 NULLs on \textsc{Prism} and 45 on
\textsc{Mirage}, concentrated in P1--P5 where it generates extended
reasoning that never resolves to a final A/B choice, behaviour
analogous to the verbosity bias noted in Section~6.2 of the main paper.
\textit{(2) Prompt sensitivity is undiminished.}  On \textsc{Prism},
Qwen3.6-27B ranges from Acc\,=\,0.49 (P1--P4) to Acc\,=\,0.83 (P6), a
gap of 0.34, wider than the gap observed for our 7B judges.  Prompt
layout remains a critical and unstable factor regardless of scale.
\textit{(3) The benchmark ordering is unchanged.}  \textsc{Mirage}
remains easier than \textsc{Prism} for both models, exactly as in the
main paper.

Capacity therefore shifts the operating point without altering the
structure of the failure.  The next section shows that the positional
signature underlying it is likewise unchanged.

\begin{table*}[htbp]
\centering
\caption{%
  \textbf{Zero-shot pairwise evaluation — Gemma-4-31B~\cite{gemmateam2026gemma4technicalreport}
  and Qwen3.6-27B~\cite{qwen3.6-27b}
  (response to Reviewers fbNz and ht4N, W1).}
  Same benchmarks and prompt configurations (P0--P6) as the main paper.
  \textbf{Acc}: accuracy over valid responses.
  \textbf{NULL}: inferences with no valid A/B answer extracted
  (grey cells).
  Both models accessed via  API, temperature\,=\,0.
  \textbf{Bold}: best accuracy per model\,/\,dataset block.
}
\label{tab:larger_models}
\scriptsize
\setlength{\tabcolsep}{3.5pt}
\renewcommand{\arraystretch}{1.25}
\definecolor{nullgray}{RGB}{235,235,235}
\resizebox{\textwidth}{!}{%
\begin{tabular}{@{} l
    r r r r
    r r r r
    r r r r
    r r r r @{}}
\toprule
& \multicolumn{4}{c}{\textbf{Gemma-4-31B~\cite{gemmateam2026gemma4technicalreport} / PRISM}
    ($n$\,=\,4200)}
& \multicolumn{4}{c}{\textbf{Gemma-4-31B~\cite{gemmateam2026gemma4technicalreport} / MIRAGE}
    ($n$\,=\,672)}
& \multicolumn{4}{c}{\textbf{Qwen3.6-27B~\cite{qwen3.6-27b} / PRISM}
    ($n$\,=\,4200)}
& \multicolumn{4}{c}{\textbf{Qwen3.6-27B~\cite{qwen3.6-27b} / MIRAGE}
    ($n$\,=\,672)} \\
\cmidrule(lr){2-5}\cmidrule(lr){6-9}
\cmidrule(lr){10-13}\cmidrule(lr){14-17}
\textbf{Prompt}
  & \textbf{Valid} & \textbf{NULL} & \textbf{Corr.} & \textbf{Acc}
  & \textbf{Valid} & \textbf{NULL} & \textbf{Corr.} & \textbf{Acc}
  & \textbf{Valid} & \textbf{NULL} & \textbf{Corr.} & \textbf{Acc}
  & \textbf{Valid} & \textbf{NULL} & \textbf{Corr.} & \textbf{Acc} \\
\midrule

P0
  & 443 & \cellcolor{nullgray}157 & 415 & \textbf{0.94}
  &  96 & 0                       &  90 & 0.94
  & 501 & \cellcolor{nullgray}99  & 333 & 0.66
  &  64 & \cellcolor{nullgray}32  &  56 & 0.88 \\

P1
  & 272 & \cellcolor{nullgray}328 & 246 & 0.90
  &  96 & 0                       &  91 & \textbf{0.95}
  & 600 & 0                       & 294 & 0.49
  &  96 & 0                       &  42 & 0.44 \\

P2
  & 296 & \cellcolor{nullgray}304 & 273 & 0.92
  &  96 & 0                       &  91 & \textbf{0.95}
  & 600 & 0                       & 294 & 0.49
  &  96 & 0                       &  42 & 0.44 \\

P3
  & 362 & \cellcolor{nullgray}238 & 336 & 0.93
  &  96 & 0                       &  90 & 0.94
  & 600 & 0                       & 294 & 0.49
  &  96 & 0                       &  42 & 0.44 \\

P4
  & 409 & \cellcolor{nullgray}191 & 376 & 0.92
  &  96 & 0                       &  89 & 0.93
  & 600 & 0                       & 294 & 0.49
  &  96 & 0                       &  42 & 0.44 \\

P5
  & 421 & \cellcolor{nullgray}179 & 361 & 0.86
  &  96 & 0                       &  90 & 0.94
  & 544 & \cellcolor{nullgray}56  & 311 & 0.57
  &  94 & \cellcolor{nullgray}2   &  52 & 0.55 \\

P6
  & 375 & \cellcolor{nullgray}225 & 331 & 0.88
  &  96 & 0                       &  89 & 0.93
  & 417 & \cellcolor{nullgray}183 & 347 & \textbf{0.83}
  &  85 & \cellcolor{nullgray}11  &  78 & \textbf{0.92} \\

\midrule
\rowcolor{headerblue}
\textbf{AVG}
  & \textbf{2578} & \textbf{1622} & \textbf{2338} & \textbf{0.91}
  & \textbf{672}  & \textbf{0}    & \textbf{630}  & \textbf{0.94}
  & \textbf{3862} & \textbf{338}  & \textbf{2167} & \textbf{0.56}
  & \textbf{627}  & \textbf{45}   & \textbf{354}  & \textbf{0.56} \\

\bottomrule
\end{tabular}%
}

\smallskip
\noindent\footnotesize
\textbf{NULL} cells (grey): no extractable A/B answer; Acc computed
over valid responses only.
Qwen3.6-27B NULLs are concentrated in P1--P5, where the model
generates extended reasoning without committing to a final choice,
mirroring the verbosity bias of LLaVA-Critic reported in
Section~6.2 of the main paper.
\end{table*}

\begin{table*}[htbp]
\centering
\caption{%
  \textbf{Positional sensitivity — per outlier/swap position
  (Gemma-4-31B vs.\ Qwen3.6-27B, \textsc{Prism}-Type1).}
  \textbf{Semantic}: semantic outlier at the given frame position.
  \textbf{Temporal}: consecutive temporal swap between adjacent positions.
  Acc computed over valid responses.
  \textbf{Bold}: best Acc per perturbation type per model.
}
\label{tab:sensitivity_position}
\scriptsize
\setlength{\tabcolsep}{4pt}
\renewcommand{\arraystretch}{1.25}
\resizebox{\textwidth}{!}{%
\begin{tabular}{@{} l l
    r r r r
    r r r r @{}}
\toprule
& &
\multicolumn{4}{c}{\textbf{Gemma-4-31B}~\cite{gemmateam2026gemma4technicalreport}} &
\multicolumn{4}{c}{\textbf{Qwen3.6-27B}~\cite{qwen3.6-27b}} \\
\cmidrule(lr){3-6}\cmidrule(lr){7-10}
\textbf{Type} & \textbf{Position}
  & \textbf{Total} & \textbf{Valid} & \textbf{Corr.} & \textbf{Acc}
  & \textbf{Total} & \textbf{Valid} & \textbf{Corr.} & \textbf{Acc} \\
\midrule

\multirow{6}{*}{\textbf{Semantic}}
& pos\,0 & 700 & 680 & 670 & \textbf{0.99} & 700 & 669 & 384 & 0.57 \\
& pos\,1 & 700 & 671 & 643 & 0.96          & 700 & 647 & 373 & 0.58 \\
& pos\,2 & 700 & 674 & 644 & 0.96          & 700 & 656 & 404 & \textbf{0.62} \\
& pos\,3 & 630 & 611 & 579 & 0.95          & 630 & 590 & 374 & \textbf{0.63} \\
& pos\,4 & 420 & 396 & 376 & 0.95          & 420 & 392 & 229 & 0.58 \\
& pos\,5 & 140 & 125 & 119 & 0.95          & 140 & 133 &  65 & 0.49 \\
\midrule

\multirow{5}{*}{\textbf{Temporal}}
& swap\,0-1 & 700 & 671 & 614 & \textbf{0.92} & 700 & 627 & 339 & \textbf{0.54} \\
& swap\,1-2 & 700 & 639 & 490 & 0.77          & 700 & 633 & 332 & 0.52 \\
& swap\,2-3 & 630 & 577 & 397 & 0.69          & 630 & 569 & 308 & 0.54 \\
& swap\,3-4 & 420 & 369 & 267 & 0.72          & 420 & 372 & 181 & 0.49 \\
& swap\,4-5 & 140 & 125 &  72 & 0.58          & 140 & 128 &  60 & 0.47 \\

\bottomrule
\end{tabular}%
}
\end{table*}

\begin{table*}[htbp]
\centering
\caption{%
  \textbf{Positional sensitivity — per prompt type
  (Gemma-4-31B vs.\ Qwen3.6-27B, \textsc{Prism}-Type1).}
  Acc averaged over all outlier/swap positions.
  \textbf{Bold}: best Acc per perturbation type per model.
}
\label{tab:sensitivity_prompt}
\scriptsize
\setlength{\tabcolsep}{3.5pt}
\renewcommand{\arraystretch}{1.05}
\resizebox{0.85\textwidth}{!}{%
\begin{tabular}{@{} l l
    r r r r
    r r r r @{}}
\toprule
& &
\multicolumn{4}{c}{\textbf{Gemma-4-31B}~\cite{gemmateam2026gemma4technicalreport}} &
\multicolumn{4}{c}{\textbf{Qwen3.6-27B}~\cite{qwen3.6-27b}} \\
\cmidrule(lr){3-6}\cmidrule(lr){7-10}
\textbf{Type} & \textbf{Prompt}
  & \textbf{Total} & \textbf{Valid} & \textbf{NULL} & \textbf{Acc}
  & \textbf{Total} & \textbf{Valid} & \textbf{NULL} & \textbf{Acc} \\
\midrule

\multirow{7}{*}{\textbf{Semantic}}
& P0 & 470 & 441 & 29 & \textbf{0.97} & 470 & 413 &  57 & 0.71 \\
& P1 & 470 & 465 &  5 & \textbf{0.97} & 470 & 470 &   0 & 0.51 \\
& P2 & 470 & 462 &  8 & \textbf{0.97} & 470 & 470 &   0 & 0.51 \\
& P3 & 470 & 448 & 22 & \textbf{0.97} & 470 & 470 &   0 & 0.51 \\
& P4 & 470 & 454 & 16 & \textbf{0.97} & 470 & 470 &   0 & 0.51 \\
& P5 & 470 & 459 & 11 & 0.91          & 470 & 438 &  32 & 0.60 \\
& P6 & 470 & 428 & 42 & 0.96          & 470 & 356 & 114 & \textbf{0.87} \\
\midrule

\multirow{7}{*}{\textbf{Temporal}}
& P0 & 370 & 344 & 26 & 0.78          & 370 & 320 &  50 & 0.52 \\
& P1 & 370 & 341 & 29 & \textbf{0.82} & 370 & 370 &   0 & 0.50 \\
& P2 & 370 & 335 & 35 & 0.80          & 370 & 370 &   0 & 0.50 \\
& P3 & 370 & 339 & 31 & 0.75          & 370 & 370 &   0 & 0.50 \\
& P4 & 370 & 346 & 24 & 0.76          & 370 & 370 &   0 & 0.50 \\
& P5 & 370 & 356 & 14 & 0.70          & 370 & 321 &  49 & 0.53 \\
& P6 & 370 & 320 & 50 & 0.81          & 370 & 208 & 162 & \textbf{0.70} \\

\bottomrule
\end{tabular}%
}
\end{table*}

\section{Positional Sensitivity Analysis on Larger Models}
\label{sec:sensitivity_larger}

The attribution of primacy and recency effects to causal masking and
RoPE in Section~3 of the main paper rests on prior literature rather
than on direct measurement.  To provide empirical support, we
replicated the positional sensitivity analysis of Section~6.3 on
Gemma-4-31B~\cite{gemmateam2026gemma4technicalreport} and
Qwen3.6-27B~\cite{qwen3.6-27b}, using the same \textsc{Prism}
perturbation protocol.

\paragraph{Setup.}
We evaluate on \textsc{Prism}-Type1 ($n=840$ records), comprising two
perturbation families.  \textbf{Semantic} ($n=470$): a contextually
irrelevant frame replaces the correct one at positions 0--5, testing
detection of content mismatches across the sequence.
\textbf{Temporal} ($n=370$): two adjacent frames are exchanged at
positions 0-1 through 4-5, testing causal order sensitivity.
Both models are evaluated zero-shot across all seven prompt
configurations at temperature~0.

\paragraph{Primacy persists at larger scale.}
On semantic perturbations, Gemma-4-31B peaks at Acc\,=\,0.99 when the
outlier occupies the opening frame and settles at 0.95--0.96 for all
later positions (Table~\ref{tab:sensitivity_position}), the same
primacy profile reported for the LLaVA judges in Figure~4 of the main
paper, displaced upward but identical in shape.

\paragraph{The temporal gradient is unchanged, and steeper.}
On temporal perturbations, Gemma-4-31B falls from Acc\,=\,0.92 at the
opening pair to 0.58 at the closing one, a gap of 0.34.  Detection
degrades as the swap moves away from the start of the sequence, and
mid-sequence swaps are substantially harder than early ones,
consistent with RoPE suppressing interactions between distant tokens.
Notably, this is the one axis on which scale offers no protection:
the model that reaches 0.91 in aggregate still loses a third of its
accuracy as a function of where the violation sits.

\paragraph{Key takeaway.}
Positional asymmetry is not an artefact of 7B-scale models or of a
particular training regime.  It survives a fourfold increase in
capacity and a change of model family, which is what one would expect
of a constraint originating in the attention mechanism itself.

\section{Visual Chronological Ordering Probe}
\label{sec:probe}

\subsection{Motivation}

The previous sections show that scale improves pairwise accuracy while
leaving the positional signature intact.  This section isolates the
underlying capability with a task that admits no shortcut through
pairwise comparison.

Temporal reasoning over visual sequences differs fundamentally from its
textual counterpart.  Asked in text whether the Roman Empire predates
the Great Wall of China, a model can draw on factual knowledge encoded
during pretraining.  Presented with \textbf{shuffled historical images
carrying no labels or timestamps}, it must infer chronological order
\emph{from visual content alone}.  This is the capability our paper
targets, and the one current LVLM judges demonstrably lack.

\subsection{Experimental Setup}

\paragraph{Task definition.}
Each probe instance consists of four historical images presented in a
\emph{randomised} order, labelled A, B, C, D.  The model receives
\emph{no} textual descriptions, captions, dates or step labels, only
the raw images.  It is asked to output the correct chronological
ordering of the four labels (e.g.\ \texttt{B-D-A-C}).
Figure~\ref{fig:probe_example} illustrates one probe instance
(S3, Space Race and Moon Landing), showing both the gold order and the
shuffled order presented to the model.

\paragraph{Prompt.}
The instruction given to all models is identical:
\begin{quote}
\small\itshape
\textbf{System:} You are an expert historian analysing visual sequences.\\[4pt]
\textbf{User:}
You are shown 4 historical images in shuffled order, labelled A, B, C, D.
Based ONLY on the visual content, with no text and no dates,
determine the correct chronological order.
Answer with ONLY the letters, e.g.\ \texttt{B\textendash A\textendash D\textendash C}.
Chronological order:
\end{quote}

\paragraph{Image sequences.}
We constructed 10 sequences spanning diverse historical periods
(Table~\ref{tab:probe_results}).  Each sequence uses four \texttt{.png}
images sourced from Wikipedia\,/\,Wikimedia Commons (public domain or
CC-licensed), resized to $448{\times}448$\,px.

\paragraph{Models evaluated.}
\begin{itemize}
  \item \textbf{LLaVA-OneVision 7B} and \textbf{LLaVA-Critic 7B}: the
    two judge backbones from the main paper, run locally on GPU (H200),
    zero-shot.
  \item \textbf{Gemma-4-31B}~\cite{gemmateam2026gemma4technicalreport}:
    a 31B dense vision-language model by Google DeepMind, accessed via
    the API.
  \item \textbf{Qwen3.6-27B}~\cite{qwen3.6-27b}: a 27B dense
    multimodal model by Alibaba, accessed via the API.
\end{itemize}
All models use temperature\,=\,0 (deterministic inference).

\paragraph{Evaluation.}
A prediction is correct (\cmark) only when the full four-element
ordering \emph{exactly} matches the gold order.

\newlength{\probecolW}\setlength{\probecolW}{0.18\textwidth}

\begin{figure*}[t]
\centering
\footnote
{\footnotesize\textbf{Gold chronological order:}
  \texttt{C\,\textrightarrow\,A\,\textrightarrow\,D\,\textrightarrow\,B}}\\[4pt]

\begin{minipage}[t]{\probecolW}
  \begin{tcolorbox}[enhanced, sharp corners=all,
      colback=white, colframe=black!50,
      boxrule=0.5pt, left=2pt, right=2pt, top=2pt, bottom=3pt,
      width=\linewidth, height=3.8cm, nobeforeafter, valign=top]
    \includegraphics[width=\linewidth, height=2.1cm, keepaspectratio=false]{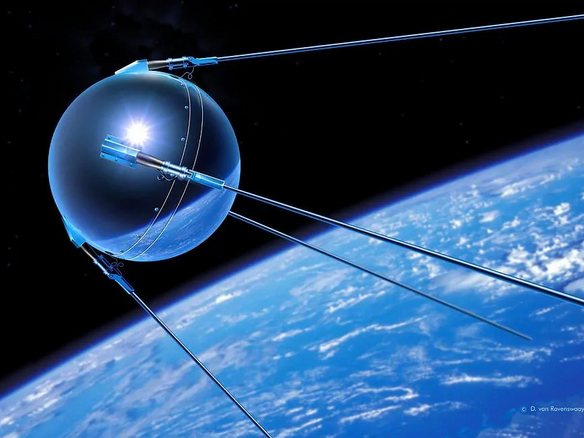}\\[3pt]
    {\tiny\textbf{1st}}\\[0.5pt]
    {\tiny\raggedright Sputnik satellite launch (1957)\par}
  \end{tcolorbox}
\end{minipage}\hfill
\begin{minipage}[t]{\probecolW}
  \begin{tcolorbox}[enhanced, sharp corners=all,
      colback=white, colframe=black!50,
      boxrule=0.5pt, left=2pt, right=2pt, top=2pt, bottom=3pt,
      width=\linewidth, height=3.8cm, nobeforeafter, valign=top]
    \includegraphics[width=\linewidth, height=2.1cm, keepaspectratio=false]{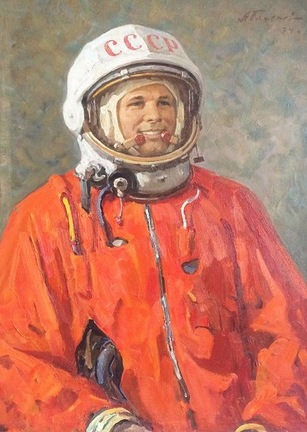}\\[3pt]
    {\tiny\textbf{2nd}}\\[0.5pt]
    {\tiny\raggedright Yuri Gagarin, first human in space (1961)\par}
  \end{tcolorbox}
\end{minipage}\hfill
\begin{minipage}[t]{\probecolW}
  \begin{tcolorbox}[enhanced, sharp corners=all,
      colback=white, colframe=black!50,
      boxrule=0.5pt, left=2pt, right=2pt, top=2pt, bottom=3pt,
      width=\linewidth, height=3.8cm, nobeforeafter, valign=top]
    \includegraphics[width=\linewidth, height=2.1cm, keepaspectratio=false]{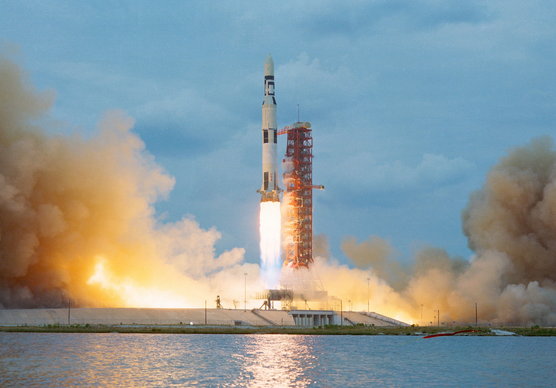}\\[3pt]
    {\tiny\textbf{3rd}}\\[0.5pt]
    {\tiny\raggedright Apollo~11 Saturn~V launch (1969)\par}
  \end{tcolorbox}
\end{minipage}\hfill
\begin{minipage}[t]{\probecolW}
  \begin{tcolorbox}[enhanced, sharp corners=all,
      colback=white, colframe=black!50,
      boxrule=0.5pt, left=2pt, right=2pt, top=2pt, bottom=3pt,
      width=\linewidth, height=3.8cm, nobeforeafter, valign=top]
    \includegraphics[width=\linewidth, height=2.1cm, keepaspectratio=false]{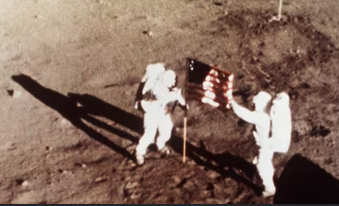}\\[3pt]
    {\tiny\textbf{4th}}\\[0.5pt]
    {\tiny\raggedright Neil Armstrong on the Moon (1969)\par}
  \end{tcolorbox}
\end{minipage}

\vspace{6pt}

{\footnotesize\textbf{Shuffled order presented to the model
  (labels A--D, no dates):}}\\[4pt]

\begin{minipage}[t]{\probecolW}
  \begin{tcolorbox}[enhanced, sharp corners=all,
      colback=white, colframe=black!50,
      boxrule=0.6pt, frame style={dash pattern=on 3pt off 2pt},
      left=2pt, right=2pt, top=2pt, bottom=3pt,
      width=\linewidth, height=3.8cm, nobeforeafter, valign=top]
    \includegraphics[width=\linewidth, height=2.1cm, keepaspectratio=false]{probe_test_img/seq3_2.png}\\[3pt]
    {\tiny\textbf{Image A}}\\[0.5pt]
    {\tiny\raggedright \textit{(no caption)}\par}
  \end{tcolorbox}
\end{minipage}\hfill
\begin{minipage}[t]{\probecolW}
  \begin{tcolorbox}[enhanced, sharp corners=all,
      colback=white, colframe=black!50,
      boxrule=0.6pt, frame style={dash pattern=on 3pt off 2pt},
      left=2pt, right=2pt, top=2pt, bottom=3pt,
      width=\linewidth, height=3.8cm, nobeforeafter, valign=top]
    \includegraphics[width=\linewidth, height=2.1cm, keepaspectratio=false]{probe_test_img/seq3_4.png}\\[3pt]
    {\tiny\textbf{Image B}}\\[0.5pt]
    {\tiny\raggedright \textit{(no caption)}\par}
  \end{tcolorbox}
\end{minipage}\hfill
\begin{minipage}[t]{\probecolW}
  \begin{tcolorbox}[enhanced, sharp corners=all,
      colback=white, colframe=black!50,
      boxrule=0.6pt, frame style={dash pattern=on 3pt off 2pt},
      left=2pt, right=2pt, top=2pt, bottom=3pt,
      width=\linewidth, height=3.8cm, nobeforeafter, valign=top]
    \includegraphics[width=\linewidth, height=2.1cm, keepaspectratio=false]{probe_test_img/seq3_1.png}\\[3pt]
    {\tiny\textbf{Image C}}\\[0.5pt]
    {\tiny\raggedright \textit{(no caption)}\par}
  \end{tcolorbox}
\end{minipage}\hfill
\begin{minipage}[t]{\probecolW}
  \begin{tcolorbox}[enhanced, sharp corners=all,
      colback=white, colframe=black!50,
      boxrule=0.6pt, frame style={dash pattern=on 3pt off 2pt},
      left=2pt, right=2pt, top=2pt, bottom=3pt,
      width=\linewidth, height=3.8cm, nobeforeafter, valign=top]
    \includegraphics[width=\linewidth, height=2.1cm, keepaspectratio=false]{probe_test_img/seq3_3.png}\\[3pt]
    {\tiny\textbf{Image D}}\\[0.5pt]
    {\tiny\raggedright \textit{(no caption)}\par}
  \end{tcolorbox}
\end{minipage}

\vspace{0.5pt}
\Description{Two rows of four historical photographs each. The top row
shows the Space Race sequence in correct chronological order, labelled
first to fourth, with captions and dates. The bottom row, outlined with
dashed borders, shows the same four images rearranged and labelled A to
D with no captions or dates.}
\caption{%
  \textbf{Probe example: S3, Space Race and Moon Landing.}
  \emph{Top}: the four images in their correct chronological order with
  labels, shown here for reference only.
  \emph{Bottom} (dashed border): the same images \textbf{shuffled} and
  presented to the model as Images A--D \emph{without} caption, date or
  label. The correct answer is \texttt{C-A-D-B}.
  Gemma-4-31B answers correctly; LLaVA-Critic answers \texttt{D-C-B-A}
  (reverse label order); Qwen3.6-27B produces no valid answer.
}
\label{fig:probe_example}
\end{figure*}
\subsection{Results and Analysis}

Table~\ref{tab:probe_results} reports the full results.

\paragraph{The 7B judges perform at chance.}
LLaVA-OneVision achieves \textbf{0/10} and LLaVA-Critic \textbf{1/10},
indistinguishable from random guessing ($1/24\!\approx\!4\%$ for a
four-element ordering).

\paragraph{LLaVA-Critic answers by position, not by content.}
LLaVA-Critic returns \texttt{D-C-B-A}, the reverse of the label order,
in \textbf{8 of 10} sequences regardless of what the images depict.
This is the positional heuristic of Section~6.3 in its purest form:
the output is a function of the input layout alone.  The Space Race
example in Figure~\ref{fig:probe_example} makes the contrast explicit,
with Gemma-4-31B recovering \texttt{C-A-D-B} while LLaVA-Critic simply
inverts the labels.

\paragraph{Larger models improve but do not solve the task.}
Gemma-4-31B reaches \textbf{7/10} and Qwen3.6-27B \textbf{3/10}.
Qwen3.6-27B frequently fails to produce an extractable answer,
generating verbose reasoning that never converges to a final label
order, the same collapse observed in
Section~\ref{sec:larger_models}.

\paragraph{Key takeaway.}
The contrast with Section~\ref{sec:larger_models} is instructive.
Gemma-4-31B reaches 0.91 on pairwise \textsc{Prism} but recovers the
correct chronological order in only 7 of 10 probe sequences, where a
single misplaced element invalidates the answer.  Pairwise
discrimination between a gold sequence and one perturbed variant is a
substantially weaker test than reconstructing order outright, which is
precisely why aggregate pairwise accuracy is a misleading proxy for
chronological competence.

\begin{table*}[htbp]
\centering
\caption{%
  \textbf{Visual chronological ordering probe, full results.}
  Given four historical images in shuffled order (labels A--D), models
  must predict the correct chronological order from visual content
  alone, with no text, timestamps or dates.
  \colorbox{correct}{\cmark}~correct order;
  \colorbox{wrong}{\xmark}~wrong order;
  \colorbox{noans}{---}~no valid answer extracted.
  \textbf{Gold}: correct label order, which depends on the shuffle.
  LLaVA models run locally, zero-shot; Gemma-4-31B and Qwen3.6-27B via
}
\label{tab:probe_results}
\scriptsize
\setlength{\tabcolsep}{3.5pt}
\renewcommand{\arraystretch}{1.35}
\newcommand{\ccell}[1]{\cellcolor{correct}{\cmark}~\texttt{#1}}
\newcommand{\xcell}[1]{\cellcolor{wrong}{\xmark}~\texttt{#1}}
\newcommand{\ncell}{\cellcolor{noans}\texttt{---}}
\resizebox{\textwidth}{!}{%
\begin{tabular}{@{} c >{\raggedright}p{2.8cm}
    >{\raggedright}p{5.8cm} c  c  c  c  c @{}}
\toprule
\textbf{ID}
  & \textbf{Topic}
  & \textbf{Chronological sequence (gold order)}
  & \textbf{Gold}
  & \makecell{\textbf{Gemma-4-31B}\\\textit{(31B, API)}}
  & \makecell{\textbf{Qwen3.6-27B}\\\textit{(27B, API)}}
  & \makecell{\textbf{LLaVA-OV}\\\textit{(7B, local)}}
  & \makecell{\textbf{LLaVA-Critic}\\\textit{(7B, local)}} \\
\midrule

S1  & Communication technology
    & Gutenberg press $\to$ Steam engine $\to$ Telephone $\to$ Internet
    & \texttt{B-D-A-C} & \ccell{B-D-A-C} & \ccell{B-D-A-C}
    & \xcell{B-C-D-A} & \xcell{D-C-B-A} \\

S2  & Roman Empire
    & Republic $\to$ Empire at peak $\to$ Decline $\to$ Fall
    & \texttt{C-B-D-A} & \xcell{B-C-D-A} & \ncell
    & \xcell{B-C-D-A} & \ccell{C-B-D-A} \\

S3  & \textbf{Space race and Moon landing} \textsuperscript{$\dagger$}
    & Sputnik $\to$ Gagarin $\to$ Apollo~11 $\to$ Moon surface
    & \texttt{C-A-D-B} & \ccell{C-A-D-B} & \ncell
    & \xcell{D-C-B-A} & \xcell{D-C-B-A} \\

S4  & World War~II
    & Rise of Nazism $\to$ Poland $\to$ D-Day $\to$ VE Day
    & \texttt{B-D-A-C} & \ccell{B-D-A-C} & \ncell
    & \xcell{B-C-D-A} & \xcell{D-C-B-A} \\

S5  & Human evolution
    & Australopithecus $\to$ Homo erectus $\to$ Cave art $\to$ Civilisation
    & \texttt{B-D-C-A} & \ccell{B-D-C-A} & \ccell{B-D-C-A}
    & \xcell{C-B-D-A} & \xcell{D-C-B-A} \\

S6  & French Revolution
    & Louis~XVI $\to$ Bastille $\to$ Terror $\to$ Napoleon
    & \texttt{C-A-D-B} & \xcell{C-D-B-A} & \ncell
    & \xcell{D-C-B-A} & \xcell{D-C-B-A} \\

S7  & Industrial Revolution
    & Hand loom $\to$ Spinning factory $\to$ Railways $\to$ Industrial city
    & \texttt{C-B-D-A} & \ccell{C-B-D-A} & \ncell
    & \xcell{B-C-D-A} & \xcell{D-C-B-A} \\

S8  & Ancient Egypt
    & Predynastic $\to$ Giza $\to$ Karnak $\to$ Cleopatra era
    & \texttt{B-D-A-C} & \xcell{D-B-A-C} & \ncell
    & \xcell{B-C-D-A} & \xcell{D-C-B-A} \\

S9  & Cold War
    & Berlin airlift $\to$ Cuban crisis $\to$ Nixon/Mao $\to$ Wall falls
    & \texttt{D-A-C-B} & \ccell{D-A-C-B} & \ncell
    & \xcell{A-C-D-B} & \xcell{D-C-B-A} \\

S10 & History of flight
    & Wright Brothers $\to$ WWI biplane $\to$ WWII fighter $\to$ Concorde
    & \texttt{B-D-A-C} & \ccell{B-D-A-C} & \ccell{B-D-A-C}
    & \xcell{D-A-C-B} & \xcell{D-C-A-B} \\

\midrule
\rowcolor{headerblue}
\multicolumn{3}{l}{\textbf{Overall accuracy}}
  & ---
  & \textbf{7/10 (70\%)}
  & \textbf{3/10 (30\%)}
  & \textbf{0/10 (0\%)}
  & \textbf{1/10 (10\%)} \\

\bottomrule
\end{tabular}%
}

\smallskip
\noindent\footnotesize
$^\dagger$~S3 is illustrated in Figure~\ref{fig:probe_example}.
\textbf{Positional heuristic in LLaVA-Critic:} outputs
\texttt{D-C-B-A} in 8 of 10 cases regardless of image content.
\textbf{Answer collapse in Qwen3.6-27B:} no valid answer extracted in
7 of 10 cases.
\end{table*}

\subsection{Theoretical Grounding: Allen's Interval Algebra}

The gap this probe exposes can be characterised formally within
\emph{Allen's Interval Algebra}~\cite{allen1983maintaining}, which
defines 13 exhaustive and mutually exclusive temporal relations between
intervals (\textsc{precedes}, \textsc{meets}, \textsc{overlaps},
\textsc{during}, \textsc{starts}, \textsc{finishes}, \textsc{equals},
and their converses).  Our probe tests only the simplest of these,
strict \textsc{precedes} chains among four ordered events, and shows
that even 31B-scale models fail to recover this ordering reliably from
visual input alone.  If the most elementary Allen relation is not
robustly available when presented visually, the twelve more complex
relations, which involve overlapping intervals, containment and
simultaneity, represent a correspondingly deeper open challenge.

The failure is not one of factual knowledge.  Asked in text whether the
Wright Brothers flew before the Second World War, these models answer
correctly.  What is missing is an explicit representation of temporal
order in the visual domain: the ability to maintain and reason over a
sequence of states rather than treating visual information as a
collection of static facts.

\subsection{Broader Relevance}

Chronological intelligence matters wherever the temporal evolution of
visual data does: \textbf{clinical imaging}, where scans must be
ordered by disease progression; \textbf{robotics}, where cause and
effect are inferred from egocentric frames; \textbf{video
understanding}, where narrative violations must be detected; and
\textbf{forensic analysis}, where timelines are reconstructed from
archival footage.  A judge that defaults to positional heuristics, as
LLaVA-Critic demonstrably does, is not merely inaccurate in these
settings but silently so.

\section{Qualitative Case Studies}
\label{sec:qualitative}

We present six representative examples comparing the outputs of a
fine-tuned model (OneVision$^\dagger$, trained on \textsc{Prism}/\textsc{Mirage})
and a zero-shot baseline, grouped by experimental condition.

\textbf{\textsc{Mirage} (Cases~1--2)} cover a real recipe progression
against a generative output with hallucinated frames, and a visual
storytelling narrative coherence task.

\textbf{\textsc{Prism}-Semantic (Cases~3--4)} show strong semantic
perturbations at different frame positions; the fine-tuned model
grounds its decision in specific frame-level evidence.

\textbf{\textsc{Prism}-Temporal (Cases~5--6)} show temporal swaps
at different positions; the fine-tuned model explicitly names the
violated causal step while the zero-shot baseline reasons globally.

\begin{table*}[htbp]
\centering

\captionof{table}{%
  \textbf{Case Study~1.} \textsc{Mirage}-Recipes.
  Real recipe progression (Seq.~B) vs.\ generative output with
  hallucinated frames (Seq.~A). Only the fine-tuned model grounds
  its decision in specific step-level evidence.}
\label{tab:example_1}
\setlength{\fboxsep}{0pt}
\begin{tcolorbox}[enhanced,sharp corners=all,colback=white,
  colframe=black!20,boxrule=0.7pt,
  left=4pt,right=4pt,top=4pt,bottom=4pt,width=\textwidth]
  \begin{tcolorbox}[enhanced,sharp corners=all,colback=black!82,
    colframe=black!82,boxrule=0pt,
    left=5pt,right=5pt,top=2pt,bottom=2pt,width=\linewidth]
    \color{white}\small\bfseries Visual Input Sequences
  \end{tcolorbox}
  \vspace{2pt}
  \begin{minipage}[t]{0.485\linewidth}
    \begin{tcolorbox}[enhanced,sharp corners=all,colback=gray!4,
      colframe=black!25,boxrule=0.4pt,
      left=2pt,right=2pt,top=2pt,bottom=2pt,width=\linewidth]
      \centering{\small\bfseries Sequence A}\\[2pt]
      \includegraphics[width=\linewidth,height=0.11\textheight,
        keepaspectratio]{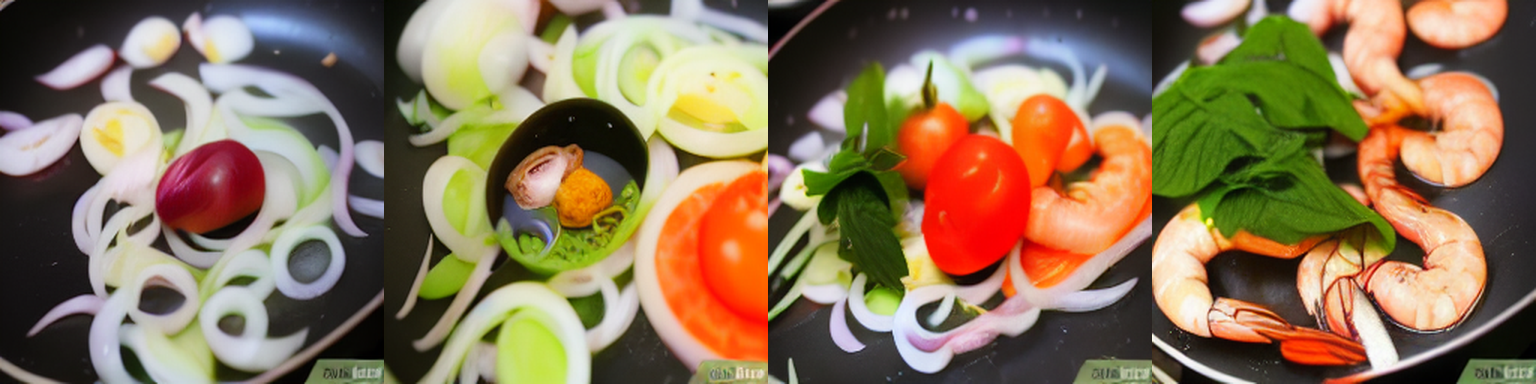}
    \end{tcolorbox}
  \end{minipage}\hfill
  \begin{minipage}[t]{0.485\linewidth}
    \begin{tcolorbox}[enhanced,sharp corners=all,colback=gray!4,
      colframe=black!25,boxrule=0.4pt,
      left=2pt,right=2pt,top=2pt,bottom=2pt,width=\linewidth]
      \centering{\small\bfseries Sequence B}\\[2pt]
      \includegraphics[width=\linewidth,height=0.11\textheight,
        keepaspectratio]{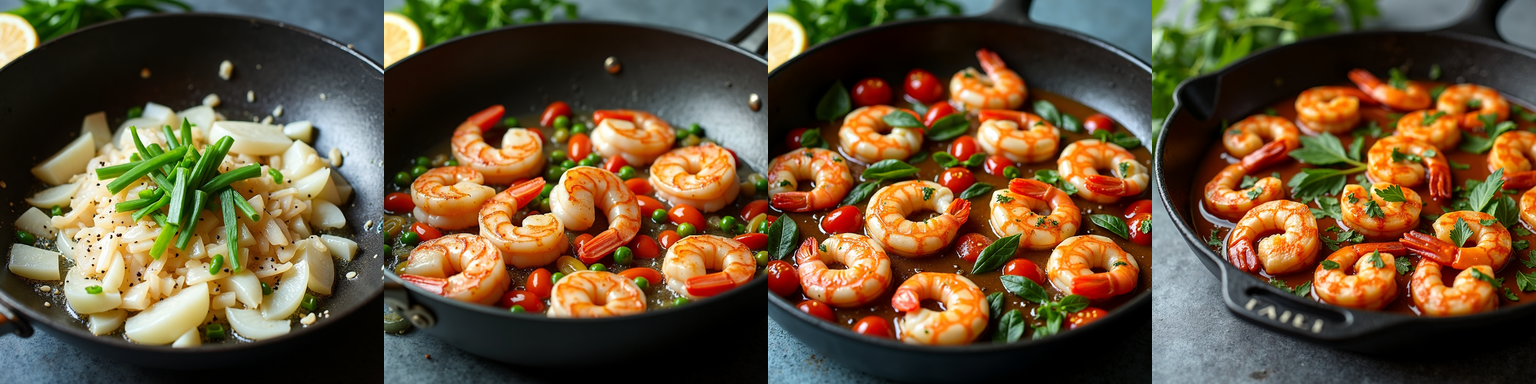}
    \end{tcolorbox}
  \end{minipage}
  \vspace{3pt}
  \begin{tcolorbox}[enhanced,sharp corners=all,colback=blue!4,
    colframe=blue!28,boxrule=0.4pt,
    left=5pt,right=5pt,top=2pt,bottom=2pt,width=\linewidth]
    {\small\bfseries Prompt:}\enspace\small\itshape
    Which sequence better represents: 1.\,Aromatics in oil;
    2.\,Shrimp added; 3.\,Callaloo simmering; 4.\,Final dish?
  \end{tcolorbox}
  \vspace{3pt}
  \begin{tcolorbox}[enhanced,sharp corners=all,colback=black!82,
    colframe=black!82,boxrule=0pt,
    left=5pt,right=5pt,top=2pt,bottom=2pt,width=\linewidth]
    \color{white}\small\bfseries Model Outputs
  \end{tcolorbox}
  \vspace{2pt}
  \noindent
  \begin{minipage}[t]{0.485\linewidth}
    \begin{tcolorbox}[enhanced,sharp corners=all,colback=green!4,
      colframe=green!40!black,boxrule=0.4pt,
      left=4pt,right=4pt,top=3pt,bottom=3pt,width=\linewidth]
      \small{\bfseries\color{green!40!black}$\blacktriangleright$
        Fine-Tuned Model (OneVision$^\dagger$)}\\[2pt]
      \textbf{Better Sequence: B.}\\[2pt]
      Sequence~B shows consistent visual evolution from raw ingredients
      through simmering. Sequence~A lacks temporal coherence at step~3,
      where callaloo integration is absent.
    \end{tcolorbox}
  \end{minipage}\hfill
  \begin{minipage}[t]{0.485\linewidth}
    \begin{tcolorbox}[enhanced,sharp corners=all,colback=orange!5,
      colframe=orange!50!black,boxrule=0.4pt,
      left=4pt,right=4pt,top=3pt,bottom=3pt,width=\linewidth]
      \small{\bfseries\color{orange!60!black}$\blacktriangleright$
        Zero-Shot Baseline (\textsc{OneVision})}\\[2pt]
      \textbf{Better Sequence: B.}\\[2pt]
      Sequence~B appears more consistent with the described process.
      The distinction is subtle; the zero-shot model cannot ground
      its preference in specific step-level violations.
    \end{tcolorbox}
  \end{minipage}
\end{tcolorbox}

\vspace{8pt}

\captionof{table}{%
  \textbf{Case Study~2.} \textsc{Mirage}-Vist.
  Narrative coherence (Washington D.C.\ fireworks story arc).
  The fine-tuned model reasons on environmental continuity;
  the zero-shot model uses visual quality as a proxy.}
\label{tab:example_2}
\begin{tcolorbox}[enhanced,sharp corners=all,colback=white,
  colframe=black!20,boxrule=0.7pt,
  left=4pt,right=4pt,top=4pt,bottom=4pt,width=\textwidth]
  \begin{tcolorbox}[enhanced,sharp corners=all,colback=black!82,
    colframe=black!82,boxrule=0pt,
    left=5pt,right=5pt,top=2pt,bottom=2pt,width=\linewidth]
    \color{white}\small\bfseries Visual Input Sequences
  \end{tcolorbox}
  \vspace{2pt}
  \begin{minipage}[t]{0.485\linewidth}
    \begin{tcolorbox}[enhanced,sharp corners=all,colback=gray!4,
      colframe=black!25,boxrule=0.4pt,
      left=2pt,right=2pt,top=2pt,bottom=2pt,width=\linewidth]
      \centering{\small\bfseries Sequence A}\\[2pt]
      \includegraphics[width=\linewidth,height=0.11\textheight,
        keepaspectratio]{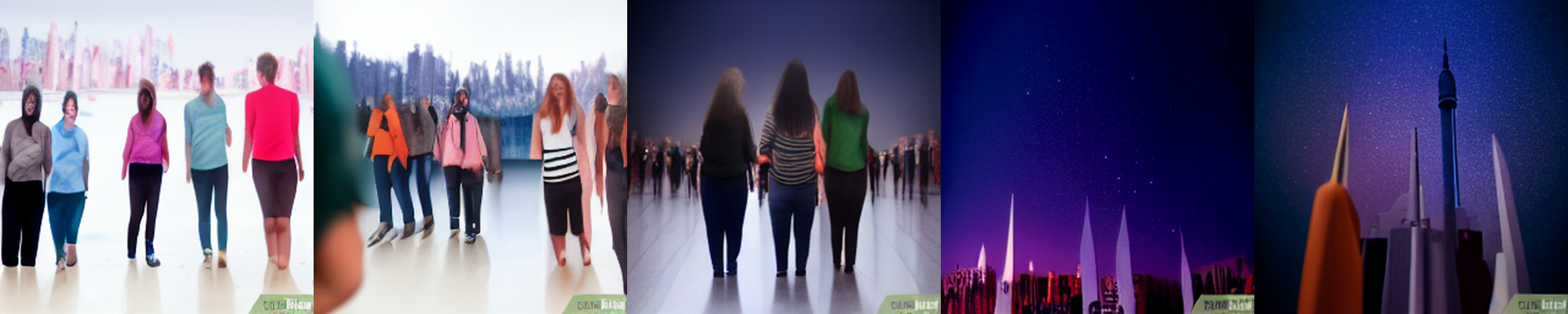}
    \end{tcolorbox}
  \end{minipage}\hfill
  \begin{minipage}[t]{0.485\linewidth}
    \begin{tcolorbox}[enhanced,sharp corners=all,colback=gray!4,
      colframe=black!25,boxrule=0.4pt,
      left=2pt,right=2pt,top=2pt,bottom=2pt,width=\linewidth]
      \centering{\small\bfseries Sequence B}\\[2pt]
      \includegraphics[width=\linewidth,height=0.11\textheight,
        keepaspectratio]{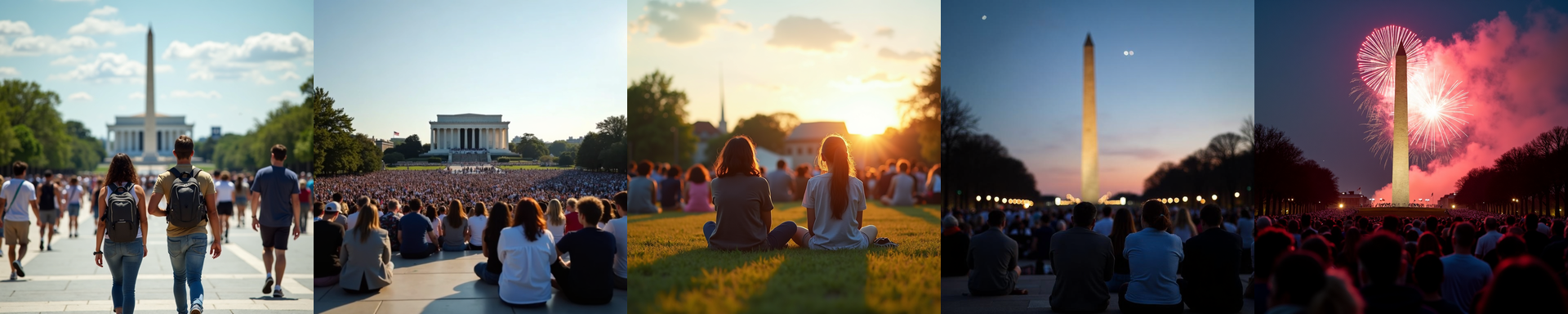}
    \end{tcolorbox}
  \end{minipage}
  \vspace{3pt}
  \begin{tcolorbox}[enhanced,sharp corners=all,colback=blue!4,
    colframe=blue!28,boxrule=0.4pt,
    left=5pt,right=5pt,top=2pt,bottom=2pt,width=\linewidth]
    {\small\bfseries Prompt:}\enspace\small\itshape
    Evaluate the narrative flow: 1.\,Walking to Memorial;
    2.\,Crowd on steps; 3.\,Friends on grass;
    4.\,Waiting at dusk; 5.\,Fireworks at night.
  \end{tcolorbox}
  \vspace{3pt}
  \begin{tcolorbox}[enhanced,sharp corners=all,colback=black!82,
    colframe=black!82,boxrule=0pt,
    left=5pt,right=5pt,top=2pt,bottom=2pt,width=\linewidth]
    \color{white}\small\bfseries Model Outputs
  \end{tcolorbox}
  \vspace{2pt}
  \noindent
  \begin{minipage}[t]{0.485\linewidth}
    \begin{tcolorbox}[enhanced,sharp corners=all,colback=green!4,
      colframe=green!40!black,boxrule=0.4pt,
      left=4pt,right=4pt,top=3pt,bottom=3pt,width=\linewidth]
      \small{\bfseries\color{green!40!black}$\blacktriangleright$
        Fine-Tuned Model (OneVision$^\dagger$)}\\[2pt]
      \textbf{Better Sequence: B.}\\[2pt]
      Sequence~B maintains stronger character and environmental
      consistency, with a coherent daylight-to-night transition
      matching the temporal arc from sightseeing to fireworks.
    \end{tcolorbox}
  \end{minipage}\hfill
  \begin{minipage}[t]{0.485\linewidth}
    \begin{tcolorbox}[enhanced,sharp corners=all,colback=orange!5,
      colframe=orange!50!black,boxrule=0.4pt,
      left=4pt,right=4pt,top=3pt,bottom=3pt,width=\linewidth]
      \small{\bfseries\color{orange!60!black}$\blacktriangleright$
        Zero-Shot Baseline (\textsc{OneVision})}\\[2pt]
      \textbf{Better Sequence: B.}\\[2pt]
      Sequence~B captures the Lincoln Memorial and Washington Monument
      with greater visual fidelity. The zero-shot model relies on
      visual quality rather than narrative logic.
    \end{tcolorbox}
  \end{minipage}
\end{tcolorbox}

\end{table*}

\begin{table*}[htbp]
\centering

\captionof{table}{%
  \textbf{Case Study~3.} \textsc{Prism}-Semantic, strong perturbation
  at position~1. Seq.~A opens with an unrelated blender frame;
  both models identify Seq.~B, but via different reasoning.}
\label{tab:example_3}
\setlength{\fboxsep}{0pt}
\begin{tcolorbox}[enhanced,sharp corners=all,colback=white,
  colframe=black!20,boxrule=0.7pt,
  left=4pt,right=4pt,top=4pt,bottom=4pt,width=\textwidth]
  \begin{tcolorbox}[enhanced,sharp corners=all,colback=black!82,
    colframe=black!82,boxrule=0pt,
    left=5pt,right=5pt,top=2pt,bottom=2pt,width=\linewidth]
    \color{white}\small\bfseries Visual Input Sequences
  \end{tcolorbox}
  \vspace{2pt}
  \begin{minipage}[t]{0.485\linewidth}
    \begin{tcolorbox}[enhanced,sharp corners=all,colback=gray!4,
      colframe=black!25,boxrule=0.4pt,
      left=2pt,right=2pt,top=2pt,bottom=2pt,width=\linewidth]
      \centering{\small\bfseries Sequence A}\\[2pt]
      \includegraphics[width=\linewidth,height=0.11\textheight,
        keepaspectratio]{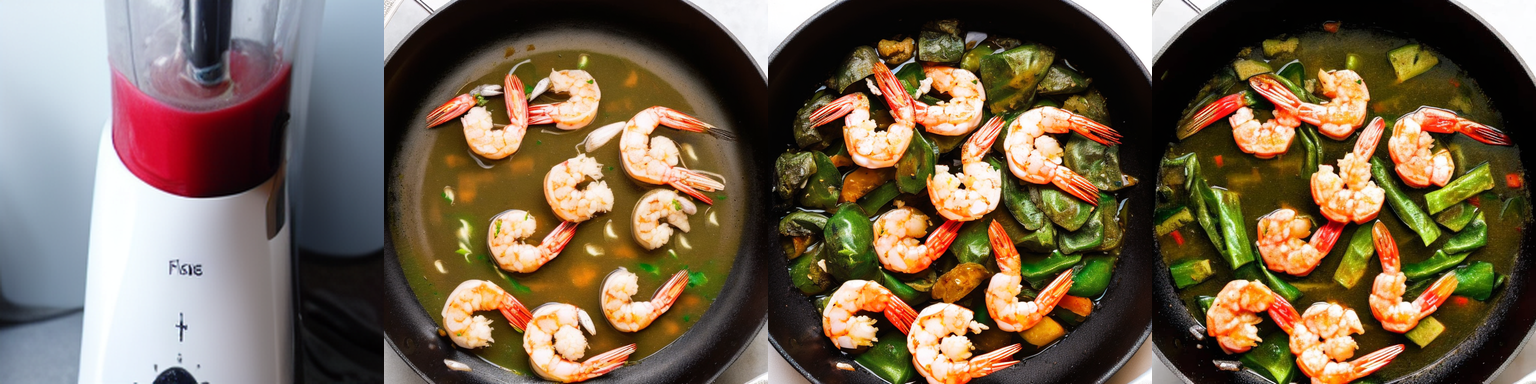}
    \end{tcolorbox}
  \end{minipage}\hfill
  \begin{minipage}[t]{0.485\linewidth}
    \begin{tcolorbox}[enhanced,sharp corners=all,colback=gray!4,
      colframe=black!25,boxrule=0.4pt,
      left=2pt,right=2pt,top=2pt,bottom=2pt,width=\linewidth]
      \centering{\small\bfseries Sequence B}\\[2pt]
      \includegraphics[width=\linewidth,height=0.11\textheight,
        keepaspectratio]{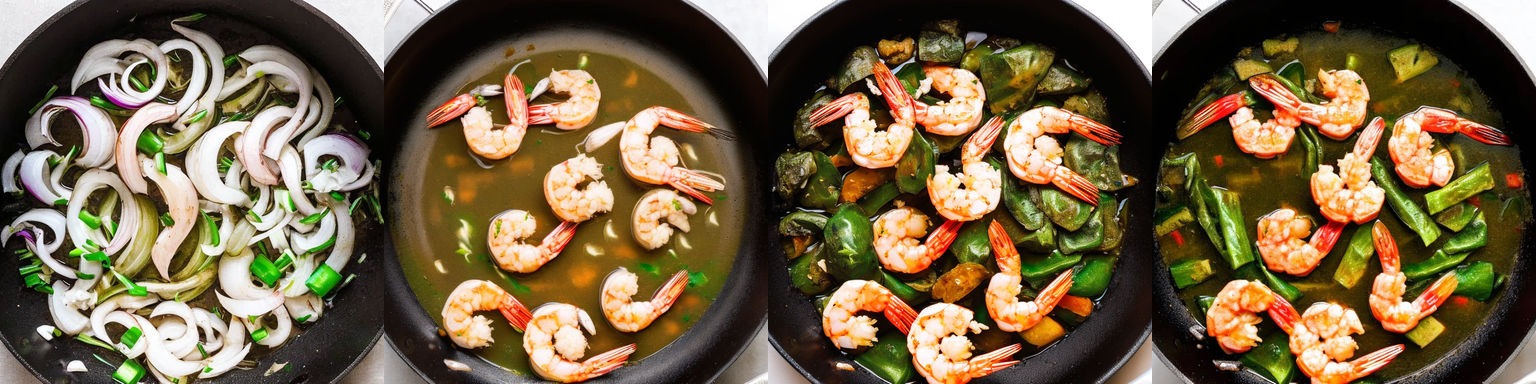}
    \end{tcolorbox}
  \end{minipage}
  \vspace{3pt}
  \begin{tcolorbox}[enhanced,sharp corners=all,colback=blue!4,
    colframe=blue!28,boxrule=0.4pt,
    left=5pt,right=5pt,top=2pt,bottom=2pt,width=\linewidth]
    {\small\bfseries Prompt:}\enspace\small\itshape
    Evaluate: 1.\,Sizzle oil; 2.\,Add shrimp;
    3.\,Simmer with callaloo; 4.\,Dish complete.
  \end{tcolorbox}
  \vspace{3pt}
  \begin{tcolorbox}[enhanced,sharp corners=all,colback=black!82,
    colframe=black!82,boxrule=0pt,
    left=5pt,right=5pt,top=2pt,bottom=2pt,width=\linewidth]
    \color{white}\small\bfseries Model Outputs
  \end{tcolorbox}
  \vspace{2pt}
  \noindent
  \begin{minipage}[t]{0.485\linewidth}
    \begin{tcolorbox}[enhanced,sharp corners=all,colback=green!4,
      colframe=green!40!black,boxrule=0.4pt,
      left=4pt,right=4pt,top=3pt,bottom=3pt,width=\linewidth]
      \small{\bfseries\color{green!40!black}$\blacktriangleright$
        Fine-Tuned Model (OneVision$^\dagger$)}\\[2pt]
      \textbf{Better Sequence: B.}\\[2pt]
      Sequence~B correctly tracks the cooking progression.
      Sequence~A opens with a blender containing red liquid ---
      unrelated to any step --- a semantic outlier that immediately
      invalidates the sequence.
    \end{tcolorbox}
  \end{minipage}\hfill
  \begin{minipage}[t]{0.485\linewidth}
    \begin{tcolorbox}[enhanced,sharp corners=all,colback=orange!5,
      colframe=orange!50!black,boxrule=0.4pt,
      left=4pt,right=4pt,top=3pt,bottom=3pt,width=\linewidth]
      \small{\bfseries\color{orange!60!black}$\blacktriangleright$
        Zero-Shot Baseline (\textsc{OneVision})}\\[2pt]
      \textbf{Better Sequence: B.}\\[2pt]
      Sequence~B maps each step correctly from oil to final dish.
      Sequence~A's first frame --- a blender with red liquid ---
      has no correspondence to step~1, breaking the flow.
    \end{tcolorbox}
  \end{minipage}
\end{tcolorbox}

\vspace{8pt}

\captionof{table}{%
  \textbf{Case Study~4.} \textsc{Prism}-Semantic, strong perturbation
  at position~4. A seeds/nuts close-up replaces the ramekin-removal
  frame. The fine-tuned model provides a more precise step-level
  rationale than the zero-shot baseline.}
\label{tab:example_4}
\begin{tcolorbox}[enhanced,sharp corners=all,colback=white,
  colframe=black!20,boxrule=0.7pt,
  left=4pt,right=4pt,top=4pt,bottom=4pt,width=\textwidth]
  \begin{tcolorbox}[enhanced,sharp corners=all,colback=black!82,
    colframe=black!82,boxrule=0pt,
    left=5pt,right=5pt,top=2pt,bottom=2pt,width=\linewidth]
    \color{white}\small\bfseries Visual Input Sequences
  \end{tcolorbox}
  \vspace{2pt}
  \begin{minipage}[t]{0.485\linewidth}
    \begin{tcolorbox}[enhanced,sharp corners=all,colback=gray!4,
      colframe=black!25,boxrule=0.4pt,
      left=2pt,right=2pt,top=2pt,bottom=2pt,width=\linewidth]
      \centering{\small\bfseries Sequence A}\\[2pt]
      \includegraphics[width=\linewidth,height=0.11\textheight,
        keepaspectratio]{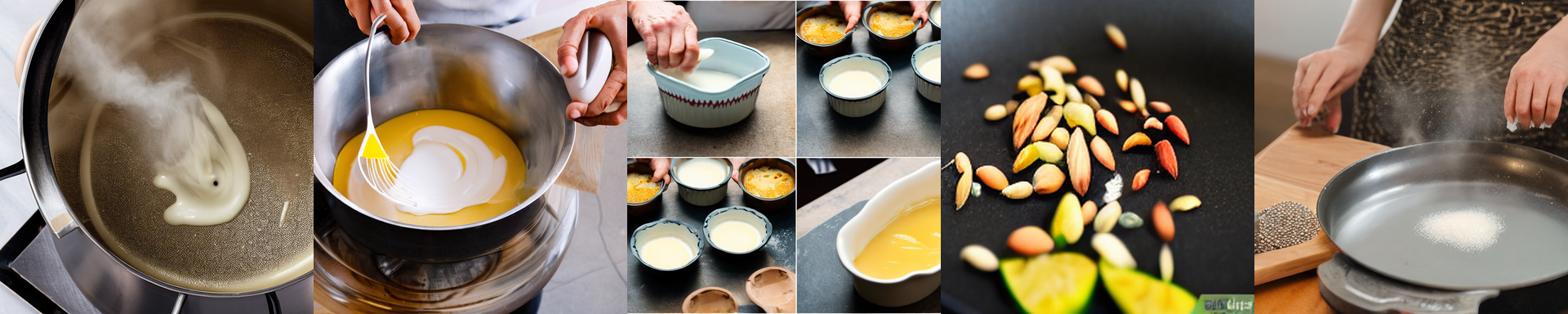}
    \end{tcolorbox}
  \end{minipage}\hfill
  \begin{minipage}[t]{0.485\linewidth}
    \begin{tcolorbox}[enhanced,sharp corners=all,colback=gray!4,
      colframe=black!25,boxrule=0.4pt,
      left=2pt,right=2pt,top=2pt,bottom=2pt,width=\linewidth]
      \centering{\small\bfseries Sequence B}\\[2pt]
      \includegraphics[width=\linewidth,height=0.11\textheight,
        keepaspectratio]{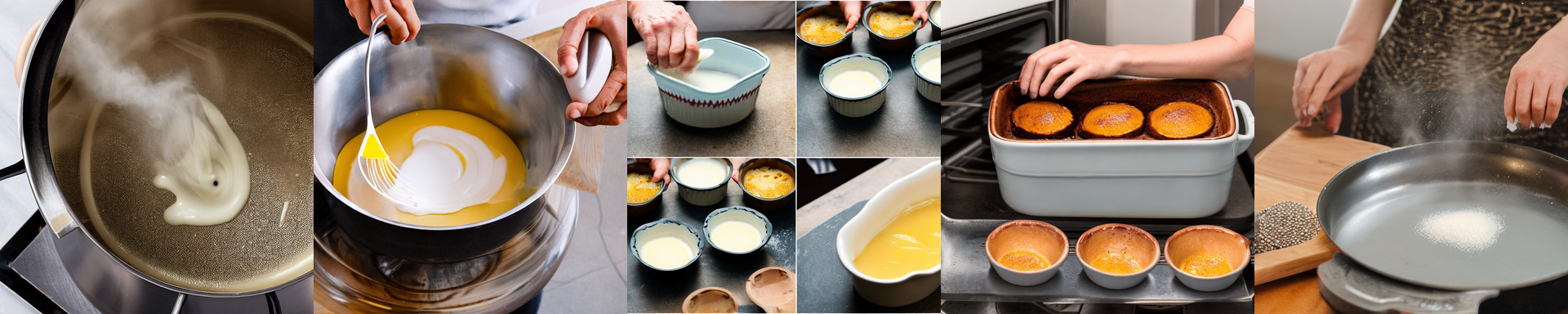}
    \end{tcolorbox}
  \end{minipage}
  \vspace{3pt}
  \begin{tcolorbox}[enhanced,sharp corners=all,colback=blue!4,
    colframe=blue!28,boxrule=0.4pt,
    left=5pt,right=5pt,top=2pt,bottom=2pt,width=\linewidth]
    {\small\bfseries Prompt:}\enspace\small\itshape
    Evaluate for: 1.\,Simmering cream; 2.\,Whisking yolks;
    3.\,Water bath; 4.\,Removing ramekins from oven;
    5.\,Sprinkling sugar.
  \end{tcolorbox}
  \vspace{3pt}
  \begin{tcolorbox}[enhanced,sharp corners=all,colback=black!82,
    colframe=black!82,boxrule=0pt,
    left=5pt,right=5pt,top=2pt,bottom=2pt,width=\linewidth]
    \color{white}\small\bfseries Model Outputs
  \end{tcolorbox}
  \vspace{2pt}
  \noindent
  \begin{minipage}[t]{0.485\linewidth}
    \begin{tcolorbox}[enhanced,sharp corners=all,colback=green!4,
      colframe=green!40!black,boxrule=0.4pt,
      left=4pt,right=4pt,top=3pt,bottom=3pt,width=\linewidth]
      \small{\bfseries\color{green!40!black}$\blacktriangleright$
        Fine-Tuned Model (OneVision$^\dagger$)}\\[2pt]
      \textbf{Better Sequence: B.}\\[2pt]
      Sequence~A fails at step~4: seeds/nuts in a pan are
      incoherent with the custard process. Sequence~B maintains
      consistent content across all five steps.
    \end{tcolorbox}
  \end{minipage}\hfill
  \begin{minipage}[t]{0.485\linewidth}
    \begin{tcolorbox}[enhanced,sharp corners=all,colback=orange!5,
      colframe=orange!50!black,boxrule=0.4pt,
      left=4pt,right=4pt,top=3pt,bottom=3pt,width=\linewidth]
      \small{\bfseries\color{orange!60!black}$\blacktriangleright$
        Zero-Shot Baseline (\textsc{OneVision})}\\[2pt]
      \textbf{Better Sequence: B.}\\[2pt]
      Step~4 in Seq.~B correctly shows ramekin removal from the oven.
      Seq.~A's step~4 --- seeds or nuts --- is an unrelated outlier
      disrupting the custard narrative.
    \end{tcolorbox}
  \end{minipage}
\end{tcolorbox}

\end{table*}

\begin{table*}[htbp]
\centering

\captionof{table}{%
  \textbf{Case Study~5.} \textsc{Prism}-Temporal swap at positions~1--2
  (vanilla extract). Seq.~B shows the filled jar before the bean is
  halved. The fine-tuned model names the causal violation explicitly.}
\label{tab:example_5}
\setlength{\fboxsep}{0pt}
\begin{tcolorbox}[enhanced,sharp corners=all,colback=white,
  colframe=black!20,boxrule=0.7pt,
  left=4pt,right=4pt,top=4pt,bottom=4pt,width=\textwidth]
  \begin{tcolorbox}[enhanced,sharp corners=all,colback=black!82,
    colframe=black!82,boxrule=0pt,
    left=5pt,right=5pt,top=2pt,bottom=2pt,width=\linewidth]
    \color{white}\small\bfseries Visual Input Sequences
  \end{tcolorbox}
  \vspace{2pt}
  \begin{minipage}[t]{0.485\linewidth}
    \begin{tcolorbox}[enhanced,sharp corners=all,colback=gray!4,
      colframe=black!25,boxrule=0.4pt,
      left=2pt,right=2pt,top=2pt,bottom=2pt,width=\linewidth]
      \centering{\small\bfseries Sequence A}\\[2pt]
      \includegraphics[width=\linewidth,height=0.11\textheight,
        keepaspectratio]{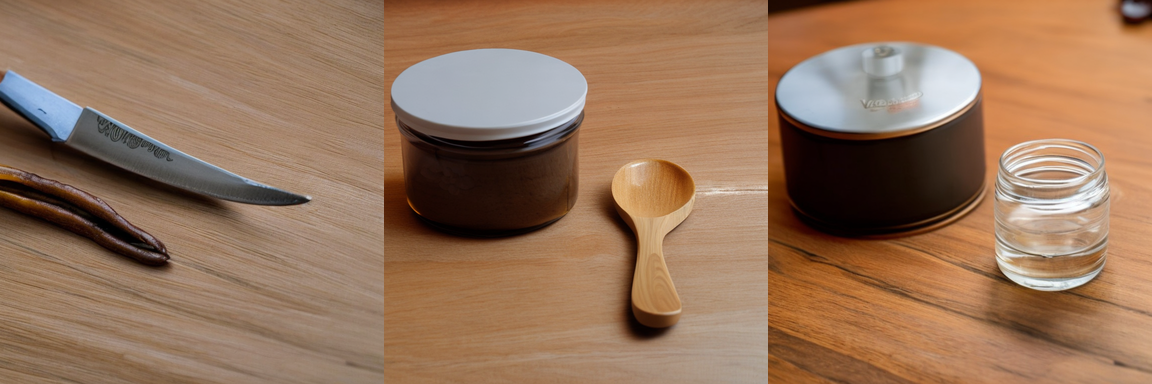}
    \end{tcolorbox}
  \end{minipage}\hfill
  \begin{minipage}[t]{0.485\linewidth}
    \begin{tcolorbox}[enhanced,sharp corners=all,colback=gray!4,
      colframe=black!25,boxrule=0.4pt,
      left=2pt,right=2pt,top=2pt,bottom=2pt,width=\linewidth]
      \centering{\small\bfseries Sequence B}\\[2pt]
      \includegraphics[width=\linewidth,height=0.11\textheight,
        keepaspectratio]{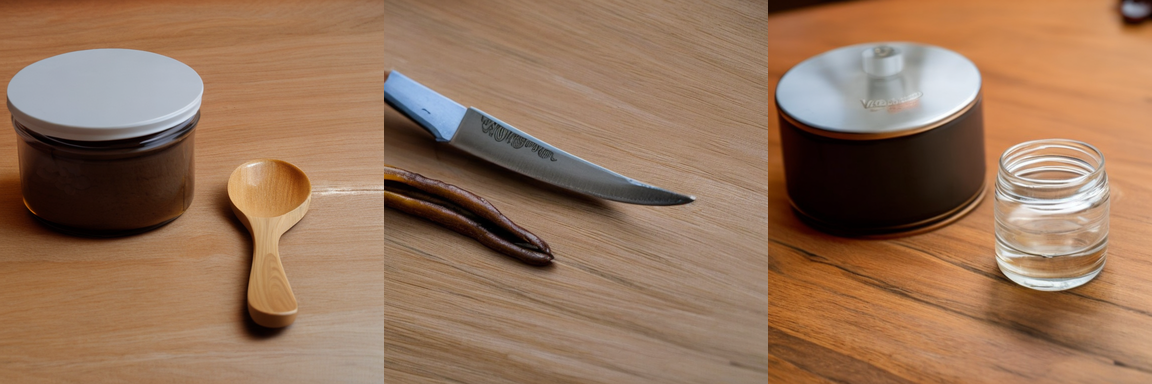}
    \end{tcolorbox}
  \end{minipage}
  \vspace{3pt}
  \begin{tcolorbox}[enhanced,sharp corners=all,colback=blue!4,
    colframe=blue!28,boxrule=0.4pt,
    left=5pt,right=5pt,top=2pt,bottom=2pt,width=\linewidth]
    {\small\bfseries Prompt:}\enspace\small\itshape
    Evaluate for: 1.\,Vanilla bean pod halved;
    2.\,Jar filled with substance; 3.\,Sealed jar next to vodka.
  \end{tcolorbox}
  \vspace{3pt}
  \begin{tcolorbox}[enhanced,sharp corners=all,colback=black!82,
    colframe=black!82,boxrule=0pt,
    left=5pt,right=5pt,top=2pt,bottom=2pt,width=\linewidth]
    \color{white}\small\bfseries Model Outputs
  \end{tcolorbox}
  \vspace{2pt}
  \noindent
  \begin{minipage}[t]{0.485\linewidth}
    \begin{tcolorbox}[enhanced,sharp corners=all,colback=green!4,
      colframe=green!40!black,boxrule=0.4pt,
      left=4pt,right=4pt,top=3pt,bottom=3pt,width=\linewidth]
      \small{\bfseries\color{green!40!black}$\blacktriangleright$
        Fine-Tuned Model (OneVision$^\dagger$)}\\[2pt]
      \textbf{Better Sequence: A.}\\[2pt]
      Sequence~A follows the correct causal order: pod halved
      (step~1) before jar filled (step~2). Sequence~B inverts
      these steps --- a direct causal violation.
    \end{tcolorbox}
  \end{minipage}\hfill
  \begin{minipage}[t]{0.485\linewidth}
    \begin{tcolorbox}[enhanced,sharp corners=all,colback=orange!5,
      colframe=orange!50!black,boxrule=0.4pt,
      left=4pt,right=4pt,top=3pt,bottom=3pt,width=\linewidth]
      \small{\bfseries\color{orange!60!black}$\blacktriangleright$
        Zero-Shot Baseline (\textsc{OneVision})}\\[2pt]
      \textbf{Better Sequence: A.}\\[2pt]
      Sequence~A matches the described process step by step.
      Sequence~B presents step~2 before step~1, disrupting
      the logical progression from preparation to containment.
    \end{tcolorbox}
  \end{minipage}
\end{tcolorbox}

\vspace{8pt}

\captionof{table}{%
  \textbf{Case Study~6.} \textsc{Prism}-Temporal swap at position~4--1
  (cake decoration). Seq.~B places the decoration frame first,
  before any batter exists. The fine-tuned model names the
  causal violation; the zero-shot baseline reasons globally.}
\label{tab:example_6}
\begin{tcolorbox}[enhanced,sharp corners=all,colback=white,
  colframe=black!20,boxrule=0.7pt,
  left=4pt,right=4pt,top=4pt,bottom=4pt,width=\textwidth]
  \begin{tcolorbox}[enhanced,sharp corners=all,colback=black!82,
    colframe=black!82,boxrule=0pt,
    left=5pt,right=5pt,top=2pt,bottom=2pt,width=\linewidth]
    \color{white}\small\bfseries Visual Input Sequences
  \end{tcolorbox}
  \vspace{2pt}
  \begin{minipage}[t]{0.485\linewidth}
    \begin{tcolorbox}[enhanced,sharp corners=all,colback=gray!4,
      colframe=black!25,boxrule=0.4pt,
      left=2pt,right=2pt,top=2pt,bottom=2pt,width=\linewidth]
      \centering{\small\bfseries Sequence A}\\[2pt]
      \includegraphics[width=\linewidth,height=0.11\textheight,
        keepaspectratio]{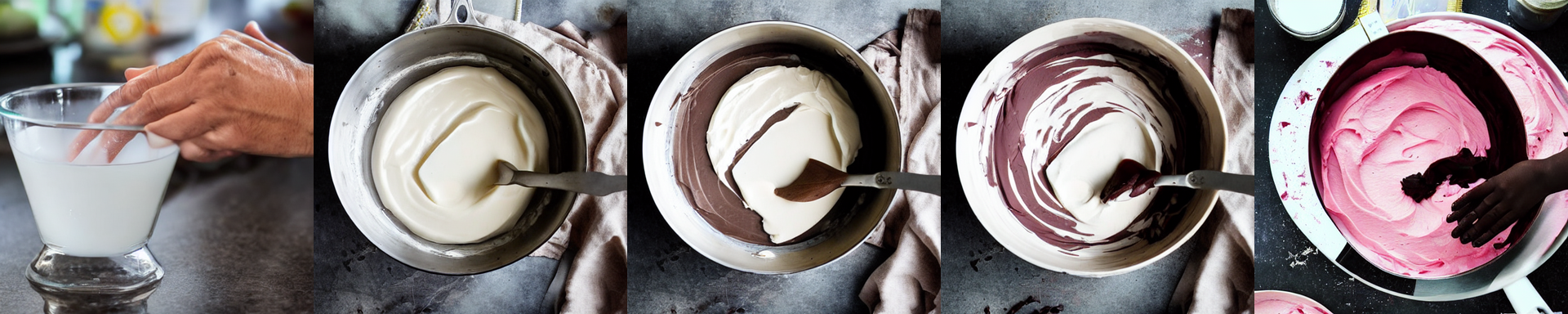}
    \end{tcolorbox}
  \end{minipage}\hfill
  \begin{minipage}[t]{0.485\linewidth}
    \begin{tcolorbox}[enhanced,sharp corners=all,colback=gray!4,
      colframe=black!25,boxrule=0.4pt,
      left=2pt,right=2pt,top=2pt,bottom=2pt,width=\linewidth]
      \centering{\small\bfseries Sequence B}\\[2pt]
      \includegraphics[width=\linewidth,height=0.11\textheight,
        keepaspectratio]{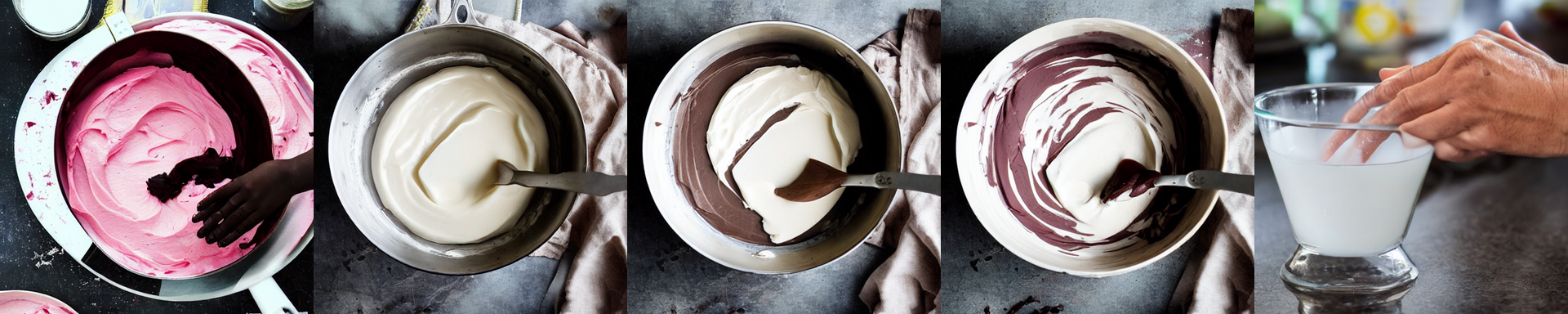}
    \end{tcolorbox}
  \end{minipage}
  \vspace{3pt}
  \begin{tcolorbox}[enhanced,sharp corners=all,colback=blue!4,
    colframe=blue!28,boxrule=0.4pt,
    left=5pt,right=5pt,top=2pt,bottom=2pt,width=\linewidth]
    {\small\bfseries Prompt:}\enspace\small\itshape
    Evaluate for: 1.\,Pouring milk; 2.\,Beating batter;
    3.\,Filling cake pan; 4.\,Mixing cr\`eme fra\^iche;
    5.\,Decorating with sprinkles.
  \end{tcolorbox}
  \vspace{3pt}
  \begin{tcolorbox}[enhanced,sharp corners=all,colback=black!82,
    colframe=black!82,boxrule=0pt,
    left=5pt,right=5pt,top=2pt,bottom=2pt,width=\linewidth]
    \color{white}\small\bfseries Model Outputs
  \end{tcolorbox}
  \vspace{2pt}
  \noindent
  \begin{minipage}[t]{0.485\linewidth}
    \begin{tcolorbox}[enhanced,sharp corners=all,colback=green!4,
      colframe=green!40!black,boxrule=0.4pt,
      left=4pt,right=4pt,top=3pt,bottom=3pt,width=\linewidth]
      \small{\bfseries\color{green!40!black}$\blacktriangleright$
        Fine-Tuned Model (OneVision$^\dagger$)}\\[2pt]
      \textbf{Better Sequence: A.}\\[2pt]
      Sequence~A preserves the correct causal order from milk
      preparation through to decoration. Sequence~B opens with
      the decoration frame --- before any batter exists ---
      a direct violation of recipe logic.
    \end{tcolorbox}
  \end{minipage}\hfill
  \begin{minipage}[t]{0.485\linewidth}
    \begin{tcolorbox}[enhanced,sharp corners=all,colback=orange!5,
      colframe=orange!50!black,boxrule=0.4pt,
      left=4pt,right=4pt,top=3pt,bottom=3pt,width=\linewidth]
      \small{\bfseries\color{orange!60!black}$\blacktriangleright$
        Zero-Shot Baseline (\textsc{OneVision})}\\[2pt]
      \textbf{Better Sequence: A.}\\[2pt]
      Sequence~A follows the process from preparation to finish.
      Sequence~B places the final decoration step first,
      disrupting the logical flow of the recipe.
    \end{tcolorbox}
  \end{minipage}
\end{tcolorbox}
\end{table*}


\end{document}